\documentclass[MNA]{mna} % 

\usepackage{qtree}
\usepackage[all]{xy}
\usepackage[normalem]{ulem}

\SHORTTITLE{Merge in function spaces}

\TITLE{Encoding syntactic objects and Merge operations in function spaces}

\AUTHORS{%
	Matilde~Marcolli\footnote{Department of Mathematics and Department of Computing and Mathematical Sciences, California Institute of Technology, CA 91125, USA.
		\EMAIL{matilde@caltech.edu}}
	\and %% remove this line and below if single author
	Robert C.~Berwick\footnote{Institute for Data Systems and Society, 
		Massachusetts Institute of Technology, Cambridge, MA 02139, USA.
		\EMAIL{berwick@csail.mit.edu}}}
\SHORTAUTHOR{M.~Marcolli, R.C.~Berwick}


\AMSSUBJ{91F20; 16Y60; 16T05; 92C20}

\SUBMITTED{August 2, 2025} 
\ACCEPTED{August 19, 2026}

\ARXIVID{2507.13501} % Edit.
\VOLUME{6}
\YEAR{2026}
\PAPERNUM{3}
\DOI{10.46298/mna.16273}

\ABSTRACT{
	We provide a mathematical argument showing that, given a representation
	of lexical items as functions (wavelets, for instance) in some function space,
	it is possible to construct a faithful representation of arbitrary syntactic objects 
	in the same function space. This space can be endowed with a commutative nonassociative
	semiring structure built using the second R\'enyi entropy. The use of this entropy function
	can be interpreted as a measure of diversity on the representation of the
	lexical items. The resulting representation of syntactic objects is compatible
	with the magma structure. The resulting set of functions is an algebra over an operad,
	where the operations in the operad model circuits that transform the input
	wave forms into a combined output that encodes the syntactic structure. A
	neurocomputational realization of these circuits is completely determined by
	a realization of a single binary gate that implements a minimization and an 
	entropy computation. The action of Merge on workspaces is faithfully implemented 
	as action on these circuits, through a coproduct and a Hopf algebra Markov chain. 
	The results obtained here provide a constructive argument showing the theoretical possibility of
	a neurocomputational realization of the core computational structure of
	syntax. We also present a particular case of this general construction
	where this type of realization of Merge is implemented as a cross frequency
	phase synchronization on sinusoidal waves. This also shows that Merge
	can be expressed in terms of the successor function of a semiring,
	thus clarifying the well known observation of its similarities with the
	successor function of arithmetic. 
} 

\DeclareMathOperator{\Ry}{Ry}
\DeclareMathOperator{\Conf}{Conf}
\DeclareMathOperator{\Magma}{Magma}
\DeclareMathOperator{\Ts}{Ts}
\DeclareMathOperator{\Dom}{Dom}
\DeclareMathOperator{\Sym}{Sym}
\DeclareMathOperator{\Id}{Id}
\DeclareMathOperator{\id}{id}
\DeclareMathOperator*{\argmin}{argmin}
\DeclareMathOperator{\Span}{span}
\DeclareMathOperator{\Sh}{Sh}

\begin{document}

%%%%%%%%%%%%%%%%%%%%%%%%%%%%%%%%%%%%%%%%%%%%%%%%%%%%%%%%%%%%%%%%%%%
%%                                                               %%
%% No need for \maketitle.                                       %%
%%                                                               %%
%%%%%%%%%%%%%%%%%%%%%%%%%%%%%%%%%%%%%%%%%%%%%%%%%%%%%%%%%%%%%%%%%%%

%%%%%%%%%%%%%%%%%%%%%%%%%%%%%%%%%%%%%%%%%%%%%%%%%%%%%%%%%%%%%%%%%%%
%%                                                               %%
%% Please replace what follows by the body of your article       %%
%% (up to the bibliography):                                     %%
%%                                                               %%
%%%%%%%%%%%%%%%%%%%%%%%%%%%%%%%%%%%%%%%%%%%%%%%%%%%%%%%%%%%%%%%%%%%

\section{Introduction}

In the context of Biolinguistics, intended broadly as the study of the biological mechanism 
underlying the structure, evolutionary development, and acquisition of human language 
(see \cite{Boeckx,Jenk}), and in particular of Neurolinguistics, focusing on the 
neural mechanisms underlying the production,
parsing, and acquisition of language (see \cite{Brenn2,Zubi}), 
an important challenge is identifying neurocomputational
mechanisms that can provide a brain correlate of the fundamental computational structure of syntax
in human language. An overview of this problem is presented, for instance, in \cite{SchBoSch,EmbPo}.
In the current model of generative linguistics, the Strong Minimalist Thesis (as developed
by Chomsky starting around 2013, see \cite{ChomskyGK,ChomskyElements}), the core
computational structure of syntax is described in terms of the ``free symmetric Merge'' that
is responsible for both structure formation (External Merge) and movement (Internal Merge).
A significant advantage of this model is that it decouples this core computational structure
from other aspects, such as parametric variation across individual languages, or the interface
between syntax and semantics, \cite{Borer}. Another advantage of this recent formulation is that it
admits a mathematical formalization (as developed in \cite{MCB}). This allows for significant
advances in terms of formulating some linguistic questions in a form suitable for the
employment of mathematical reasoning and modeling. In this paper, we discuss how the
mathematical formulation can be used to gain some theoretical insights on the question
of possible neurocomputational realization of the Merge operation.

Different aspects of this broad question on  neurocomputational realizations of syntactic
structures have been analyzed, see for instance \cite{BickSza,Frie,Fukui,Martin1,Murphy1}.
(Additional literature will be discussed briefly in \S \ref{ROSEsec}.)
In this paper we take a more theoretical viewpoint and we give a mathematical proof
that, if lexical items are encoded as brain waves (which mathematically means that
there is an embedding of the set of lexical items in a space of wavelet functions),
then this map can be extended to an embedding of syntactic objects, which moreover
preserves the algebraic structure (nonassociative commutative magma) that is crucial
to structure formation. We show that this embedding can be obtained through
neural circuits made of a composition of gates that realize a simple binary operation
consisting of a minimization and an entropy computation. It is important to stress that
this argument is a theoretical (constructive) proof of concept: it does not imply that
an actual neurocomputational realization will necessarily have to be exactly of this form, but it
shows a possible theoretical mechanism through which such a realization (with the
correct computational properties) can be obtained.  
A longer term goal of this type of mathematical questions would be to develop possible general
theoretical guidelines, primarily based on algebraic properties, that can constrain 
neurocomputational implementations of Merge and therefore may suggest novel
constraints for experimental testing. Additional assumptions, beyond what is discussed here,
would be required to derive such constraints, but the main approach that we suggest  
is inspired by the modeling of physical systems, where a historically very successful strategy has 
been the use of algebraic properties for the purpose of constraining the model, so as to 
drastically reduce the size of the space of possibilities. 

Thus, we focus here on the problem of how to create an embedding of syntactic objects, and of
the action of Merge on workspaces, inside a space of functions, in terms of a (nonlinear) superposition 
of certain wavelets (elements of some function space). The reason for investigating this type of
encoding as functions (typically some kind of wavelets) comes from neurolinguistic questions.

It was variously observed in the neuroscience of language that speech and language parsing
and processing involve measurable effects of phase synchronization, 
phase coupling, phase-amplitude coupling (PAC) in brain waves and oscillations, for
instance \cite{Baasti1, Baasti2, BreMartin,CoopMartin, CoopMartin2,Ding, Kau,Martin1,Meyer1,Meyer2, Molinaro,Oever,Poe,Prysta, Rimme,Sega,Slaats,Zhao1, Zhao2}.  
An important question is how to separate effects
due to statistical properties of strings of text (similar to mechanisms now seen in artificial models of
language production) from effects that capture the core computational mechanism of syntax
and the hierarchical structures present in language. Most of the literature in the 
neuroscience of language focuses on effects of speech listening. There are a number of
empirical results that point more specifically to connections between synchronization and syntax,
for instance \cite{BaiMartin,BreMartin, MeyerGum, WeissMartin}. 
In \cite{WeissMartin} in particular, the statistical and the structural effects and their interactions 
are analyzed and distinguished. Ultimately, what one would like to achieve is a theoretical
model that can be fully formalized in a mathematical and computational way, and that
leads to empirically verifiable descriptions of how the structure of syntax and the fundamental
structure building operations of syntax, as a computational system, can be realized in 
neural population activity and in observable rhythms and brain oscillations.

While we do not try here to address this complex problem, we will provide a flexible
template of how an operation with the prescribed algebraic properties of Merge and
the resulting syntactic objects generated by a magma operation can be faithfully
embodied in spaces of wavelet functions. This mathematical model can in principle
be flexibly adapted to at least some of the existing theoretical models proposed in 
language neuroscience, providing a way of incorporating an operation with the
correct formal properties of the syntactic Merge.

For instance,  we discuss briefly some aspects of two recent theoretical models:
the Compositional Neural Architecture model of \cite{Martin} (see also
\cite{MartinDou1,MartinDou2,MartinDou3}), based on a 
symbolic-connectionist DORA (Discovery of Relations by Analogy) model and
on a ``time-based binding'' that produces synchronization phenomena related
to hierarchical structures, and the ROSE (Representation, Operation, Structure, Encoding) 
theoretical model of \cite{Murphy1,Murphy2,Murphy3}, that proposes
a neurocomputational realization of Merge in terms of brain oscillations
and synchronization and modulation phenomena across different types of
brain waves. Inspired by these and 
other discussions in neurolinguistics on representations
of hierarchical structures starting from waves or time series
representations of lexical items, we develop a special
form of our model where the action of Merge in such 
representations is implemented via phase synchronizations.

Another motivation comes from brain-machine interface experiments that indicate
neural representations of internal and vocalized words, at both the single neuron and population levels,
\cite{Dash,Metz,Moses,WBPLLA}. Current brain-machine interface
models are typically based on phonetic (not lexical or semantic) representations, and  
use ML analysis of high gamma activity for word prediction, rather than  
identifying interpretable neural representations. Further experiments (see \cite{Proix})
use cross-frequency coupling for decoding, rather than just high or low frequency activity. 
All these different experiments do, in any case, 
raise the theoretical question of whether representations of
hierarchical structures of syntax would also be similarly possible, going beyond just 
speech as a sequence of words, and on to the level of language as hierarchical structures,
\cite{Huij1, Huij2}. We do not discuss in this paper
the interface between syntax and morphology (which can be articulated within the
mathematical formulation of Minimalism, \cite{SentMar}), though brain-wave models
of the syntax-morphology interface have also been proposed in \cite{Martin, Murphy1}:  
an investigation of possible neurocomputational models of the algebraic structure
described in \cite{SentMar} is an interesting question.

The point of view we follow in this note is simply to consider the underlying mathematical
question, and provide a constructive proof that it is possible indeed to faithfully and efficiently 
encode syntactic objects with the algebraic structure of commutative nonassociative
magma, and movement by Internal Merge, starting with a given embedding of the set of
lexical items and syntactic features in a function space.

As a model of the core computational structure of syntax, we follow here the 
formulation of Chomsky's free symmetric Merge in
the Strong Minimalist Thesis, \cite{ChomskyGK,ChomskyElements}, 
in the mathematical  form developed in the work of Marcolli, Chomsky, Berwick, \cite{MCB}.

The construction that we present here can be seen in the same light as the
model of syntax-semantics interface discussed in \cite{MCB}. However, we
focus here on a somewhat different type of property and a very different
interpretation. At the level of formal properties, the embedding we consider takes
values in a (nonassociative) semiring, 
where the nonassociative but commutative semiring addition will account for the
magma structure operation on syntactic objects. (The reader should keep in mind
that we will use the term {\em semiring} by including the 
possibility of {\em nonassociative semirings}, where multiplication is associative 
but addition is not, so multiplication is a monoid but addition is a magma. Note that this type of
algebraic structure is very different from nonassociative algebras, where it is the
multiplication operation that fails associativity. Most examples of such nonassociative
semirings arise, as in the case we consider, from deformations of ordinary 
associative semirings.) The main reasons for this choice of algebraic structure
are that it reflects the key algebraic properties of the unconstrained free 
Merge operation in the Minimalism model of syntax, commutativity and nonassociativity, and that such
target semiring structure can be realized in spaces of functions, through an
averaging operation optimized with respect to an information function. This is a setting
that, as we will discuss, more closely relates to computational mechanisms
like cross-frequency phase synchronization on wave functions that are known to play
a role in neurocomputation. The action of Merge 
can then be described as an action on circuits that compute this embedding
of syntactic objects starting from the lexical items.
As mentioned, to obtain the nonassociative addition operation in the semiring,
the construction we present crucially uses an information measure, the second R\'enyi entropy,
which can be seen as a diversity measure for the dictionary encoding the lexical items.

Our main goal is to provide a model that embeds syntactic objects with the Merge operation
into a space of functions, assuming that lexical items are themselves encoded in the same
function space. To this goal we focus on two key properties: mapping the Merge operation
to a corresponding operation that has the same algebraic properties (specifically, commutativity
and nonassociativity), and ensuring the faithfulness of the embedding of syntactic objects in the
target space of functions. The reason for requiring the same algebraic structure is that it is
a first key {\em necessary condition} for a possible realization of the action of Merge: if 
two operations have different algebraic structures then they certainly cannot be equivalent
operations, so if we want to propose a possible realization of the unconstrained free symmetric 
Merge as an operation on a certain set of functions, then the operation we construct has to, 
at the very least, satisfy these properties.  We will also comment more extensively in \S \ref{revMergeSec}
below, on the specific meaning of the nonassociativity and commutativity properties in the
linguistic model: that will provide additional reasons why we focus mainly on these properties.
There are additional properties to Merge, beyond the commutativity and nonassociativity
of the magma of syntactic objects, and one can see more algebraic structure when considering 
the action of Merge on workspaces. Such structures also involve topological notions, when
considering the resulting dynamical system, for example as discussed in \cite{MarSki}, or
in interface with semantics (as in Chapter 3 of \cite{MCB}). We will not discuss all these
aspects in the present paper, but we will discuss passing from a model of the magma
of syntactic objects to a model of the action of Merge on workspaces in \S \ref{IMsec}.
The faithfulness requirement, which needs a fair amount of technical mathematical work
to establish, is also a key basic necessary condition. If we want to propose a mechanism that
realizes syntactic objects, then what we present as the representation of the abstract 
syntactic objects inside a space of functions (such as wavelets, sinusoidal waves, for instance) 
has to be able, at the very least, to reconstruct the syntactic objects faithfully, otherwise
we just cannot claim to have obtained something equivalent. An encoding that is non-faithful
is only a partial encoding with possibly significant loss of information and structure. 
Note that, when we talk about obtaining a faithful representation of syntactic objects
in a given space of functions, we mean this as an equivalence in a structural algebraic sense,
not in terms of a specific biological implementation.

In more direct mathematical terms, our first goal here is to construct a magma representation
of the magma of syntactic objects (which we recall in \S \ref{revMergeSec}) in a vector space
of wavelet-type functions, in such a way that the corresponding operation can be instantiated
in terms of operations well known in the neuroscience setting, such as cross-frequency phase
synchronization of sinusoidal waves. Because of nonassociativity, the notion of ``representation''
is much more subtle than for associative algebraic structures. In the associative case, a linear
representation is a morphism to an algebra of linear operators, in which the associative operation
becomes composition. To obtain something valuable and nontrivial in a nonassociative setting,
we need the target vector space to be itself endowed with a nonassociative operation that
is in an appropriate sense compatible with the linear space structure and that can then receive
a morphism of nonassociative magmas from the magma of syntactic objects. A good choice of
structure that does satisfy these properties is the class of (nonassociative) semirings, called
thermodynamic semirings, introduced in \cite{MarThor}, where the linear addition of the underlying
vector space becomes the (associative) semiring multiplication, while a compatible idempotent 
addition operation (the usual tropical semiring)  is deformed to a nonassociative form, endowing the underlying
vector space with the type of structure that makes it possible to talk about magma representations
in a meaningful way. The resulting functions associated to syntactic objects and the 
(nonassociative) addition operation of the thermodynamic semiring 
look quite different from the original combinatorial syntactic objects with their magma operation.
Thus, in order to establish that this is indeed a good way of encoding syntactic objects along with their 
magma structure into a vector space of wavelet-type functions, we need to also ensure that
the original syntactic objects remain ``visible'' (and recoverable) in their image, and this is the
reason why we need to also include a more elaborate argument on faithfulness. 

It is important to stress that 
the model we present is simply aimed at showing that such encoding of syntactic
objects and syntactic operations is theoretically possible, given an encoding of lexical items 
by certain wavelet functions: it does not necessarily provide a direct
realization in neurocomputational terms. We will give some indication as to the
realizability of the entropy computations required in this model, to show
that it is not unreasonable to expect those to be implementable in neuronal circuits.

In particular, in the last section of this paper, we discuss some aspects of the
existing literature and possible relations to the model we present here. We include
a brief discussion of some aspects of the Compositional Neural Architecture model of \cite{Martin} 
and of the ROSE model of \cite{Murphy2}, in the light of the construction presented here. 
Although we do not try to formalize or implement any of these existing models, we
focus on one particular aspect they share, which is the role of cross-frequency
and phase-phase synchronization of waves. We use this as motivation for presenting
a simpler instantiation of the general model described in the previous sections in terms of
phase synchronization of simple sinusoidal waves, where one can make some
properties of our model more explicit.

In particular, this construction via phase synchronizations shows an interesting
property of Merge, namely how Merge can be expressed in terms of a ``successor function''.
Chomsky had noted, on various occasions in the past, how Merge bears some similarities 
to the successor function of arithmetic (see for instance \cite{Chomsky2020}). However,
these two operations have different algebraic properties, so they cannot be directly
related. On the other hand, the thermodynamic semirings of \cite{MarThor} that we use 
in this paper also have a successor function (generalizing the usual successor function
of arithmetic associated to the semiring $\mathbb{N}$), as shown in \cite{MarThor}. We show here
that this kind of successor function is indeed the one that can be used to describe Merge.
This clarifies, in a mathematically precise way, the relation between Merge and the
successor function, by identifying a natural modification to the successor function of
arithmetic, which is still the successor function of another more general type of semiring
different from the natural numbers, and which can be directly related to Merge.

Our model of the Merge operation in function spaces of wavelet-like signals also answers
a question posed by Andrea Martin in \cite{Martin}: whether composition via vector addition
is a sufficient operation for compositionality in natural language. We confirm the fact that
the compositionality of Merge does not have the properties of a tensor product type
operation (multiplicative combination), but rather behaves like a sum (additive combination), 
as observed in \cite{Martin}. However, we show that vector addition does not have the
right algebraic properties to implement structure formation by Merge. On the other hand,
the solution we present, that does faithfully encode the properties of Merge, is obtained
by optimizing over convex combinations of vectors using an entropy functional that
can be chosen in a natural way (the R\'enyi entropy) so as to obtain the correct
algebraic properties for the Merge operation.

\section{Syntactic Objects and Weighted Combinations} \label{WeightcombSec}

The underlying theoretical model of syntax that we assume in this paper is
the most recent form of Minimalism in generative linguistics, as developed
by Chomsky starting around 2013, and presented in \cite{ChomskyUCLA}
and \cite{ChomskyGK} and in an extensive introductory form in 
\cite{ChomskyElements}. This model was given a mathematical formalization
in the book \cite{MCB} by Chomsky and the authors.

Our goal in the present paper is not to seek empirical verifications for 
this model, but simply, assuming it as our theoretical model, to verify the
theoretical feasibility of faithfully realizing, in function spaces suitable for
neurocomputation, the same core computational structure that 
we identified in \cite{MCB} as underlying the syntax-formation action of Merge. 
As we summarize the basic linguistics background in \S \ref{revMergeSec},
we will also comment further on how to interpret what follows in the rest of
the paper, in terms of theory development versus empirical aspects.

\subsection{Summary of Merge and its mathematical model}\label{revMergeSec}

We start with a very short summary of the main notions, from the mathematical
formulation of the free symmetric Merge of Minimalism developed in \cite{MCB},
that we need to use in the rest of the paper. We refer the reader to \cite{MCB}
for the details and more extensive discussions.

In this theoretical model, syntax formation is subdivided into a core computational mechanism,
the free symmetric Merge, and two fundamental interfaces, usually referred to as
Externalization (or Sensory-Motor interface) and Syntax-Semantics interface (or 
Conceptual-Intentional interface). In practice, what this distinction boils down to is the
idea that a single common computational structure is responsible for the generation of
syntax in human language, in a universal form that is, in itself, independent of which particular 
language an individual learns during the process of language acquisition. This core
computational mechanism simply builds syntactic objects from data of lexical items
and syntactic features, and performs transformational movement on these objects. 
The results of this dynamical activity are subjected to systems of filtering that eliminate
non-viable syntactic objects. This filtering is of three different types. There are
filters that are still independent of any particular language, such as those that govern
the structure of {\em phases} (the head--complement--specifier structure, the
head projection, and the hierarchical arrangement of these structures), the assignment
of theta roles (predicates and arguments). There are filters that instead depend on
the particular language that is acquired (setting of syntactic parameters), such as
specific word order structure. Moreover, there is also further filtering at the Syntax-Semantics
interface for viable parsing in semantics. The simplicity and elegance of this theoretical
model lies in its capacity for decoupling these structures, which is a key factor allowing 
for a good mathematical modeling of the theory.  Since the language we observe in
empirical data is the result of this additional filtering, it is difficult to envision
appropriate empirical tests that would provide direct access to the core computational
mechanism. On the other hand, one can capitalize on the fact that the core mechanism 
of free symmetric Merge has a clear and simple mathematical model, and try to 
derive from that model possible consequences that may be closer to empirical settings
where the model can be probed. While this is in general a broad goal, beyond what
we focus on in this paper, it motivates the approach that we follow here. Namely,
here we assume the theoretical model as it is, and we prove that its formal structure 
allows for a faithful realization in a computational setting closely related to 
neurocomputational models.

We recall the main aspects we will need of the mathematical model of the core syntax-formation 
process via free symmetric Merge, as formulated in \cite{MCB}.

As in \cite{MCB}, we denote by $\mathcal{S}\mathcal{O}_0$ the (finite) set of lexical items and syntactic features, and
by $\mathcal{S}\mathcal{O}$ the (countable) set of syntactic objects. The set $\mathcal{S}\mathcal{O}$ has an algebraic structure: it is
the free nonassociative, commutative magma generated by the set $\mathcal{S}\mathcal{O}_0$,
\[
 \mathcal{S}\mathcal{O}=\Magma_{na,c}(\mathcal{S}\mathcal{O}_0,\mathfrak{M}) 
\]
where $\mathfrak{M}$ denotes the magma operation. This implies that $\mathcal{S}\mathcal{O}$ can be identified with the
set $\mathfrak{T}_{\mathcal{S}\mathcal{O}_0}$ of abstract (nonplanar) binary rooted trees with leaves decorated by elements
of $\mathcal{S}\mathcal{O}_0$. As in \cite{MCB}, we will use the notations $\mathcal{S}\mathcal{O}$ and $\mathfrak{T}_{\mathcal{S}\mathcal{O}_0}$ interchangeably
(the notation $\mathcal{S}\mathcal{O}$ is the one that was in use in some of the linguistics literature and is suggestive
of the linguistic role, while the notation $\mathfrak{T}_{\mathcal{S}\mathcal{O}_0}$ is more directly suggestive 
of the type of mathematical objects involved, trees with leaves labelled by elements of the set $\mathcal{S}\mathcal{O}_0$). 
We will denote by $V^o(T) \subset V(T)$ the set of non-leaf vertices of a tree $T\in \mathfrak{T}_{\mathcal{S}\mathcal{O}_0}$.
We will also use the common terminology ``cherry tree'' to denote a full binary rooted tree $T\in \mathfrak{T}_{\mathcal{S}\mathcal{O}_0}$ with 
two leaves, namely $T=\mathfrak{M}(\alpha, \beta)$ and we call a ``cherry-tree substructure'', or a ``cherry subtree'' of
a full binary rooted tree $T\in \mathfrak{T}_{\mathcal{S}\mathcal{O}_0}$ a subtree $T_v \subset T$ consisting of all descendants of a 
non-root vertex $v$, that happens to be a cherry, namely $T_v=\mathfrak{M}(\alpha, \beta)$ with root $v$ and two leaves.

There is an important issue here that concerns the difference between words
and lexical items and syntactic features. While empirically assigning a wavelet representation
to words is done in neuroscience experiments, an analogous encoding for the more abstract
elements of the set $\mathcal{S}\mathcal{O}_0$ is a more delicate issue. Since we are primarily interested here
in outlining the mathematical structure of possible neurocomputational syntax encoding models, 
we will for simplicity ignore these differences at this stage. Linguistic proposals for a simplifying
setting for $\mathcal{S}\mathcal{O}_0$ can be found, for instance, in \cite{Elb}. However, even considering such
possible simplifications, there remain important differences between lexical items and
words. At the level of theoretical linguistics these would involve an interface between syntax 
and morphology (see \cite{SentMar}) and a mechanism of vocabulary insertion. How the difference
between words and lexical items can be understood in the neuroscience setting is discussed, for
instance, in Chapter 5 of \cite{Brenn2}, where a notion of ``lexical grains'' and different approaches
to modeling these (``full decomposition theory'' versus ``partial decomposition'') are compared.
For the purpose of this paper, it is not necessary to explore these notions in detail , as the
main constructions, in the generality in which we discuss them here, are not sensitive
to this level of detail.
The {\em nonassociativity} of the magma operation $\mathfrak{M}$ is what allows for the existence of 
nontrivial hierarchical structures in syntax. The important property being encoded here
is the notion of {\em constituency} and the key difference between simple concatenation
of strings of words (an associative operation) and the hierarchical structure of phrases.
Constituency in syntax refers to the presence of (nested) substructures in sentences
that are naturally grouped together and function as subunits: phrases, which are not
in themselves complete sentences, but carry completed subunits of structure. Thus,
for example it is usually argued that verb-object is such a unit (because for example
in many languages verb-object combinations can give rise to indecomposable idioms),
while the subject completes this subunit to a further structure, 
$\{ \text{ subject }, \{ \text{ verb }, \text{ object } \} \}$. There is extensive empirical
cross-linguistic evidence for such constituencies. Moreover, 
very simple examples can show that the same string of words can represent different sentences
with different underlying tree structure (``I saw someone with a telescope'' is a very
simple such example: the fact that this is not one but two different sentences is an instance
of the nonassociativity of the syntax-formation operation).  Lack of associativity
is crucial in order to encode the hierarchical structures of syntax, and this is again a
key reason why we focus on a realization of the $\mathfrak{M}$ operation on wavelet-like
objects that will be able to accommodate these same hierarchical structures: the
nonassociativity of the operation is necessary to the purpose of encoding the actual
structure of syntax beyond the associative concatenation of strings of words.
The {\em commutativity} of the magma operation $\mathfrak{M}$ is where this model of ``free symmetric'' 
Merge in the most recent formulation of Minimalism of \cite{ChomskyUCLA,
ChomskyGK, ChomskyElements} differs from previous models of
generative linguistics. Commutativity of $\mathfrak{M}$ means that, at the level of this core
computational structure, syntax formation is only handling hierarchical structures and is
not incorporating any word order requirements. All word order constraints are modeled 
as part of the Externalization interface and are language specific, while in the underlying
universal structure the only proximity relations are happening in terms of the tree structure.
In mathematical terms, the commutativity of $\mathfrak{M}$ means that the full binary rooted trees
in $\mathfrak{T}_{\mathcal{S}\mathcal{O}_0}$ are {\em not} planar trees. They are just abstract trees with no
assignment of planar embedding. A (language-specific) assignment of planar
embedding is part of the Externalization interface, which corresponds to the effect
of fixing by language learning those syntactic parameters that pertain to word order
properties.

In addition to the magma of syntactic objects, the model of syntax formation by the
free symmetric Merge also incorporates transformations on syntactic objects (movement)
performed by what is called in the linguistics literature Internal Merge. In the current model,
an overall Merge dynamical system combines the External Merge (which acts like the $\mathfrak{M}$ 
magma operation, merging two trees at the root) and Internal Merge, which extracts
an {\em accessible term} (a subtree $T_v$ made of all descendants of a non-root vertex $v$ of
a syntactic object) and merges it back at the root. To properly formulate this dynamics,
Chomsky introduced the notion of {\em workspaces} (in mathematical terms, forests
$F=T_1\sqcup \cdots \sqcup T_r$ with components $T_i \in \mathfrak{T}_{\mathcal{S}\mathcal{O}_0}$). In the
mathematical formulation of \cite{MCB}, the set $\mathfrak{F}_{\mathcal{S}\mathcal{O}_0}$ of workspaces spans
a Hopf algebra, where the coproduct performs the extraction of accessible terms needed
for movement via Internal Merge. The Merge dynamical system is then implemented by
operators of the form
\begin{equation}\label{MSS}
	\mathfrak{M}_{S,S'}=\sqcup \circ (\mathcal{B} \otimes \Id) \circ  \delta_{S,S'}  \circ \Delta 
\end{equation} 
where the coproduct $\Delta$ performs all the possible extractions of
(forests of) accessible terms, with
\begin{equation}\label{coprodMerge}
	\Delta(T)= T\otimes 1 + 1 \otimes T + 
	\sum (T_{v_1}\sqcup \cdots \sqcup T_{v_r})\otimes T/(T_{v_1}\sqcup \cdots \sqcup T_{v_r})\, ,
\end{equation} 
the operation $\delta_{S,S'}$ searches and selects an extracted accessible term of the
form $S\sqcup S'$, to which the merging $\mathcal{B}: S\sqcup S' \mapsto \mathfrak{M}(S,S')$ is applied.
The new resulting workspace is then re-assembled using the product operation $\sqcup$.
The operations $\mathfrak{M}_{S,S'}$ can be combined in a single operation 
\begin{equation}\label{MergeK}
	\mathcal{K}=\sum_{S,S'} \mathfrak{M}_{S,S'}= \sqcup \circ (\mathcal{B} \otimes \Id) \circ \Pi_{(2)} \circ \Delta\, , 
\end{equation} 
which, together with its iterates, describes the core dynamical system of syntax formation.
In the form \eqref{MergeK}, 
the projection $\Pi_{(2)}$ selects those terms of the coproduct with $T_{v_1} \sqcup T_{v_2}$
on the left-hand-side  (including the case of a single term, with either $T_{v_1}$ or $T_{v_2}$
the formal empty tree, the unit of the Hopf algebra), and $\sqcup$ is the Hopf algebra product
that re-assembles the resulting workspace. The resulting dynamical system describing the
action of Merge on workspaces is a Hopf algebra Markov chain in the sense of \cite{DPR}, and
is discussed in detail in \cite{MarSki}.

In the linguistics literature the operation $\delta_{S,S'}$ of \eqref{MSS} that identifies,
in a given workspace, accessible terms to be used by Merge, is associated to an
optimal search algorithm, called Minimal Search. An extensive discussion, within
the mathematical model of \cite{MCB}, of optimization aspects such as Minimal Search,
and other types of cost functions for Merge operations, is given in \cite{MarLarHuij} and \cite{MarSki}.  
We will not discuss cost functions in this paper, but we will give an interpretation of
Minimal Search (as efficient location and extraction of accessible terms) in \S \ref{AccTermsSec}
and formulate more precisely the question of implementation in our model. 

We will focus here mostly on the realization of the magma of syntactic objects, as that
is our main goal in this paper, but we will also discuss workspaces and the action of
Merge in \S \ref{IMsec}. We will return in forthcoming work to discuss other possible 
neurocomputational realizations of the Hopf algebra Markov chain dynamical system
of \cite{MarSki}.
There is one more aspect of the linguistic model that we need to recall here, as we will be using
it later in the paper, namely the notion of syntactic head, which is part of the requirements of a 
well formed syntactic structure, and its abstract mathematical formulation 
as ``head function''. Indeed a syntactic head is needed for a well formed organization
of constituency-substructures in syntactic objects, especially those substructures called phases,
that are already sufficiently complete to be interpretable at the syntax-semantics interface, since
it governs the organization of specifier-head-complement constituencies, and the completed
substructure determined by the ``maximal projection'' of the head from leaf towards the interior of
the syntactic tree. It also provides a labeling algorithm for the internal vertices of the trees, given
the lexical items and features at the leaves. 
The main properties of the syntactic head are described in the linguistic formulation of 
\cite{ChomskyBare} and reformulated mathematically as head functions in \S 1.13 of \cite{MCB},
with the discussion of phases and labeling algorithm in \S 1.14 and \S 1.15, respectively. 
It is shown in \cite{MCB} that the mathematical definition is equivalent to the definition of 
head given in \cite{ChomskyBare}. Thus, for our purposes, it suffices to recall this mathematical
definition.

As in \cite{MCB}, a head function $h_T$ on a syntactic object $T\in \mathfrak{T}_{\mathcal{S}\mathcal{O}_0}$ is a function $h_T: V^o(T) \to L(T)$
from the non-leaf vertices of $T$ to the leaves, with the property that $h_T(v)\in L(T_v)$, where $T_v\subset T$ is the accessible
term at $v$ (the full subtree rooted at $v$) and that if $T_w \subset T_v$ and $h_T(v)\in L(T_w)$ then $h_T(v)=h_T(w)$.
Assigning a head function $h_T$
to $T$ is the same thing as assigning at each vertex $v\in V^o(T)$ a marking to one of the two edges below $v$ (the direction 
pointing to the leaf $h_T(v)$). In particular, in a pair $(T,h_T(v))$ there is an assigned way of selecting one of the two
edges below each vertex. We write $h: T \mapsto h_T$ for an assignment of a head function and $\Dom(h) \subset \mathcal{S}\mathcal{O}$
for the set of syntactic objects that have a head function. As discussed in \cite{MCB} and \cite{MHL}, $\Dom(h)$
is not a submagma: incorporating head functions leads to extending the magma structure to a hypermagma. On
the other hand, the coproduct structure is compatible with the presence of head functions.

The linguistics model of
phases can be found in \cite{ChomskyBare, Chomsky-phase}   
and the mathematical formulation is given in \cite{MHL}. 
In this paper we do not try to model this more refined structure of phases, and we will make 
very limited use of head functions, so we will not go into further detail here, as we have
recalled all that we will need.  We will discuss in \S \ref{HeadSec} how the
head function can be modeled, in a particular form of the model we discuss in this paper.
Head functions will also play an auxiliary role in \S \ref{PolyTree}, although the result of the construction
described there will be independent of these additional data. 

\subsection{Binary trees in spaces of functions} 

We first analyze the magma $\mathcal{S}\mathcal{O}$ of syntactic objects recalled above, and how it can 
be encoded in a space of functions.

Let $\mathcal{F}$ denote a function space. We want to be fairly broad for now in terms of the properties of
this space, but it would typically be an infinite dimensional topological vector space (or Banach space,
or Hilbert space) of functions on some underlying topological space $X$. We think of the space $X$
as representing areas of the brain as well as an interval of time over which observations of brain
activity are performed. We can assume that $X$ is a compact space. This will include cases where $\mathcal{F}$
is the space of continuous functions $\mathcal{C}(X,\mathbb{R})$, or of smooth functions $\mathcal{C}^\infty(X,\mathbb{R})$, or
integrable or square-integrable functions, $L^1(X,\mathbb{R})$ or $L^2(X,\mathbb{R})$. We can also allow generalizations 
that include spaces of distributions, rather than functions. In practice, while what we say in this section 
will hold at this level of generality, we can for simplicity think of the  
case of continuous functions on an interval $X=[0,1]$, which 
suffices to describe functions of time, over a certain window of observation (though in general
we would want, more realistically to include in $X$ also positional coordinates identifying
brain locations).

Of particular interest to us are spaces of functions generated by a wavelet basis, as wavelet analysis is
ordinarily employed for signal analysis and denoising, by decomposing a signal in a wavelet basis.
For example, in wavelet analysis of magnetoencephalography (MEG) data (see \cite{Dash}) a
wavelet decomposition using discrete wavelet transform provides components corresponding to the 
high-gamma (62--125 Hz), gamma (31--58 Hz), beta (16--30 Hz), alpha (8--16 Hz), and theta (4--8 Hz) 
frequency bands of the neural signal, with the higher frequency components removed for denoising. 
In \cite{Dash}, the spectral-temporal characteristics of the signal are encoded in a representation, 
obtained via continuous wavelet transform, of the denoised signal in a basis of Morlet wavelets.

Let then $\mathcal{F}$ be a linear space of functions endowed with a chosen basis as above.
Suppose given an injective function $\varphi: \mathcal{S}\mathcal{O}_0 \to \mathcal{F}$. This would represent an encoding
of lexical items (together with syntactic features) as certain linear combinations in that basis (say,
combinations of wavelets). Such a map can be constructed empirically by methods such as 
those discussed in \cite{Dash, WBPLLA}. We will just assume here that it is a given input 
of our theoretical model.

As discussed in \S \ref{revMergeSec} above, for the purpose of this paper we will
not discuss the difference between words and lexical items in relation to these kinds of
wavelet representation, and we will only take the existence of a representation
$\varphi: \mathcal{S}\mathcal{O}_0 \to \mathcal{F}$ as a given input for our construction.

Our main working assumptions about the function $\varphi: \mathcal{S}\mathcal{O}_0 \to \mathcal{F}$ are summarized as follows.

\begin{remark}\label{SO0rem}
		We assume that the map
		$\varphi: \mathcal{S}\mathcal{O}_0 \to \mathcal{F}$ is injective, namely that its image can disambiguate any
		different items in $\mathcal{S}\mathcal{O}_0$. We also assume the stronger property that, with respect to the 
		vector space structure of $\mathcal{F}$, all the $\varphi(\alpha)$, for $\alpha\in \mathcal{S}\mathcal{O}_0$, are 
		linearly independent vectors. 
\end{remark}

The linear independence assumption for representations of lexical items makes
the embedding $\varphi: \mathcal{S}\mathcal{O}_0 \to \mathcal{F}$ more akin to the type of ``one-hot''
embeddings representing words in large, sparse, high-dimensional spaces,
rather than to the kind of ``word embeddings'' where linear relations account for
certain semantic similarities. This is in part motivated by adherence to the 
modularity of the theoretical linguistics model we are considering, where 
the interface with semantics is processed separately, and the linear independence
assumption reflects this aspect of the model. There is also a technical reason for
the linear independence assumption, which will be clear shortly, since a first
step of our construction uses convex linear combinations, and the faithfulness
we seek in representing syntactic objects would fail if we start forming linear
combinations of vectors that already have linear dependence relations.

The question we consider here is how to extend the map  $\varphi: \mathcal{S}\mathcal{O}_0 \to \mathcal{F}$ to a
map $\varphi: \mathcal{S}\mathcal{O} \to \mathcal{F}$ that is still an embedding, and that also encodes the magma
structure of $\mathcal{S}\mathcal{O}$. We also want this map to be in some sense a solution to an
optimization problem in the computational and information theoretic sense.  The idea
that the Merge operation in syntax should be characterized by optimization properties
in terms of efficiency of computation (often referred to in the literature as ``third factor'' 
constraints) has been a key idea in the development of Minimalism,
especially in its current form. The mathematical model of Minimalism of \cite{MCB}
formulates some of these notions of optimality in a more directly quantifiable way,
which in turn can be used for deriving linguistic consequences, as in \cite{MarLarHuij}.
Here we will identify another possible form of optimality, in the specific realization
that we propose in function spaces, namely the fact that the nonassociativity
property of Merge is obtained through an entropy minimization, and the particular
choice of entropy functional used for minimization (the second R\'enyi entropy)
has itself optimality properties, as we discuss in \S \ref{2ndRenyiSec}.

The key parts of this problem can be rephrased as the following questions.

\begin{question}\label{SOques}
		\begin{enumerate}
			\item Given an abstract binary rooted tree $T\in \mathfrak{T}_{\mathcal{S}\mathcal{O}_0}$ with $\{ \alpha_\ell \}_{\ell\in L(T)}$ the
			elements $\alpha_\ell \in \mathcal{S}\mathcal{O}_0$ assigned to the leaves $\ell\in L(T)$, construct a (not necessarily linear) 
			combination of the functions $\varphi(\alpha_\ell)\in \mathcal{F}$ that faithfully encodes the combinatorial tree structure of $T$.
			\item Show that this construction is compatible with the magma structure, in the sense that,
			if $\varphi(T)$ and $\varphi(T')$ are obtained as above, then $\varphi(\mathfrak{M}(T,T')) = \Gamma(\varphi(T),\varphi(T'))$
			for a commutative nonassociative binary operation $\Gamma$ on $\mathcal{F}$.
			\item Show that the function computing $\varphi(T)$ from the $\varphi(\alpha_\ell)$ is optimal
			with respect to an entropy/information functional. 
	\end{enumerate}
\end{question}

We will answer these three questions in steps. We first show in \S \ref{PolyTree} that it is possible to associate to
a syntactic object $T\in \mathfrak{T}_{\mathcal{S}\mathcal{O}_0}$ with a head function (syntactic head) a probability distribution $A=(a_\ell)_{\ell\in L(T)}
$ on the set of leaves $L(T)$, so that $a_\ell=P_\ell(\Lambda)$ is a polynomial function of variables $\Lambda=(\lambda_v)$
with $\lambda_v\in [0,1]$ associated to each non-leaf vertex $v\in V^o(T)$. We show that the function $A=P(\Lambda)$
completely determines the pair $(T,h_T)$ of the syntactic object and the head function. This construction gives an
embedding of syntactic objects into a much larger space of functions than the one where the lexical items are embedded,
as the function associated to $T$ depends on a collection $\Lambda=(\lambda_v)$ of auxiliary variables, in addition
to the variables in $X$ that the functions $\varphi(\alpha_\ell)$ depend on. This is not yet satisfactory because one wants
both lexical items and syntactic objects to all live in the same function space (without a rapidly growing set of
additional variables) and, moreover, one wants the images of syntactic objects in this vector space to still carry
the algebraic operation that gives the magma structure and that is the base of the structure formation via Merge.
To move towards that goal, we then observe in \S \ref{CombOp} that the operation
$\Gamma(\varphi(T), \varphi(T')) = \lambda \varphi(T) + (1-\lambda) \varphi(T')$ that combines two
functions in a convex combination has some of the algebraic properties we want for Merge (nonassociativity)
but not others (commutativity). This suggests that a modification of this kind of operation that
achieves commutativity but maintains nonassociativity, and that also identifies an optimal choice
of the value of $\lambda$, can provide the next step of the construction. We do this in \S \ref{RenyiSec}
by endowing the space
of functions $\mathcal{F}$ with a nonassociative commutative addition that is obtained as a deformation of the
$(\min, +)$ semiring (tropical semiring) with a deformation parameter that behaves like a thermodynamical
inverse temperature and where the convex combination 
$\Gamma(\varphi(T), \varphi(T')) = \lambda \varphi(T) + (1-\lambda) \varphi(T')$ is corrected
by adding a term $-\beta^{-1} S(\lambda, 1-\lambda)$, where $S$ is an information measure.
The resulting $\Gamma(\varphi(T), \varphi(T')) -\beta^{-1} S(\lambda, 1-\lambda)$ is then
minimized over $\lambda$: for suitable choices of $S$ this operation is indeed nonassociative
but commutative. The question then is whether with this operation one can obtain embeddings
of syntactic objects in $\mathcal{F}$. This question is answered in \S \ref{WallSec} and \S \ref{TransvSec},
with some further discussion of the remaining cases in \S \ref{RepCopSec}. The compatibility
with the magma operation is discussed in \S \ref{magmaSec} and the implementation of the
Hopf algebra structure and the Merge operation in \S \ref{IMsec}.

\subsection{Polynomial functions from trees}\label{PolyTree}

We assume here that we have a head function $h: T \mapsto h_T$ as recalled above,
with $\Dom(h)\subset \mathcal{S}\mathcal{O}$. 
As we will see below, this choice is only an auxiliary part of the construction we describe and 
does not matter in the end in our argument, since the dependence on the head function will be eliminated
at a later step of our construction. In other words, the realization of Merge that we propose does not in itself depend
on the use of the head function. (Indeed this has to be the case, as head functions are not compatible with the magma structure, as
discussed in \S 1.13 of \cite{MCB} and in \cite{MHL}.)

We first extend the map $\varphi: \mathcal{S}\mathcal{O}_0 \to \mathcal{F}$ to a map $\varphi: \Dom(h) \subset \mathcal{S}\mathcal{O} \to \mathcal{F}\otimes_\mathbb{R} \mathbb{R}[\underline{\Lambda}]$ with values in the tensor product of the vector space $\mathcal{F}$ with polynomials in a countable 
set of variables $\underline{\Lambda}$.

Note that this kind of extension, in itself, does not answer the main question we are asking in this paper. In fact, we
are looking for an extension of $\varphi: \mathcal{S}\mathcal{O}_0 \to \mathcal{F}$ from lexical items to syntactic objects that
maintains  values in {\em the same} function space where
the lexical items are mapped, while here we extend $\mathcal{F}$ by a countable set of arbitrary new variables, whose
nature is not directly justifiable in terms of a given description of wavelet-like signals associated to lexical items.
However, this first extension $\varphi: \Dom(h) \subset \mathcal{S}\mathcal{O} \to \mathcal{F}\otimes_\mathbb{R} \mathbb{R}[\underline{\Lambda}]$
serves as an intermediate step towards obtaining the kind of extension we want, in which the additional
variables in $\underline{\Lambda}$ will be appropriately optimized over, leaving resulting functions in $\mathcal{F}$.
As we will see, that same optimization process will also simultaneously implement the correct algebraic properties
of the magma operation $\mathfrak{M}$. 

Some comments on notation: as in \cite{MCB}, we write here and in the following $\ell \in L(T)$ for the leaves of a
tree $T\in \mathfrak{T}_{\mathcal{S}\mathcal{O}_0}$, and we write $\alpha_\ell \in \mathcal{S}\mathcal{O}_0$ for the lexical items and
syntactic features associated to the leaves $\ell \in L(T)$. This notation allows for the fact
that there may be repetitions in the assignments of items at the leaves, so that one may
have $\alpha_\ell = \alpha_{\ell'} \in \mathcal{S}\mathcal{O}_0$ for some $\ell\neq \ell' \in L(T)$. This notation
is consistent with the definition of the set of trees $\mathfrak{T}_{\mathcal{S}\mathcal{O}_0}$ as the free
nonassociative commutative magma generated by the set $\mathcal{S}\mathcal{O}_0$ (the magma contains,
for example, trees $T$ where the same element $\alpha\in \mathcal{S}\mathcal{O}_0$ is assigned to all the leaves). 
We will mention later (especially with respect to the faithfulness argument) when it is convenient to 
assume that there are no repetitions among the items attached at the leaves, and how to extend
the argument to the case where repetitions are present.

\begin{proposition}\label{phiTprop}
	Given an injective map $\varphi: \mathcal{S}\mathcal{O}_0 \to \mathcal{F}$ whose image consists of a collection of linearly independent vectors in $\mathcal{F}$, 
	and a syntactic object $T\in \mathcal{S}\mathcal{O}=\mathfrak{T}_{\mathcal{S}\mathcal{O}_0}$
	that is in $\Dom(h)$,
	let $V^o(T)$ denote the set of non-leaf vertices of $T$. Let $\varphi(\alpha_\ell)$ be the
	functions associated to the elements $\alpha_\ell \in \mathcal{S}\mathcal{O}_0$ at the leaves $\ell \in L(T)$. These
	determine a function $\varphi(T)\in \mathcal{F}[\{ \lambda_v \}_{v\in V^o(T)}]$, the vector space of polynomials
	in a set of variables $\Lambda=\{ \lambda_v \}_{v\in V^o(T)}$ associated to the non-leaf vertices,
	\begin{equation}\label{phiT}
		\varphi(T)(\Lambda)=\sum_{\ell \in L(T)} a^T_\ell(\Lambda) \varphi(\alpha_\ell) \, ,
	\end{equation}
	where the $a^T_\ell(\Lambda)$ are polynomials in the variables $\lambda_v$. The polynomials
	$a^T_\ell(\Lambda)$ are determined recursively via the following relation.
	Let $T=\mathfrak{M}(T_1,T_2)\in \Dom(h)$ with root vertex $v_T$ and with $h_T(v_T)=h_{T_1}(v_{T_1})$, where $v_{T_1}$ 
	is the root vertex of $T_1$. 
	Then  
	\begin{equation}\label{aT}
		a^T_{\ell}= \left\{ \begin{array}{ll} \lambda_{v_T} \, a^{T_1}_\ell & \text{ for } \ell\in L(T_1)\subset L(T) \, , \\
			(1-\lambda_{v_T}) \, a^{T_2}_\ell & \text{ for } \ell\in L(T_2)\subset L(T) \, ,
		\end{array} \right. 
	\end{equation}
	where the recursion starts at $T=\ell$ a single leaf labelled by $\alpha_\ell$, for which we take $a^{\ell}_\ell=1$.
	The case of $h_T(v_T)=h_{T_2}(v_{T_2})$ is analogous with $\lambda_v$ replaced by $1-\lambda_v$.
\end{proposition}

\begin{proof} 
The construction of $\varphi(T)$ follows the construction of $T$ in the magma $\mathcal{S}\mathcal{O}$,
starting with the elements of $\mathcal{S}\mathcal{O}_0$. Given a pair $\alpha,\beta \in \mathcal{S}\mathcal{O}_0$ and a cherry tree substructure of $T$ of the form
\[
 T_v =\{ \alpha, \beta \} = \Tree[ $\alpha$ $\beta$ ] \, , 
 \]
with root vertex $v$, the functions $\varphi(\alpha)$ and $\varphi(\beta)$, and a variable $\lambda_v$,
form the combination
\[
 \varphi(T)=\lambda  \varphi(\alpha) + (1-\lambda) \varphi(\beta) \, , 
\]
if $\alpha$ is the head of $T$ (and with $\lambda$ and $1-\lambda$ exchanged otherwise). 
When further applying the magma operation $\mathfrak{M}$ we assign to
\[
 \mathfrak{M}(T,\gamma)=\{ \{ \alpha,\beta \}, \gamma \} = \Tree[ [ $\alpha$ $\beta$ ] $\gamma$ ] 
\]
(where the trees are nonplanar) the combination
\[ 
\lambda' \lambda  \varphi(\alpha) + \lambda'  (1-\lambda) \varphi(\beta) + (1-\lambda') \varphi(\gamma) \, , 
\]
where $\lambda, \lambda'$ are two parameters. The combination is depicted here in the case where $\alpha$ is the
head of the resulting tree. We see that this procedure
can be continued, resulting in an expression $\varphi(T)$ that depends on a collection of parameters $\{ \lambda_v \}_{v\in V^o(T)}$,
with $V^o(T)$ the set of non-leaf vertices of $T$. This can be seen inductively as phrased in the statement.
If we have constructed $\varphi(T)(\Lambda)$ and $\varphi(T')(\Lambda')$, where $\Lambda=(\lambda_v)_{v\in V^o(T)}$ and
$\Lambda'=(\lambda'_w)_{w\in V^o(T')}$, for two given syntactic objects $T,T'\in \mathcal{S}\mathcal{O}$,
then we define 
\[
 \varphi(\mathfrak{M}(T,T'))(\Lambda, \Lambda', \lambda_u)= \lambda_u \varphi(T)(\Lambda) + (1-\lambda_u) \varphi(T')(\Lambda')\, , 
 \]
where $\lambda_u$ is the additional parameter associated to the new root vertex of $\mathfrak{M}(T,T')$, and we assume that the head of $\mathfrak{M}(T,T')$
is the head of $T$ (if it is the head of $T'$ we replace $\lambda_u\leftrightarrow 1-\lambda_u$).
The result of this construction can be seen as a function
\[ 
\varphi(T)=\sum_{\ell \in L(T)} a_\ell \,  \varphi(\alpha_\ell) \, , 
\]
where $\alpha_\ell \in \mathcal{S}\mathcal{O}_0$ are the items associated to the leaves $\ell\in L(T)$, with $\varphi(\alpha_\ell) \in \mathcal{F}$,
where the coefficients $a_\ell$ are polynomial functions of the variables $\lambda_v$, for $v\in V^o(T)$,
\[
 a_\ell = P_\ell(\lambda_v) \in \mathbb{Z}[\{ \lambda_v \}_{v\in V^o(T)} ] \subset \mathbb{R}[\{ \lambda_v \}_{v\in V^o(T)} ]  
 \]
as stated. Since the parameters $\lambda_v$ are regarded as variables, and not evaluated here at a
particular point of the unit interval, the resulting $\varphi(T)$ is an element of the vector space of 
polynomials in the variables $\lambda_v$, with coefficients in the vector space $\mathcal{F}$, namely, 
in $\mathcal{F}\otimes \mathbb{R}[\{ \lambda_v \}_{v\in V^o(T)} ]$ (which is a vector space not a ring, unless $\mathcal{F}$ is also an algebra and
not just a vector space). 
\end{proof}

The presence of the head function $h_T$ on $T$ is simply used here to decide which of the edges below a vertex $v$
contributes a factor $\lambda_v$ (marked edge in the head direction) or $1-\lambda_v$ (the other edge). More precisely,
this can be phrased as the following observation.

\begin{remark}\label{remDomh}
		The construction of $\varphi(T)$ follows the generative process of the magma $\mathcal{S}\mathcal{O}$, since the
		recursive construction uses the magma operation $T=\mathfrak{M}(T_1,T_2)$. However, in the assignment
		\[
		 \varphi(T)=\lambda  \varphi(T_1) + (1-\lambda) \varphi(T_2) \, , 
		 \]
		the presence of a head function on $T$ is used to decide which of the two terms gets the variable $\lambda$
		and which $1-\lambda$. The domain $\Dom(h)$ is not a submagma of $\mathcal{S}\mathcal{O}$: in  general 
		$\mathfrak{M}(T_1,T_2)$ is not necessarily in $\Dom(h)$ even though $T_1$ and $T_2$ are. Thus, in
		Proposition~\ref{phiTprop} we need to assume $T\in \Dom(h)$. 
\end{remark}

\begin{corollary}\label{aellTpoly}
	A head function $h_T$ on a syntactic object $T$ determines a partition of the set of vertices $V(T)$
	into a disjoint union of paths $\gamma_\ell$ for $\ell \in L(T)$, whose edges are the marked edges
	determined by the head function.
	The recursive form \eqref{aT} gives the explicit form of the polynomial dependence $A=P(\Lambda)$ as
	\begin{equation}\label{aTell}
		a_\ell^T =\prod_{v\in \gamma_{v_0,\ell}} \epsilon_v \, , \ \ \text{ with } \epsilon_v= \left\{ \begin{array}{ll} \lambda_v & e_v\in \gamma_{h_T(v)} \\
			1-\lambda_v & e_v\notin \gamma_{h_T(v)}  \, ,
		\end{array} \right. 
	\end{equation}
	with $v_0$ the root of $T$, where $\gamma_{v_0,\ell}$ is the path from $v_0$ to the leaf $\ell$ and $e_v$ denotes 
	the edge of the path $\gamma_{v_0,\ell}$ with source vertex $v$ (in the orientation away from the root). 
\end{corollary}

\begin{proof} As recalled above, the head function marks an edge below each vertex. The
paths $\gamma_\ell$ can be obtained by following for each $\ell \in L(T)$ the path along these
marked edges.  Some of the paths $\gamma_\ell$ are trivial and consist of a single leaf and no edges.
For each nontrivial path, the highest vertex $v_\ell$ on the path is the maximal projection of the
head $\ell$, see \cite{MCB} for a discussion of all these properties of the head function.
By  \eqref{aT} , at each step along the path $\gamma_{v_0,\ell}$ from the root $v_0$ to the
leaf $\ell$, we multiply by either a factor $\lambda_v$ if the edge is marked or $1-\lambda_v$ if it is not,
that is, according to whether the edge $e_v$ is on 
the path $\gamma_{h_T(v)}$ (the path from $v$ to the leaf $h_T(v)$)  or not, resulting in the
product \eqref{aTell}.
\end{proof}

\subsection{Reconstructing the tree}\label{TreePoly}

We show the injectivity of the map constructed in Proposition~\ref{phiTprop}, namely the fact that the syntactic
object $T\in \mathfrak{T}_{\mathcal{S}\mathcal{O}_0}$ can be reconstructed from its image 
$\varphi(T)$ in $\mathcal{F}\otimes \mathbb{R}[\Lambda]$.
To do this, we consider a convenient parameterizing space for metric trees (trees with weights $w_e > 0$
on the edges) and we think of the weights $\lambda_v$ and $1-\lambda_v$ assigned to the pair of edges 
below a non-leaf vertex $v$ as coordinates in this space. A construction of such parameterizing space for 
metric trees is done in \cite{BHV}  and they are usually referred to as BHV moduli spaces. We do not need
the full details from the construction of these spaces, but we do use one key idea. For full binary trees 
$T$ with $n$ labelled leaves, the data $( w_e )_{e\in E(T)}$ form an open cell homeomorphic to a 
hypercube of dimension the number of edges. These $(2n-3)!!$ open cells (one for each tree topology) 
are then glued together to form the compactified moduli space along faces that correspond to the limit 
$w_e \to 0$ for one of the edge lengths, so that the boundary strata can be thought of as parameterizing 
metric trees where some vertices are of higher valence. Crossing a wall corresponds to shrinking to zero
one of the two edges below one of the vertices and re-expanding it to a nonzero length on the other side
of the vertex, thus changing the tree topology. Any two different tree topologies are obtainable from one
another via a sequence of such wall crossing operations. This is the main property we will use here.

\begin{proposition}\label{Trec}
	One can reconstruct the tree $T$ and the head function $h_T$ from the function $\varphi(T)$
	of the variables $\lambda_v$, under the assumption that the $\varphi(\alpha)$, for $\alpha\in \mathcal{S}\mathcal{O}_0$, 
	are linearly independent. 
\end{proposition}

\begin{proof}
For a fixed set of leaves $L$ and distinct items $\alpha_\ell \in \mathcal{S}\mathcal{O}_0$, for $\ell\in L$,
associated to the leaves, we need to show that, if $T \neq T'$ are two nonisomorphic trees with $L(T)=L(T')=L$,
then for some $\ell\in L$ the polynomials $a_\ell(T)$ and $a_\ell(T')$ are also different, $a_\ell(T)\neq  a_\ell(T')$,
so that $\varphi(T)\neq \varphi(T')$. 
First observe that we can associate to the pair $(T,\Lambda)$ with $\Lambda=(\lambda_v)_{v\in V^o(T)}$ a
point in the BHV moduli space (see \cite{BHV} and \S 3.4.2 of \cite{MCB}), by endowing each edge $e$ of $T$
with a weight $w_e$ equal to either $\lambda_v$ or $1-\lambda_v$, depending on whether in the
function $\varphi(T)$ the polynomials $a^T_\ell$ of all the leaves $\ell$ in the subtree below the edge $e$  
contain a factor $\lambda_v$ or $1-\lambda_v$. In the BHV moduli
space a nonplanar binary rooted tree $T$ with assigned weights $w_e > 0$ at the edges corresponds to a point
inside one of the open cells of the moduli space. The walls between the open cells correspond to degenerate
trees with higher valence vertices, where one (or more in the deeper strata) of the edge weights take the
limiting value $w_e=0$. Thus, passing from one to another nonisomorphic tree with the same labelled leaves
is achieved by a number of wall crossings in the BHV moduli space. Each such wall crossing corresponds to
one edge $e$ with source $v$ moving across the vertex $w$ above its source $v$, by contracting to zero length the
edge between $v$ and $w$ and creating a nonzero length edge on the other branch below $w$. With $v'$
denoting the other vertex below $w$, and $T_v =\mathfrak{M}(T_{v,1}, T_{v,2})$, each such crossing corresponds to a change 
\[ 
T_w = \Tree[ $T_v$ $T_{v'}$ ] = \Tree[ [ $T_{v,1}$ $T_{v,2}$ ] $T_{v'}$ ] \mapsto \Tree[ $T_{v,1}$ [ $T_{v,2}$ $T_{v'}$ ]] 
\]
which corresponds to a change 
\begin{multline*}
 \lambda' \lambda  \varphi(T_{v,1}) + \lambda'  (1-\lambda) \varphi(T_{v,2}) + (1-\lambda') \varphi(T_{v'}) \mapsto \\
  \lambda'   \varphi(T_{v,1}) + (1-\lambda')  \lambda \varphi(T_{v,2}) + (1-\lambda') (1-\lambda) \varphi(T_{v'}) 
\end{multline*}
in the associated function. 

The functions $\varphi(T_{v,1})$, $\varphi(T_{v,2})$, $\varphi(T_{v'})$ are linearly independent in $\mathcal{F}$ (for any fixed
values of the parameters $\lambda_v$) as linear combinations of different sets of $\alpha_\ell$ (since the sets
$L(T_{v,1})$, $L(T_{v,2})$ and $L(T_{v'})$ do not overlap). Thus, by the same argument given above, the two
expressions
\begin{align*}
 \lambda' \lambda  \varphi(T_{v,1}) + \lambda'  (1-\lambda) \varphi(T_{v,2}) + (1-\lambda') \varphi(T_{v'}) \\
  \lambda'   \varphi(T_{v,1}) + (1-\lambda')  \lambda \varphi(T_{v,2}) + (1-\lambda') (1-\lambda) \varphi(T_{v'}) 
  \end{align*}
are different for any value $\lambda, \lambda'\notin \{ 0, 1\}$. 
Since each successive wall crossing will involve similar transformations at different internal vertices $w$,
affecting different subsets of leaves, each time we obtain different resulting functions, so that the
different trees $T$ with same $L=L(T)$ will give rise to different functions $\varphi(T)$.
For trees in $\mathcal{S}\mathcal{O}$ that do not have the same set of leaves the resulting functions are necessarily
different because they either depend on a different set of variables, if the sets of leaves do not have
the same cardinality, or have different elements in $\mathcal{S}\mathcal{O}_0$ at the leaves hence give combinations
of different sets of independent vectors.  This means that, in all cases, the functions $\varphi(T)$ suffice
to reconstruct the syntactic objects $T\in \mathcal{S}\mathcal{O}$. Corollary~\ref{aellTpoly} shows that the head function
$h_T$ is also encoded in $\varphi(T)$: indeed, assigning $h_T$ is the same thing as assigning which
edge below each vertex is marked, and this can be obtained from the formula \eqref{aTell}, by checking
which factor is a $\lambda_v$ or a $1-\lambda_v$.
\end{proof}

Combining Proposition~\ref{phiTprop} and Proposition~\ref{Trec} we obtain the following result.

\begin{corollary}\label{InfPolys}
	Let $\underline{\Lambda}=\cup_{T\in \mathfrak{T}} \{ \lambda_{T,v} \,|\, v\in V^o(T) \}$ be a countable set of
	variables with $T\in \mathfrak{T}$ ranging over the countable set of all nonplanar binary rooted trees. The
	construction of Proposition~\ref{phiTprop} extends an injective map $\varphi: \mathcal{S}\mathcal{O}_0 \to \mathcal{F}$, whose image 
	consists of a collection of linearly independent vectors in $\mathcal{F}$, to an injective map 
	\begin{equation}\label{phiLambdaInf}
		\varphi: \Dom(h) \subset \mathfrak{T}_{\mathcal{S}\mathcal{O}_0} \to \mathcal{F}\otimes_\mathbb{R} \mathbb{R}[\underline{\Lambda}]  \, . 
	\end{equation}
\end{corollary}

\subsection{Combination operations}\label{CombOp}

For any evaluation of the parameters $\lambda_v$ at particular points $\lambda_v\in (0,1)$, we obtain a
function $\varphi(T)\in \mathcal{F}$. In general we do not expect that fixing the values of the $\lambda_v$
variables will still yield an injective map of the syntactic objects to $\mathcal{F}$. Thus, the question, in terms
of these evaluations of $\varphi(T)(\Lambda)$ is whether there is an optimal way of choosing the
values of the $\lambda_v$ variables. We discuss in \S  \ref{RenyiSec} a way to pass from 
functions $\varphi(T)\in \mathcal{F}\otimes \mathbb{R}[\underline{\Lambda}]$ to functions in $\mathcal{F}$ that is motivated by
the algebraic properties of the iterated convex combinations involved in  the construction of $\varphi(T)$.
We first observe the following property. 

\begin{lemma}\label{nonass}
	For given $\lambda, \lambda'\in (0,1)$, the operation
	\begin{equation}\label{GammaT}
		\Gamma(\varphi(T), \varphi(T')) = \lambda \varphi(T) + (1-\lambda) \varphi(T') 
	\end{equation} 
	is nonassociative. It is also non-commutative, except when  $\lambda = \lambda' =1/2$. 
\end{lemma}

\begin{proof}
The nonassociativity is seen by the fact that, since the $\varphi(\alpha)$, for $\alpha\in \mathcal{S}\mathcal{O}_0$, are linearly independent,
setting
\[
 \lambda' \lambda  \varphi(\alpha) + \lambda'  (1-\lambda) \varphi(\beta) + (1-\lambda') \varphi(\gamma) =
\lambda'   \varphi(\alpha) + (1-\lambda')  \lambda \varphi(\beta) + (1-\lambda') (1-\lambda) \varphi(\gamma) 
\]
would imply $\lambda' \lambda=\lambda'$ and $\lambda'  (1-\lambda)=(1-\lambda')  \lambda$ and
$1-\lambda' =(1-\lambda') (1-\lambda)$, which would give $\lambda, \lambda' \in \{ 0, 1\}$ which are excluded values. 
Similarly, by linear independence, we see that $\lambda \varphi(T) + (1-\lambda) \varphi(T') =\lambda \varphi(T') + (1-\lambda) \varphi(T)$
iff $\lambda=1-\lambda$  and $\lambda'=1-\lambda'$, i.e. $\lambda=\lambda'=1/2$. 
In other words, the lack of commutativity is
measured by the transformation that exchanges $\lambda \leftrightarrow (1-\lambda)$.
\end{proof}

We want to improve on this representation, by obtaining a faithful mapping of
syntactic objects to a space of functions, so that the resulting functions $\varphi(T)$
are obtained (not necessarily as linear combinations) from the functions $\varphi(\alpha_\ell)$ 
associated to the lexical items at the leaves, in such a way that the mapping preserves 
the structure of {\em commutative, nonassociative magma}.  In order to do this, 
we use the formalism of thermodynamic semirings introduced in \cite{MarThor}.

\section{Thermodynamic Semirings and the R\'enyi Entropy}\label{RenyiSec}

We have obtained a faithful encoding of the syntactic objects (with a head function) in a vector 
space $\mathcal{F}\otimes_\mathbb{R} \mathbb{R}[\underline{\Lambda}]$ of functions. The question then is how to improve on
this construction so that a faithful encoding can happen in the same space $\mathcal{F}$ and in a way
that also encodes the magma operation.

One issue about encoding the magma operation, as already mentioned in Remark~\ref{remDomh}, 
is that the domain $\Dom(h)$
of a head function is not compatible with the magma structure. This issue can be resolved
by passing to hypermagmas as in \cite{MHL}. 
There is, however, another issue discussed in Lemma~\ref{nonass}:
commutativity for the operation $\Gamma(\varphi(T), \varphi(T'))$ only holds up to the
transformation $\lambda \leftrightarrow 1-\lambda$, while the magma operation is
commutative, $\mathfrak{M}(T,T')=\mathfrak{M}(T',T)$.

We show that these two problems (maintaining the same space $\mathcal{F}$ of the embedding
and encoding the magma structure) can be solved at the same time using a construction
developed in \cite{MarThor} of ``thermodynamic semirings'' involving an entropy measure. 
The entropy functional will provide an addition operation that is nonassociative and
commutative, which will implement the magma operation of syntactic objects, while
performing an optimization over the variables $\lambda_v$, which reduces the embedding
space back to $\mathcal{F}$ by fixing an optimal value of these additional variables.

\subsection{Probabilities and entropy}\label{LTprob} 
The probability distribution we consider here, associated to a syntactic object $T$, is
the $A=(a_\ell)_{\ell\in L(T)}$ constructed as in Corollary~\ref{aellTpoly}. We verify
that it is indeed a probability distribution as follows.

\begin{lemma}\label{aprob}
	For all $T\in \mathfrak{T}_{\mathcal{S}\mathcal{O}_0}$ and for all $\Lambda=(\lambda_v)_{v\in V^0(T)} \in (0,1)^{\# V^0(T)}$, the vector $A=(a^T)_{\ell\in L(T)}$,
	with $a^T_\ell(\Lambda)$ as in Proposition~\ref{phiTprop} is a probability distribution on the set $L(T)$, namely $a^T_\ell(\Lambda)\geq 0$ and
	\begin{equation}\label{aT1}
		\sum_{\ell\in L(T)} a^T_\ell(\Lambda) =1 \, . 
	\end{equation}
\end{lemma}

\begin{proof} It is clear that, if all the $\lambda_v$ are in $(0,1)$ then the polynomials $a^T_{\ell}(\Lambda)$ take non-negative values,
since each $a^T_{\ell}(\Lambda)$ is a product $a^T_{\ell}(\Lambda)=\prod_{w\in \pi_\ell} c_w$ where $\pi_\ell$ is the path of vertices from the
leaf $\ell$ to the root vertex of $T$ and each $c_w$ is either $\lambda_w$ or $1-\lambda_w$, so $a^T_{\ell}(\Lambda)>0$. 
We can also show inductively that $\sum_\ell a_\ell=1$. 
For $\# L(T)=2$ we simply have $\lambda$ and $1-\lambda$
as the values of $a^T_\ell$. Similarly, with the $a^T_\ell$ equal to $ \lambda' \lambda$, $\lambda'  (1-\lambda)$, and $(1-\lambda')$
in the case of a tree $T$ with three leaves. Assuming the normalization \eqref{aT1} holds up to a given size, it suffices to show it also holds 
for $\mathfrak{M}(T,T')$, to inductively increase the size. For $\ell\in L(T)$ we have $\sum_{\ell} a^T_\ell =1$ and 
for $\ell'\in L(T')$ we also have $\sum_{\ell'} a^{T'}_{\ell'} =1$. Then $L(\mathfrak{M}(T,T'))=L(T)\sqcup L(T')$, with
$a^{\mathfrak{M}(T,T')}_\ell =\lambda a^T_\ell$ and $a^{\mathfrak{M}(T,T')}_{\ell'}=(1-\lambda) a_{\ell'}(T')$, so that one still
has a probability distribution. 
\end{proof} 

Since $A(T)=(a_\ell(T))_{\ell\in L(T)}$ is a probability distribution, we can compute its entropy/information
according to different kinds of information functionals, such as Shannon entropy, R\'enyi and Tsallis entropies. 

\begin{definition}\label{Sdef}
	A {\em binary information measure} is a continuous concave function $S: [0,1] \to \mathbb{R}_{\geq 0}$ with $S(0)=0=S(1)$
	and $S(p)=S(1-p)$. 
	The particular case of the Shannon entropy also satisfies the condition 
	\[
	 S(p)+(1-p)S(\frac{q}{1-p}) = S(p+q) + (p+q) S(\frac{p}{p+q})\, . 
	\]
\end{definition}

We will equivalently write either $S(p)$ or $S(p,1-p)$, where the latter notation explains the term ``binary information measure''.

\begin{definition}\label{tropical}
	The tropical semiring consists of the set $\mathbb{R} \cup \{ \infty \}$ endowed with two operations $\oplus$ and $\odot$.
	The product operation $\odot = + $ is the ordinary sum of real numbers, extended with the rule that
	$x+\infty=\infty=\infty +x$ for any $x\in \mathbb{R} \cup \{ \infty \}$. The identity element for $\odot$ is $0$.
	The sum operation $\oplus = \min$ is the minimum $x\oplus y=\min\{ x,y \}$, with identity element $\infty$.
	Both operations satisfy associativity and commutativity and have
	identity elements, and the product $\odot$ distributes over the sum $\oplus$. 
	However, since there are in general no additive inverses, this is a semiring rather than a ring. 
\end{definition}

We recall that, as mentioned in the introduction, we use here the terminology ``semiring'' to include
the possibility of {\em nonassociative semirings} (just as the terminology ``algebra'' that usually
refers to associative algebras is also routinely used, for example, to refer to Jordan algebras that are nonassociative).
The main difference here, with respect to the case of algebras, is that in the nonassociative semirings
we consider it is the addition that is a nonassociative magma, while the multiplication remains an associative monoid
(while in nonassociative algebras it is the multiplication that becomes nonassociative while the addition
remains associative). Thus, semirings that include this nonassociative case will have two 
operations, one of which (addition) is a magma while the other (multiplication) is a monoid, with  
a compatibility relation given by distributivity of multiplication over addition.

The following deformations of the tropical semirings were studied in \cite{MarThor}. We summarize
here the main construction.

\begin{remark}\label{thermo}
		A binary information functional $S$ determines a modified addition operation $\oplus_{S,\beta}$ on the same 
		set $\mathbb{R}\cup \{ \infty \}$. Instead of  $x\oplus y=\min\{ x,y \}$, one defines
		\begin{equation}\label{thermosum}
			x\oplus_{S,\beta} y= \min_{p\in [0,1]} \{ px + (1-p) y - \beta^{-1} S(p) \}\, .  
		\end{equation} 
		Then $\mathbb{T}_{S,\beta}=(\mathbb{R}\cup \{ \infty \}, \oplus_{S,\beta}, \odot)$ is still a semiring (called the
		{\em thermodynamic semiring} with entropy $S$ (see \cite{MarThor})). 
		The commutativity of the operation $\oplus_{S,\beta}$ follows from the symmetric property 
		$S(p)=S(1-p)$ of the binary information functional, the fact that $0$ is still the additive unit
		follows from the property $S(0)=0=S(1)$. However, in general $\mathbb{T}_{S,\beta}$ is {\em nonassociative}.
		Associativity in fact holds if and only if $S$ is the Shannon entropy. For any other entropy
		functional with $S(p)=S(1-p)$ and $S(0)=0=S(1)$ the addition operation in the thermodynamic semiring 
		is a nonassociative commutative magma.
\end{remark}

For our purposes here we will be choosing an entropy functional that is not the Shannon entropy, so that
the addition in the resulting thermodynamic semiring will be commutative but nonassociative:
the R\'enyi entropy will be the main example.

We introduce some preliminary notation that will be useful in the following. First, it will be convenient
to distinguish between the tree and the bracket notation for a given syntactic object.

\begin{definition}\label{bracketT}
	For a syntactic object $T\in \mathfrak{T}_{\mathcal{S}\mathcal{O}_0}$, with $\alpha_\ell \in \mathcal{S}\mathcal{O}_0$, for $\ell\in L(T)$ the labels at the leaves, 
	we write $\mathcal{B}_T((\alpha_\ell)_{\ell\in L(T)})$ for the ``bracket notation'' of $T$. For example:
	\begin{equation}\label{3BT}
		T= \Tree[ $\alpha_1$ [ $\alpha_2$ $\alpha_3$ ] ] \ \ \ \longleftrightarrow \ \ \  \mathcal{B}_T(\alpha_1,\alpha_2,\alpha_3) = \{ \alpha_1, \{ \alpha_2, \alpha_3 \} \} \, . 
	\end{equation}
\end{definition}

Then we want to introduce the relation between full binary trees and elements in a thermodynamic
semiring. To do this, we introduce a preliminary step before dealing with the actual syntactic objects,
where we consider binary rooted trees with leaves decorated by real numbers (instead of lexical items
and syntactic features as in the case of syntactic objects).

We recall here the setting considered in \S 10 of \cite{MarThor}. Given a full binary rooted tree $T$ with $n$ 
labelled leaves, where the leaves are labelled by real numbers $x_\ell \in \mathbb{R}$ (or in $\mathbb{R} \cup \{ \infty \}$),
one can assign to $T$ an $n$-ary information measure
$S_T$, which is obtained by considering the bracketing $\mathcal{B}_T$ description of the full binary rooted tree as in Definition~\ref{bracketT}. 
We introduce some notation that will be useful in the following.

\begin{definition}\label{bracketplusS}
	We write $\mathcal{B}_{T,S,\beta} (x_1,\cdots, x_n)$ for the bracketing of the sums $\oplus_{S,\beta}$ of the values $x_\ell$ 
	at the leaves, associated to the $\alpha_\ell \in \mathcal{S}\mathcal{O}_0$, according
	to the bracketing $\mathcal{B}_T (\alpha_1,\cdots,\alpha_n)$ that describes the full binary tree $T$: for example  
	\begin{equation}\label{ex3BTS}
		\mathcal{B}_{T,S,\beta}(x_1,x_2,x_3)\coloneqq x_1 \oplus_{S,\beta} (x_2 \oplus_{S,\beta} x_3) \ \ \ \text{ for } \    T = \Tree[ $x_1$ [ $x_2$ $x_3$ ] ]  \, . 
	\end{equation}  
	Note that since $\oplus_{S,\beta}$ is commutative, the bracketed sum $\mathcal{B}_{T,S,\beta} (x_1,\cdots,x_n)$ 
	only depends on $T$ as an abstract tree and does not require the choice of a planar structure.
\end{definition}

Given the explicit form \eqref{thermosum} of the addition $\oplus_{S,\beta}$, we can express the
bracketing $\mathcal{B}_T (\alpha_1,\cdots,\alpha_n)$ as an optimization over a combination of expressions
as in \eqref{thermosum}. To this purpose it is convenient to introduce another notation that keeps
track of each of these expressions.

\begin{definition}\label{sumprobP}
	We use the notation $\mathcal{B}_{T,S,\beta} (x_1,\cdots,x_n; \Lambda)$, with $\Lambda=(\lambda_v)_{v\in V^0(T)}$ a collection of
	probability distributions $(\lambda_v, 1-\lambda_v)$ at the
	non-leaf vertices of the tree $T$, to indicate the expression obtained by recursively computing the expression
	\[
	 \lambda_v x_{v_1} + (1-\lambda_v) x_{v_2} - \beta^{-1} S(\lambda_v) 
	\]
	where $x_{v_1}$ and $x_{v_2}$ are the previously computed expressions at the  two vertices $v_1$, $v_2$ below $v$. 
\end{definition}

The following property was proved in \S 10 of \cite{MarThor}. We rephrase it here according to our present setting and
choice of notation.

\begin{proposition}[\cite{MarThor}]\label{STconstr}
	Given a binary information measure $S$ that makes the associated thermodynamic semiring $\mathbb{T}_{S,\beta}$ commutative but nonassociative,
	the element $\mathcal{B}_{T,S,\beta} (x_1,\cdots, x_n)\in \mathbb{T}_{S,\beta}$ associated to a $T\in \mathfrak{T}_{\mathcal{S}\mathcal{O}_0}$ as in Definition~\ref{bracketplusS}
	satisfies
	\[
	 \mathcal{B}_{T,S,\beta} (x_1,\cdots,x_n) = \min_{\Lambda \in [0,1]^{\# V^o(T)}} \mathcal{B}_{T,S,\beta} (x_1,\cdots, x_n; \Lambda) \, . 
	\]
	As in \eqref{aTell}, let $a^T_\ell =P_\ell(\Lambda)$ be the polynomials describing the probability distribution at the leaves of the tree $T$, determined
	as in Proposition~\ref{phiTprop} from the $\Lambda=(\lambda_v) \in [0,1]^{\# V^o(T)}$. We write this simply as $A=P(\Lambda)$.
	There is an $n$-ary information measure $S_T$ associated to the tree $T$, which is recursively determined on $T=\mathfrak{M}(T_1,T_2)\in \Dom(h)$ (with a
	planarization by the head function), with the following property.
	For a probability distribution $A=(a_1,\cdots, a_n)$ at the leaves of $T$ we also define
	\begin{equation}\label{BTSA}
		\mathcal{B}_{T,S,\beta} (A; x_1,\cdots, x_n)\coloneqq \sum_{\ell\in L(T)} a_\ell \, x_\ell - \beta^{-1} S_T(A) \, . 
	\end{equation}  
	Then, for all $\Lambda \in [0,1]^{\# V^o(T)}$, we have the identity
	\[
	\mathcal{B}_{T,S,\beta} (x_1,\cdots, x_n; \Lambda) = \mathcal{B}_{T,S,\beta} (A=P(\Lambda); x_1,\cdots, x_n) \, . 
	\]
	This also implies the identity
	\[
	 \mathcal{B}_{T,S,\beta} (x_1,\cdots,x_n) = \min_{A\in \Delta_{\# L(T)}} \{ \sum_\ell a_\ell \, x_\ell - \beta^{-1} S_T(A) \} = \min_{A\in \Delta_{\# L(T)}} \mathcal{B}_{T,S,\beta} (A; x_1,\cdots, x_n) \, . 
	\]
\end{proposition}

In the example of \eqref{ex3BTS} this can be seen explicitly as follows.

\begin{example}\label{exBP}
		For the case of \eqref{ex3BTS}, we have two non-leaf nodes $v\in V^o(T)$, hence two probabilities $(\lambda_v, 1-\lambda_v)$. Moreover,
		we have a probability $A=(a_1,a_2,a_3)$ at the leaves, so we obtain in this case
		\begin{multline*}
		 \lambda_2 x_1 + (1-\lambda_2) \lambda_1 x_2 + (1-\lambda_2) (1-\lambda_1) x_3 - \beta^{-1} ( (1-\lambda_2) S(\lambda_1) + S(\lambda_2)) \\
		 = a_1\, x_1 + a_2\, x_2 + a_3\,  x_3 - \beta^{-1} (S(\frac{a_2}{1-a_1}) + (1-a_1) S(a_1))\, , 
		\end{multline*}
		with $\lambda_1= a_2 (1-a_1)^{-1}$ and $\lambda_2=a_1$. Thus, we obtain that the information measure at the leaves, determined by the tree $T$ is 
		\[ S_T(A)= S(\frac{a_2}{1-a_1}) + (1- a_1) S( a_1) \, , \]
		and the resulting element that is the image of $T$ in the thermodynamic semiring is given by 
		\begin{align*}
		 \mathcal{B}_{T,S,\beta}(x_1,x_2,x_3) & = \min_{\lambda_1,\lambda_2\in [0,1]} \{ \lambda_2 x_1 + (1-\lambda_2) \lambda_1 x_2 + (1-\lambda_2) (1-\lambda_1) x_3 \\
		 &\qquad \qquad \qquad \qquad - \beta^{-1} ( (1-\lambda_2) S(\lambda_1) + S(\lambda_2))  \} \\
		 & =
		\min_{A \in \Delta_2} \{ \sum_{i=1}^3 a_i \, x_i - \beta^{-1} (S(\frac{a_2}{1-a_1}) + (1-a_1) S(a_1)) \} \\
		& = \min_{A \in \Delta_2} \{ \sum_{i=1}^3 a_i\, x_i - \beta^{-1} S_T(A) \}  \, . 
	\end{align*}
\end{example}

\subsection{Thermodynamic semirings of functions}\label{ThermoFuncSec}

As in \cite{MarThor}, in addition to considering the deformations $\mathbb{T}_{S,\beta}$ of the tropical semiring,
reviewed above,  one can consider semirings of functions, for example continuous functions $\mathcal{C}(X,\mathbb{T})$, with
values in the tropical semiring, endowed with the pointwise semiring operations.  The thermodynamic
deformations of the tropical semiring then induce corresponding deformations $\mathcal{C}(X,\mathbb{T}_{S,\beta})$.
We consider here the function space $\mathcal{F}_{S,\beta}\coloneqq\mathcal{C}(X,\mathbb{T}_{S,\beta})$, endowed with this
semiring structure.  In particular if the original function space is taken to be $\mathcal{F}=\mathcal{C}(X,\mathbb{R})$, it can
be seen as contained inside $\mathcal{C}(X,\mathbb{T})$ and inside $\mathcal{F}_{S,\beta}\coloneqq\mathcal{C}(X,\mathbb{T}_{S,\beta})$ (with
different addition operations and with the original addition playing the role of the semiring
multiplication).

We consider here again the given $\varphi: \mathcal{S}\mathcal{O}_0\to \mathcal{F}$ which we now interpret as
$\varphi: \mathcal{S}\mathcal{O}_0\to \mathcal{F}_{S,\beta}$, providing a set of
functions that encode the lexical items. Instead
of considering combinations as elements of $\mathcal{F}\otimes \mathbb{R}[\Lambda]$ as in the previous
section, as a way of combining the functions $\varphi(\alpha_\ell)$ according to the
tree structure of a syntactic object,  we want to use a particular, optimal, choice of $\lambda_v$,
and correct the combination $\sum_\ell a_\ell \varphi(\alpha_\ell)$ so as to obtain an iterative form
of an operation that is nonassociative and commutative.

\begin{remark}\label{NLrem}
		One important change, with respect to the construction of \S \ref{WeightcombSec}, is that now
		the optimality of the choice is achieved {\em pointwise} in $X$, resulting in variables $\lambda_v(t)$ 
		that are themselves functions of the point $t\in X$. Thus, the combinations of the wavelet 
		functions $\varphi(\alpha_\ell)$ associated to the lexical items will no longer be linear combinations
		(i.e.~constant coefficients $a_\ell$) but combinations of the form $\sum_\ell a_\ell(t) \, \varphi(\alpha_\ell)(t)$, 
		with a probability distribution $A(t)=(a_\ell(t))_{\ell\in L(T)}$ that changes with $t\in X$, and that is 
		determined by a pointwise optimization procedure. 
\end{remark}

Before describing this procedure in more detail, we recall some properties of the information measure
that we will use for the thermodynamic semiring.

\subsubsection{The second R\'enyi entropy}\label{2ndRenyiSec}

Some of the results in  Section~\ref{magmaSec} (Theorem~\ref{magmahom}) apply to a broad range of
choices of information functional $S$, for the thermodynamic deformations of the tropical
semiring, namely any such functional for which the resulting addition $\oplus_{S,\beta}$ is
commutative but nonassociative. These include, for example, the R\'enyi  entropies $\Ry_\alpha$
for any value of $\alpha\neq 1$, or the Tsallis entropies $\Ts_q$ for any value $q\neq 1$.
(The case at $1$ gives back the Shannon entropy for which $\oplus_{S,\beta}$ is both
commutative and associative.) However, for the case of Theorem~\ref{Pembed}, we make a specific
choice of information measure $S$, namely a special value of the R\'enyi  entropy for $\alpha=2$
\begin{equation}\label{SRy2}
	S(p) = \Ry_2(p) \, .
\end{equation}
The second R\'enyi  entropy $\Ry_2$ is also known as the ``collision entropy''.
There are various reasons for making this choice, some conceptual and some practical.
The practicality lies in making some computations simpler than in the more general case
of arbitrary $\Ry_\alpha$ or $\Ts_q$. The more conceptual reason lies in a 
particularly useful property of the special value $\Ry_2$ of the R\'enyi entropy, namely
the fact that the second R\'enyi entropy has an interpretation in terms of 
optimization of redundant non-orthogonal basis expansion (see for instance \cite{Hon} in
signal analysis, and also \cite{BoPoPla} in quantum information). Moreover, R\'enyi entropy
is computable in brain activity, and has been used in models of visual perception,
image segmentation and tracking (for instance \cite{Sahoo, Tozzi}). It also has good estimation
algorithms (for instance \cite{HuNa,OmSk,Pal}). The connection between
entropy (Shannon entropy in this case) and brain signals, in relation to language, was
considered in \cite{Brenn, Coop}.  
While these connections of the second R\'enyi entropy
to some optimization properties in signal analysis and the occurrence of entropy functionals in
neurocomputational settings seem to support the choice of this particular type of function, the
main motivation here is not directly coming from empirical data but from algebraic structural reasons,
as we discuss below. It remains an interesting question, though, to relate more directly the role that 
it has in the optimization we consider with signal processing optimization and possibly with explicit
neurocomputational implementation. While this is not the focus of the present paper, we recall
briefly its role in signal analysis interpreted in terms of the lexical representation, for possible
use in future work. 

In the signal analysis setting one has a signal $\psi$ and a finite {\em dictionary} $\psi_1,\cdots, \psi_N$
of elementary signals that $\psi$ can be decomposed into. Assuming linearity, this means $\psi$ is
(approximately) equal to a linear combination $\sum_i a_i\, \psi_i$. In the nonlinear cases, the
combination is replaced by a nonlinear kernel.  In general one can assume that the
$\psi_i$ are linearly independent, but not necessarily that they are an orthogonal basis with respect to
an inner product. The dictionary itself may be very large, but some sparsity assumption is considered.
In our case, the dictionary would literally be a  ``dictionary'' consisting of the set $\mathcal{S}\mathcal{O}_0$ of lexical items 
and syntactic features and the set of their images $\varphi(\mathcal{S}\mathcal{O}_0)$ inside the space of functions
$\mathcal{F}_{S,\beta}$. Since the space $\mathcal{F}_{S,\beta}$ is not endowed with an inner product, we are exactly 
in the situation where we cannot rely on orthogonality of the basis for signal analysis. 
In such cases one can introduce a measurement of heterogeneity of the decomposition, a diversity
measure that shows the terms of the decomposition are sufficiently dissimilar. An effective such measure is provided
by the second R\'enyi entropy (see \cite{Hon}).

In our setting, we look at a different perspective, though.
We form (pointwise in $X$) certain combinations of atoms $\varphi(\alpha)$ with $\alpha\in \mathcal{S}\mathcal{O}_0$ with 
coefficients given by a probability distribution $A=(a_\ell)_{\ell\in L}$ (dependent on the point $t\in X$),
which depends on a binary rooted tree $T$ with $L=L(T)$ the set of leaves. We can
think of the resulting objects $\sum_\ell a_\ell \varphi(\alpha_\ell)$ as a random variable, depending
on the probability distribution $A=A(\Lambda)$, with $\Lambda=(\lambda_v)_{v\in V^o(T)}$. 
Thus, instead of just evaluating the $n$-ary second R\'enyi entropy $\Ry_2(A)$ as a diversity measure 
(as in  \cite{Hon}), we consider the $n$-ary entropy functionals built from the binary second R\'enyi entropy
and the tree $T$, namely the $S_T(a_1,\cdots, a_n) = \Ry_{2,T}(a_1,\cdots, a_n)$, with $n=\# L(T)$. 
Unlike the $n$-ary second R\'enyi entropy, the diversity measure $\Ry_{2,T}$ is adapted 
to the presence of a fixed tree structure $T$. Maximization of 
$\Ry_{2,T}(a_1,\cdots, a_n)$ corresponds to minimization of $-\beta^{-1} \Ry_{2,T}(a_1,\cdots, a_n)$.

As we discuss below, the choice of an optimal $A=A(\Lambda)$ follows this idea, but instead of
just minimizing $-\beta^{-1} \Ry_{2,T}(a_1,\cdots, a_n)$, we minimize the combination
\[
 \mathcal{B}_{T,\Ry_2,\beta}(A, \varphi(\alpha_1),\cdots,\varphi(\alpha_n))=
\sum_\ell a_\ell \varphi(\alpha_\ell) - \beta^{-1} \Ry_{2,T}(a_1,\cdots, a_n) \, . 
\]
Note that, since the $\varphi(\alpha_\ell)$ are functions on an underlying space $X$,
the minimization above is pointwise in $t\in X$, hence the resulting $A=A(\Lambda)$
is also a function of $t\in X$, rather than a constant probability distribution, as pointed
out in Remark~\ref{NLrem} above. 
This means that this decomposition performed pointwise in $t\in X$ is globally a decomposition with
coefficients $a_\ell(t)$ in a ring of functions of $t\in X$. We can then interpret the second R\'enyi
entropy as a measure of diversity for the coefficients $a_\ell(t)$ and the choice of $A$ that
is incorporated in the minimization of
$\mathcal{B}_{T,\Ry_2,\beta}(A, \varphi(\alpha_1),\cdots,\varphi(\alpha_n))$, for fixed 
$\{ \varphi(\alpha_\ell) \}$.

This discussion is only aimed at justifying heuristically
our choice of $S=\Ry_2$. The rigorous discussion in the rest of this section is
independent of these considerations.

In fact, as mentioned above, for us the main reason for choosing the second R\'enyi 
entropy is that the construction
described here of a composition of the $\varphi(\alpha_\ell)$ with coefficients given
by functions $a_\ell$ on $X$, determines an embedding of syntactic objects in the
space $\mathcal{F}_{\Ry_2,\beta}$ that is also a morphism of magmas, thus 
capturing the full algebraic structure of the set of syntactic objects.

We can compute explicitly, for $S=\Ry_2$ the form of the addition $\oplus_{\Ry_2,\beta}$.

\begin{lemma}\label{lemRy2add}
	For $S=\Ry_2$, the addition
	\[
	 x\oplus_{S,\beta} y =\min_{\lambda \in [0,1]} \{ \lambda x + (1-\lambda) y - \beta^{-1} S(\lambda) \} 
	\]
	takes the form 
	\begin{equation}\label{addRy2}
		x\oplus_{\Ry_2,\beta} y = 
		\min_{\lambda \in [0,1]} \{ \lambda x + (1-\lambda) y + \beta^{-1} \log (\lambda^2 + (1-\lambda)^2) \} \, ,
	\end{equation}
	with the minimum achieved at a value $0<  \lambda_{\min}(x,y)  < 1$ of the form
	\begin{equation}\label{lambdaminxyu}
		\lambda_{\min}(x,y) = \frac{u^{-1}}{2} (1 + u - \sqrt{ 1- u^2}) \ \ \  \text{ for } u=\beta (y-x)/2 \, ,
	\end{equation} 
	for $x,y\in \mathbb{T}_{S,\beta}$ such that $|y-x| < 2/\beta$. When $|y-x| \geq  2/\beta$ the
	minimum is at the boundary, $\lambda_{\min}(x,y) \in \{ 0,1 \}$. 
\end{lemma}

\begin{figure}[h]
	\begin{center}
		\includegraphics[scale=0.35]{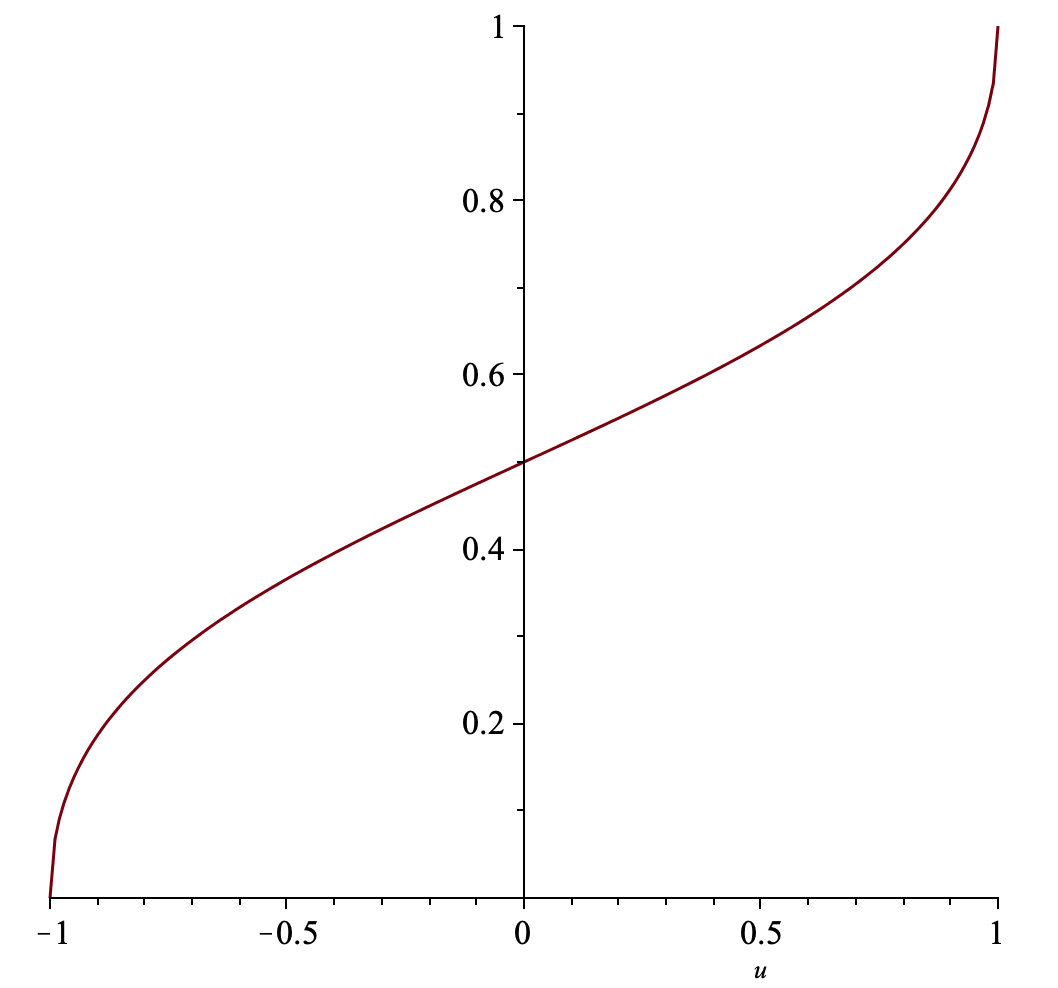} 
		\caption{The behavior of the function $\lambda_{\min}(x,y)$ of  \eqref{lambdaxyu}, 
		in the variable $u=\beta (y-x)/2$ in the range $|u|<1$.
			\label{FigLambdaMin}}
	\end{center}
\end{figure}

\begin{proof}
In the case $S=\Ry_2$ we have
\[
 x\oplus_{\Ry_2,\beta} y =\min_{\lambda \in [0,1]} \{ \lambda x + (1-\lambda) y - \beta^{-1} \Ry_2(\lambda) \}
= \min_{\lambda \in [0,1]} \{ \lambda x + (1-\lambda) y + \beta^{-1} \log (\lambda^2 + (1-\lambda)^2) \} \, . 
\]
We write
\begin{equation}\label{Flambda}
	F(\lambda)\coloneqq \lambda x + (1-\lambda) y + \beta^{-1} \log (\lambda^2 + (1-\lambda)^2)\, ,
\end{equation}
so that we obtain 
\[
\frac{\partial}{\partial \lambda} F(\lambda) = x-y + \beta^{-1} \frac{4\lambda -2}{\lambda^2 + (1-\lambda)^2} 
\]
and setting $\frac{\partial}{\partial \lambda} F(\lambda)=0$ we see that, for $y-x \geq 2\beta^{-1}$
we have $ \frac{\partial}{\partial \lambda} F(\lambda) <0$ inside the interval, so the minimum
is achieved at the boundary $\lambda=1$. When $0< y-x <2\beta^{-1}$, on the other hand, we have two
real solutions, with the smaller one being the minimum. This gives a solution $0< \lambda_{\min}(x,y) < 1$ of the form
\begin{equation}\label{lambdaxyu}
	\lambda_{\min}(x,y) = \frac{ -2\beta^{-1} + x-y + \sqrt{4 \beta^{-2} -(x-y)^2} }{2 (x-y) } = \frac{u^{-1}}{2} (1 + u - \sqrt{ 1- u^2}) 
\end{equation} 
with the change of variable $u=\beta (y-x)/2$, with $0< u <1$.  Symmetrically, 
under $u\mapsto -u$, for $u= \beta (x-y)/2$,
\begin{equation}\label{lambdaxyu2}
\lambda_{\min}(u) =\frac{1}{2}\left(1-\frac{u}{1+\sqrt{1-u^2}}\right)\, ,
\end{equation} 
extended by $1$ for $u<-1$ and by $0$ for $u>1$.

The minimum $\lambda_{\min}(u)$ as in \eqref{lambdaxyu} satisfies $\lambda_{\min}(u)=1-\lambda_{\min}(-u)$
for $|u|\leq 1$. Thus, when $y-x<0$, we can rewrite the minimization of \eqref{Flambda} as
\begin{multline}
	(1-\lambda_{\min}(-u)) x + \lambda_{\min}(-u) y + \beta^{-1} \log (\lambda_{\min}(-u)^2 + (1-\lambda_{\min}(-u))^2)\\
  = x + \lambda_{\min}(-u) \, 2\beta^{-1} u + \beta^{-1} \log (\lambda_{\min}(-u)^2 + (1-\lambda_{\min}(-u))^2) \, . 
 \label{xysign}
\end{multline}
We then have, for $y-x \leq - 2\beta^{-1}$ the minimum at $\lambda=0$ and for $ -2\beta^{-1}< y-x <0$
the minimum at $\lambda_{\min}(u)$ as in \eqref{lambdaxyu}.
The behavior of the minimum \eqref{lambdaxyu} in the range $|u|<1$ is
illustrated in Figure~\ref{FigLambdaMin}. Note the differential singularities at $u=\pm 1$, so that
the function is continuous (equal to $0$ for $u<-1$ and to $1$ for $u>1$) but only piecewise differentiable. 
\end{proof}

\begin{remark}\label{symmu}
		Note that the commutativity property $x\oplus_{\Ry_2,\beta} y=y\oplus_{\Ry_2,\beta} x$ 
		is also ensured by \eqref{xysign} since we have 
		$\lambda_{\min}(u) x + (1-\lambda_{\min}(u)) y = (1-\lambda_{\min}(-u)) x + \lambda_{\min}(-u) y$ and we also have
		$\Ry_2(\lambda)=\Ry_2(1-\lambda)$.
\end{remark}

\begin{definition}\label{LambdaOpt}
	Given a function $\varphi: \mathcal{S}\mathcal{O}_0 \to \mathcal{C}(X,\mathbb{R})$, and assuming $S=\Ry_2$ is
	the second R\'enyi entropy, for a given syntactic object $T\in \Dom(h)\subset \mathfrak{T}_{\mathcal{S}\mathcal{O}_0}$,
	with $L=L(T)$ and data $\{ \alpha_\ell \}_{\ell\in L}$ at the leaves, the {\em optimal $\Lambda$} is
	the function 
	\[
	 \Lambda=\Lambda_{T, \{\alpha_\ell\}_{\ell\in L(T)}} : X \to [0,1]^{\# V^o(T)}
	 \]
	given by the solution of the pointwise optimization
	\begin{equation}\label{argmin}
		\Lambda =\argmin_{\Lambda} \mathcal{B}_{T,S,\beta} (\varphi(\alpha_1),\cdots, \varphi(\alpha_n); \Lambda) 
		=\argmin_{\Lambda} \mathcal{B}_{T,S,\beta} (A=P(\Lambda); \varphi(\alpha_1),\cdots, \varphi(\alpha_n)) \, .
	\end{equation}
\end{definition}

\section{Embedding syntactic objects}\label{EmbSOsec}

The construction of Proposition~\ref{STconstr}  answers the question of whether 
there is a preferred choice of the probabilities $\Lambda \in [0,1]^{\# V^o(T)}$
used to map the tree $T$ to the vector space $\mathcal{F}$ as a combination $\varphi(T)$ of the functions $\varphi(\alpha_\ell)$ at the
leaves with coefficients $a_\ell^T = P_\ell(\Lambda)$.

We now discuss how to use the result of Proposition~\ref{STconstr} to construct embeddings of
syntactic objects inside a thermodynamic semiring of functions as described in \S \ref{ThermoFuncSec}. 
Namely, we want to construct embeddings of syntactic objects in $\mathcal{C}(X,\mathbb{R})$, in such
a way as to ensure that we can also capture the magma operation on syntactic objects, in terms of
a commutative, nonassociative operation on $\mathcal{C}(X,\mathbb{R})$.

As above let $\mathcal{F}_{S,\beta}=\mathcal{C}(X,\mathbb{T}_{S,\beta})$ be the semiring with the pointwise
operations induced by $\mathbb{T}_{S,\beta}$.
In general we are interested in the case where $X\subset \mathbb{R}^D$ is a compact 
region that has a nonempty dense open set $\mathcal{U} \subset X$. In fact, for
simplicity, we can just assume (possibly by enlarging the region) that $X$ 
is a compact $D$-dimensional smooth manifold without boundary. 
We then consider the dense subalgebra $\mathcal{C}^\infty(X,\mathbb{R})$ of
the algebra $\mathcal{C}(X,\mathbb{R})$ of continuous functions (with the usual
pointwise addition and multiplication). We can endow  $\mathcal{C}^\infty(X,\mathbb{R})$
with the thermodynamic semiring operations $\oplus_{S,\beta}$, $\odot$
by restriction from viewing $\mathcal{C}^\infty(X,\mathbb{R})\subset \mathcal{F}_{S,\beta}$. 
We write $\mathcal{F}_{S,\beta}^\infty$ for $\mathcal{C}^\infty(X,\mathbb{R})$ with these
semiring operations.

We view a function $\varphi: \mathcal{S}\mathcal{O}_0 \to \mathcal{C}(X,\mathbb{R})$
as above as taking values in $\mathcal{F}_{S,\beta}$. We will also assume here that these
functions are in fact smooth, namely $\varphi: \mathcal{S}\mathcal{O}_0 \to \mathcal{C}^\infty(X,\mathbb{R})$. By the density
of $\mathcal{C}^\infty(X,\mathbb{R})$ in $\mathcal{C}(X,\mathbb{R})$ this can always be achieved by approximation.

\begin{definition}\label{phiSbetaTdef}
	Given $\varphi: \mathcal{S}\mathcal{O}_0 \to \mathcal{C}^\infty(X,\mathbb{R})$ as above, and $S=\Ry_2$.
	We define the function $\varphi_{S,\beta}: \mathfrak{T}_{\mathcal{S}\mathcal{O}_0} \to \mathcal{F}_{S,\beta}$ in
	the following way. Given a finite set $L$ and a collection
	$\{ \alpha_\ell \}_{\ell \in L} \subset \mathcal{S}\mathcal{O}_0$, set
	\begin{equation}\label{phiSbeta}
		\varphi_{S,\beta}(T)=\mathcal{B}_{T,S,\beta}(\varphi(\alpha_1),\cdots, \varphi(\alpha_n)) = \sum_\ell a_\ell(T) \varphi(\alpha_\ell) -\beta^{-1} S_T(A) \, ,   
	\end{equation}
	for optimal $A=(a_\ell)=P(\Lambda)$, 
	with optimal $\Lambda$ as in \eqref{argmin}.
\end{definition}

\subsection{Uniform bound and high temperature regime} \label{HighTSec}

We recall here some properties of spaces of functions that we will need in the following.
We assume here that the space $X$ is compact. For example, a closed bounded interval $[a,b]\subset \mathbb{R}$ or the
product $[a,b]\times \Omega$ of a closed bounded region with nonempty interior, $\Omega\subset \mathbb{R}^D$, 
typically with $D=2$ or $D=3$, representing a region of brain locations and a closed interval of time measurement.

\begin{remark}\label{compXrem}
		For $X$ compact, continuous functions $f\in \mathcal{C}(X,\mathbb{R})$ have a finite maximum and minimum in $X$. Thus,
		given a finite set $\{ \varphi(\alpha_\ell) \}_{\ell \in L}$ in $\mathcal{F}_{S,\beta}=\mathcal{C}(X,\mathbb{T}_{S,\beta})$ that take
		values in $\mathbb{R}$ (rather than $\mathbb{R} \cup \{ \infty \}$), there is a maximum $M_L$ so that
		\begin{equation}\label{MLmax}
			| \varphi(\alpha_\ell)(t) | \leq M_L < \infty  \, , 
		\end{equation} 
		for all $t\in X$ and $\ell\in L$. 
\end{remark}

\begin{definition}\label{betaLdef}
	For $X$ compact, and for a chosen set $L$ of leaves and corresponding functions
	$\{ \varphi(\alpha_\ell) \}_{\ell \in L}$ in $\mathcal{F}_{S,\beta}$, with $M_L$ the maximum
	as in \eqref{MLmax}, we choose a $\beta=\beta_L$ with the
	property that $\beta_L = M_L^{-1}$. The range $\beta< \beta_L$ 
	is the high temperature regime for the syntactic object $T$ with $L(T)=L$
	with the given data $\{ \varphi(\alpha_\ell) \}_{\ell \in L}$ at the leaves. We can also consider the set
	$\{ \varphi(\alpha) \}_{\alpha\in \mathcal{S}\mathcal{O}_0}$. Since $\mathcal{S}\mathcal{O}_0$ is a finite set, there is a
	global maximum $M_{X,\mathcal{S}\mathcal{O}_0}< \infty$ for all the $\varphi(\alpha)$ on the compact $X$, 
	with $M_L\leq M_{X,\mathcal{S}\mathcal{O}_0}$ for any $L$,
	and such that
	\[
	 | \varphi(\alpha)(t) | \leq M_{X,\mathcal{S}\mathcal{O}_0} < \infty \ \ \  \forall \alpha\in \mathcal{S}\mathcal{O}_0 \, . 
	 \]
	We refer to 
	\begin{equation}\label{lowtempX}
		\beta< \beta_{X,\mathcal{S}\mathcal{O}_0}=M_{X,\mathcal{S}\mathcal{O}_0}^{-1}
	\end{equation}
	as the {\em global high temperature regime}.
\end{definition}

Then we obtain the following, as a direct consequence of Definition~\ref{betaLdef}.

\begin{lemma}\label{HighTu1}
	In the high temperature regime $\beta< \beta_L$, the estimate $|u(t)|<1$ holds for all $t\in X$, for
	\begin{equation}\label{ut}
		u(t) = \frac{\beta}{2} ( \varphi(\alpha_\ell)(t) - \varphi(\alpha_{\ell'})(t) ) 
	\end{equation} 
	for any $\ell, \ell' \in L$. In the global high temperature regime $\beta< \beta_{X,\mathcal{S}\mathcal{O}_0}$
	the same $|u(t)|<1$ estimate holds for 
	\[
	 u(t) = \frac{\beta}{2} ( \varphi(\alpha)(t) - \varphi(\alpha')(t) ) 
	 \]
	for any $\alpha,\alpha'\in \mathcal{S}\mathcal{O}_0$.
\end{lemma}

\begin{proof} We have 
\[
 |u(t)|=\frac{\beta}{2} |\varphi(\alpha_\ell)(t) - \varphi(\alpha_{\ell'})(t)| \leq \frac{\beta}{2} (|\varphi(\alpha_\ell)(t)|+|\varphi(\alpha_{\ell'})(t)|)
\leq \beta M_{X,\mathcal{S}\mathcal{O}_0} < 1 \, . 
\]
\end{proof}

\subsection{Single wall-crossing}\label{WallSec}

Any two trees $T\neq T'$ with $L(T)=L(T')=L$ and the same assignment $\{ \alpha_\ell \}_{\ell\in L}$
at the leaves, will differ from
one another by a series of moves that cross one of the boundary strata in the BHV$_n$ moduli space, 
for $n=\# L$, see \cite{BHV}. Namely, if we can assign to both $T,T'$ constant edge length equal to $1$ at all edges,
moving to a boundary stratum corresponds to shrinking one (or more, for the higher codimension strata of the
boundary) of the edges to zero length, hence producing a vertex of higher valence. Crossing a one dimensional
boundary stratum corresponds to moving one branching across a vertex, by first shrinking an edge to
zero and then creating a new edge of growing length along the other side of the ``collision vertex''.   
Any $T\neq T'$ with the same set of labelled leaves differ by repeated application of a number of 
such operations. We first consider  the case of $T\neq T'$ that differ by a single such operation 
at one of their non-leaf vertices. We then discuss how to deal with an arbitrary pair $T\neq T'$
with the same data at the leaves.

\begin{theorem}\label{3Tcase}
	Consider a pair of trees $T\neq T'$ with $L(T)=L(T')=L$
	and with the same assignment $\{ \alpha_\ell \}_{\ell\in L}$, that
	differ by a single wall crossing in the BHV$_n$ moduli space, with $\# L=n$. 
	Then for such a pair 
	\[
	 \varphi_{S,\beta}(T)=\mathcal{B}_{T,S,\beta}(\varphi(\alpha_1),\cdots, \varphi(\alpha_n))  \neq
	\varphi_{S,\beta}(T')=\mathcal{B}_{T',S,\beta}(\varphi(\alpha_1),\cdots, \varphi(\alpha_n)) \, . 
	\]
\end{theorem}

\begin{proof}
In terms of $\Lambda$, this transformation
of the tree corresponds to the transformation that maps  
\begin{equation}\label{BHVwall}
	\xymatrix{ &  & v  \ar@{-}[dl]_{\lambda_v} \ar@{-}[ddrr]^{1-\lambda_v} & &  \\
		& w \ar@{-}[dl]_{\lambda_w} \ar@{-}[dr]^{1-\lambda_w}  & & &  \\  w_1  &  & w_2 & & v_2  } \ \  \longleftrightarrow \ \ \ 
	\xymatrix{ &  & w  \ar@{-}[ddll]_{\lambda_w} \ar@{-}[dr]^{1-\lambda_w} & &  \\
		& & & v \ar@{-}[dl]_{\lambda_v} \ar@{-}[dr]^{1-\lambda_v}  &  \\  w_1 &  & v_1=w_2 & & v_2 }
\end{equation}
In the left-hand-side of the diagram \eqref{BHVwall} the coefficients $\lambda_v$ and $\lambda_w$
are evaluated at the optimal $\Lambda=\Lambda_{T, \{ \alpha_\ell \}}$ and in the right-hand-side at the optimal
$\Lambda'=\Lambda_{T', \{ \alpha_\ell \}}$. More precisely, since optimization is done one step at a time in
the tree, according to the bracketing, with each bracketing computing a single $\oplus_{S,\beta}$ operation,
this means that in the left-hand-side of the diagram \eqref{BHVwall} the coefficients $\lambda_v$ and $\lambda_w$
realize the optimization that computes the term 
\begin{align}\nonumber
	\varphi_{S,\beta}(T_v) & =  \left( \mathcal{B}_{T_{w_1},S,\beta}((\alpha_\ell)_{\ell\in L(T_{w_1})}) \oplus_{S,\beta} \mathcal{B}_{T_{w_2},S,\beta}((\alpha_\ell)_{\ell\in L(T_{w_2})}) \right)
	\oplus_{S,\beta} \mathcal{B}_{T_{v_2},S,\beta}((\alpha_\ell)_{\ell\in L(T_{v_2})})  \\
\label{Xiv}
 & =  (\varphi_{S,\beta}(T_{w_1}) \oplus_{S,\beta} \varphi_{S,\beta}(T_{w_2}) ) \oplus_{S,\beta} \varphi_{S,\beta}(T_{v_2}) \, 
\end{align}
while in the right-hand-side they realize the optimization  that computes the term 
\begin{align}\nonumber
	\varphi_{S,\beta}(T'_v) &= \mathcal{B}_{T_{w_1},S,\beta}((\alpha_\ell)_{\ell\in L(T_{w_1})}) \oplus_{S,\beta} 
	\left( \mathcal{B}_{T_{v_1},S,\beta}((\alpha_\ell)_{\ell\in L(T_{v_1})}) 
	\oplus_{S,\beta} \mathcal{B}_{T_{v_2},S,\beta}((\alpha_\ell)_{\ell\in L(T_{v_2})}) \right) 
\\
\label{Xiv1}
& = \varphi_{S,\beta}(T_{w_1}) \oplus_{S,\beta} ( \varphi_{S,\beta}(T_{w_1}) \oplus_{S,\beta} \varphi_{S,\beta}(T_{v_2}) ) \, . 
\end{align}
Note how the subtrees $T_{w_1}$, $T_{w_2}$, $T_{v_2}$ are the same in both $T$ and $T'$ hence
they map to the same $\varphi_{S,\beta}(T_{w_1})$, $\varphi_{S,\beta}(T_{w_2})$, $\varphi_{S,\beta}(T_{v_2})$
in both \eqref{Xiv} and \eqref{Xiv1}. We then show that
\begin{equation}\label{3noteq}
	(\varphi_{S,\beta}(T_{w_1}) \oplus_{S,\beta} \varphi_{S,\beta}(T_{w_2}) ) \oplus_{S,\beta} \varphi_{S,\beta}(T_{v_2}) \neq
	\varphi_{S,\beta}(T_{w_1}) \oplus_{S,\beta} ( \varphi_{S,\beta}(T_{w_1}) \oplus_{S,\beta} \varphi_{S,\beta}(T_{v_2}) ) \, , 
\end{equation}
for a general choice of the function $\varphi: \mathcal{S}\mathcal{O}_0 \to \mathcal{F}_{S,\beta}$. 
Consider the function $X \to \mathbb{R}^3$ that maps 
\begin{equation}\label{Vmap}
	\varphi_v : t \mapsto (x(t),y(t),z(t))\coloneqq(\varphi_{S,\beta}(T_{w_1})(t), \varphi_{S,\beta}(T_{w_2}) (t), \varphi_{S,\beta}(T_{v_2})(t))\, . 
\end{equation} 
The condition \eqref{3noteq} states that there is a nonempty set of $t\in X$ such that the image $\varphi_v(t)=(x(t),y(t),z(t))$ under
this map $\varphi_v$ is not contained in the locus 
\begin{equation}\label{3hyp}
	H = \{ (x,y,z) \in \mathbb{R}^3 \,|\, x\oplus_{S,\beta} (y \oplus_{S,\beta} z) = (x\oplus_{S,\beta} y) \oplus_{S,\beta} z \} \, . 
\end{equation}
The locus \eqref{3hyp} inside $\mathbb{R}^3$, if nonempty, is of codimension at least one, so it is always possible,
up to a small (in the supremum norm in $\mathcal{C}(X,\mathbb{R})$) perturbation, to have a function $\varphi: \mathcal{S}\mathcal{O}_0 \to \mathcal{C}(X, \mathbb{R})$
for which the image of $\varphi_v$ of \eqref{Vmap} is not contained in \eqref{3hyp}, hence \eqref{3noteq} holds.
In fact, we can ensure that the set of $t\in X$ for which $\varphi_v(t)\notin H$ is a dense open subset $\mathcal{U}_v$
of $X$. This gives us that $\varphi_{S,\beta}(T_v) \neq \varphi_{S,\beta}(T'_v)$ at the vertex $v$ where the move
described in \eqref{BHVwall} is performed. If $T$ and $T'$ differ only by this single move, the rest of the trees 
$T$ and $T'$ is then obtained by merging of $T_v$ (respectively, $T'_v$) with the same $T_{\tilde v}$ with
$\tilde v$ the twin vertex of $v$ in both trees, and then repeatedly merge in other trees (always the same
ones for both $T$ and $T'$ until the root is reached). Thus it suffices to show that if 
$\varphi_{S,\beta}(T_v) \neq \varphi_{S,\beta}(T'_v)$, then also $\varphi_{S,\beta}(T_v)\oplus_{S,\beta} \varphi_{S,\beta}(T_{\tilde v}) \neq \varphi_{S,\beta}(T'_v)\oplus_{S,\beta} \varphi_{S,\beta}(T_{\tilde v})$. For this one can consider the properties of the 
successor function of the thermodynamic semiring, as in \S 9 of \cite{MarThor}, namely the function
$x \oplus_{S,\beta} 0$. Indeed, we have 
\[
 x \oplus_{S,\beta} y = y +  ((x-y) \oplus_{S,\beta} 0) = y+ \min_\lambda \{ \lambda(x-y)-\beta^{-1}S(\lambda) \}\, . 
 \]
We can then see that the successor function $x \mapsto x \oplus_{S,\beta} 0$ is injective as a function $\mathbb{R} \to \mathbb{R}$.
For the choice $S=\Ry_2$ of the second R\'enyi entropy, this can be seen directly from \eqref{lambdaxyu}, using
the fact that 
\[ 
\frac{\partial}{\partial u} \lambda_{\min}(u)=\frac{1-\sqrt{1-u^2}}{2u^2 \sqrt{1-u^2}} > 0 
\]
for $|u|<1$, which holds in the high temperature regime by Lemma~\ref{HighTu1},
or from general considerations as in \S 9 of \cite{MarThor}. 
The injectivity of the successor function then implies that, for $\varphi_{S,\beta}(T_v) 
\neq \varphi_{S,\beta}(T'_v)$ we also have
\[ (\varphi_{S,\beta}(T_v)-\varphi_{S,\beta}(T_{\tilde v}))\oplus_{S,\beta} 0 \neq 
(\varphi_{S,\beta}(T'_v)-\varphi_{S,\beta}(T_{\tilde v}))\oplus_{S,\beta} 0 
\]
which gives the condition we want, which can then be iterated to obtain $\varphi_{S,\beta}(T)\neq \varphi_{S,\beta}(T')$.
We can consider the condition that $\varphi_v(t)\notin H$ for each $v\in V^o(T)$, which ensures
that \eqref{3noteq} holds for the tree $T'$ obtained from $T$ by the same single move at the vertex $v$. The
intersection $\cap_v \mathcal{U}_v$ gives again a dense open set on which all the $\varphi_v(t)\notin H$  conditions 
hold at once, hence a generic $\varphi: \mathcal{S}\mathcal{O}_0 \to \mathcal{C}(X, \mathbb{R})$ for which \eqref{3noteq} holds for every $v\in V^o(T)$.
Thus we obtain $\varphi_{S,\beta}(T)\neq \varphi_{S,\beta}(T')$ whenever $T$ and $T'$ differ by a single move as
in \eqref{BHVwall} at any of the vertices. 
\end{proof}

We then need to extend this argument to the case where $T$ and $T'$ differ by multiple moves of the type 
described in \eqref{BHVwall} performed at several vertices. One can iterate the same argument, where
at each vertex where
a move of type \eqref{BHVwall} happens, one is checking the transversality of a function
(obtained from the previous steps) to the same hypersurface \eqref{3hyp}. It is however
more convenient to just argue directly in terms of the whole tree using the locus 
\[
 \mathcal{B}_{T,S,\beta}(x_1,\cdots, x_n) = \mathcal{B}_{T',S,\beta}(x_1,\cdots, x_n) 
 \]
inside $\mathbb{R}^n$, as in Theorem~\ref{Pembed} below.

\subsection{Transversality and embeddings of syntactic objects} \label{TransvSec}

We then obtain the following general result. 

\begin{theorem}\label{Pembed}
	There is a dense open set of functions $\varphi\in \mathcal{C}^\infty(X,\mathbb{R})^{\mathcal{S}\mathcal{O}_0}$, 
	with the property that, for any finite set $L$ with $\# L \leq \# \mathcal{S}\mathcal{O}_0$  
	and any collection $\{ \alpha_\ell \}_{\ell \in L} \subset \mathcal{S}\mathcal{O}_0$ with
	$\alpha_\ell \neq \alpha_{\ell'}$ for $\ell\neq \ell' \in L$, the 
	function $\varphi_{S,\beta}$ of Definition~\ref{phiSbetaTdef}
	satisfies $\varphi_{S,\beta}(T)\neq \varphi_{S,\beta}(T')$
	for any pair $T\neq T'$ with the same $L(T)=L(T')=L$ and the same data $\{ \alpha_\ell \}_{\ell \in L}$ at the leaves.
\end{theorem}

\begin{proof} Given the data $L$ and $\{ \alpha_\ell \}_{\ell \in L} \subset \mathcal{S}\mathcal{O}_0$, we have a function
$L \stackrel{\alpha}{\rightarrow} \mathcal{S}\mathcal{O}_0 \stackrel{\varphi}{\rightarrow} \mathcal{C}^\infty(X,\mathbb{R})$, 
hence a continuous map $\alpha^*: \mathcal{C}^\infty(X,\mathbb{R})^{\mathcal{S}\mathcal{O}_0} \to \mathcal{C}^\infty(X,\mathbb{R})^{L}$, equivalently seen
as a map $\alpha^*: \mathcal{C}^\infty(X,\mathbb{R}^N) \to \mathcal{C}^\infty(X,\mathbb{R}^n)$, for $N=\# \mathcal{S}\mathcal{O}_0$ and $n=\# L$. 
For $T\neq T'$, with the same $L(T)=L(T')=L$ and $\{ \alpha_\ell \}_{\ell \in L}$, the expressions
$\varphi_{S,\beta}(T)=\mathcal{B}_{T,S,\beta}(\varphi(\alpha_1),\cdots, \varphi(\alpha_n))$ and
$\varphi_{S,\beta}(T')=\mathcal{B}_{T',S,\beta}(\varphi(\alpha_1),\cdots, \varphi(\alpha_n))$ of \eqref{phiSbeta}
differ by the bracketing of the sequence of binary operations $\oplus_{S,\beta}$ performed at each
node of the tree. Since the semiring  $\mathbb{T}_{S,\beta}$ is nonassociative, {\em in general} 
\[
 \mathcal{B}_{T,S,\beta}(x_1,\cdots, x_n) \neq \mathcal{B}_{T',S,\beta}(x_1,\cdots, x_n) 
 \]
for a dense open set of $(x_1,\cdots,x_n)\in \mathbb{R}^n$. The locus 
\begin{equation}\label{HTT}
	H_{T,T'}=\{ (x_1,\cdots,x_n)\in \mathbb{R}^n\,|\,  \mathcal{B}_{T,S,\beta}(x_1,\cdots, x_n) = \mathcal{B}_{T',S,\beta}(x_1,\cdots, x_n) \} 
\end{equation}
if nonempty, is of codimension at least one. In fact, for the specific case of $S=\Ry_2$, the
second R\'enyi entropy, we know from the explicit computation \eqref{lambdaxyu} of the minimizer,
that the repeated application of this formula over the sequence 
of binary operations $\oplus_{S,\beta}$ performed at each
node of the tree yields an expression that is almost everywhere a smooth function, except at
the vanishing points of the expressions under square roots, where it can have singularities.
By the implicit function theorem $H_{T,T'}$ is (away from a lower-dimensional singularity locus) 
a smooth submanifold of $\mathbb{R}^n$ of codimension one (a hypersurface with singularities). 
By \eqref{lambdaxyu}, the singularities occur when $|u|=1$, in each application of \eqref{lambdaxyu}
for each $\oplus_{S,\beta}$ involved. 

Thus, for each pair $T\neq T'$
we are looking at the relative position of the two maps
\[
 \xymatrix{ X \ar[r]^{\alpha^*(\varphi)} & \mathbb{R}^n \\ & H_{T,T'} \ar@{^{(}->}[u] }.
 \] 
In general, given a smooth function  $\phi: X \to \mathbb{R}^n$, $\phi(t)=(\phi_1(t),\cdots, \phi_n(t))$, 
with $\phi_i \in \mathcal{C}^\infty(X,\mathbb{R})$, and a subset $H \subset \mathbb{R}^n$ that is a smooth submanifold 
of positive codimension, the function $\phi$ is transverse to $H$ if, for every $t\in \phi^{-1}(H)$
\begin{equation}\label{HXtransv}
	d_t \phi (T_t X) + T_{\phi(t)} H = T_{\phi(t)} \mathbb{R}^n \simeq \mathbb{R}^n \, .
\end{equation}
Namely, the span of the image of the tangent space of $X$ and the tangent space of $H$ span
all of the $n$ dimensions of the ambient space $\mathbb{R}^n$. One writes $\phi \pitchfork H$ for
this transversality condition. More generally, for $K \subset X$, one writes $\phi \pitchfork_K H$
to denote the fact that \eqref{HXtransv} holds at all $x\in K \cap \phi^{-1}(H)$. We also use the notation
\begin{equation}\label{transvMNK}
	\pitchfork_K (X,\mathbb{R}^n; H) \coloneqq \{ \phi\in \mathcal{C}^\infty(X,\mathbb{R}^n) \,|\, \phi \pitchfork_K H \} \, . 
\end{equation}
The transversality theorem (see \cite{Hirsch}) shows that, if $H$ is closed and $K$ is compact, 
$\pitchfork_K (X,\mathbb{R}^n; H)$ is open and dense in $\mathcal{C}^\infty(X,\mathbb{R}^n)$. 

We are here assuming that $X$ 
is a compact $D$-dimensional smooth manifold without boundary, where
in general $1\leq D\leq 4$. Thus, except in the case where $n=3$, we
have that $D$ is smaller than both $N$ and $n$. In the case of $n=3$,
there is a single hypersurface $H_{T,T'}$ corresponding to the
condition $(x_1\oplus_{S,\beta} x_2) \oplus_{S,\beta} x_3=x_1\oplus_{S,\beta} (x_2 \oplus_{S,\beta} x_3)$
in $\mathbb{R}^3$. This is the case we have already discussed in Theorem~\ref{3Tcase}.
Thus, we can assume here that $n\geq 4$ and that $D\leq n$ (hence also $D\leq N$ by
hypothesis). Since $D\geq 1$ the transversality theorem can apply, with $H$ a smooth
hypersurface in $\mathbb{R}^n$, since $\dim d_t \phi (T_t X)=D$ and $\dim T_{\phi(t)} H=n-1$
and $D+n-1\geq n$. 

Since the locus $H=H_{T,T'}$ has singularities at which $T_{x} H$ is not well defined,
the transversality condition \eqref{HXtransv} can only hold on the complement of the
singular locus of $H_{T,T'}$. A way to avoid this problem is to introduce a small
smooth deformation of $H_{T,T'}$ in a small tubular neighborhood in $\mathbb{R}^n$ around 
the singular locus, and unchanged everywhere else. We denote by $H_{T,T'}^\epsilon$
such a small smooth deformation, depending on a small $\epsilon>0$. Then 
$H_{T,T'}^\epsilon$ is a smooth hypersurface in $\mathbb{R}^n$ and the transversality theorem
applies, ensuring the existence of an open dense subset of functions
\[
 \mathcal{U}^{\epsilon}_{T,T'} \subset \mathcal{C}^\infty(X,\mathbb{R}^n) 
\]
such that $\phi \pitchfork H_{T,T'}^\epsilon$ for all $\phi\in \mathcal{U}^{\epsilon}_{T,T'}$.

Considering then all the pairs $(T,T')$ with $T\neq T'$ and $L(T)=L(T')=L$ with $\# L=n$,
we have 
\[
  \mathcal{U}^{\epsilon}_n \coloneqq \bigcap_{T,T'}  \mathcal{U}^{\epsilon}_{T,T'} \subset \mathcal{C}^\infty(X,\mathbb{R}^n) 
\]
which is a finite intersection of dense open sets, hence itself a dense open set,
where transversality simultaneously holds for all the $H_{T,T'}^\epsilon$. 

This means that, if we can ensure that $\alpha^*(\varphi)\in \mathcal{U}^{\epsilon}_n$,
by an appropriate choice of $\varphi\in \mathcal{C}^\infty(X,\mathbb{R}^N)$, and for fixed
$\alpha$, we can achieve transversality in all the diagrams 
\begin{equation}\label{transvepsilon}
	\xymatrix{ X \ar[r]^{\alpha^*(\varphi)} & \mathbb{R}^n \\ & H^\epsilon_{T,T'} \ar@{^{(}->}[u] }
\end{equation}

The map $\alpha^*: \mathcal{C}^\infty(X,\mathbb{R}^N)\to \mathcal{C}^\infty(X,\mathbb{R}^n)$ acts by mapping the elements $\ell\in L$
that label the standard orthonormal basis of $\mathbb{R}^n$ to elements $\alpha_\ell \in \mathcal{S}\mathcal{O}_0$, where
$\mathcal{S}\mathcal{O}_0$ labels the standard orthonormal basis of $\mathbb{R}^N$, and restricting a map 
$\varphi(t)=(\varphi_\alpha(t))$ to a map 
$\alpha^*(\varphi)(t)=(\alpha^*(\varphi)_\ell(t))=(\varphi_{\alpha_\ell}(t))$.
Thus, the preimage 
\[
 (\alpha^*)^{-1}(\mathcal{U}^{\epsilon}_n)=\{ \varphi=(\varphi_\alpha) \,|\, (\varphi_{\alpha_\ell})_{\ell \in L} \in \mathcal{U}^{\epsilon}_n \} 
\subset \mathcal{C}^\infty(X,\mathbb{R}^N)  
\]
is the subset on which transversality in all the diagrams \eqref{transvepsilon} is achieved. 
If the data $\{ \alpha_\ell \}_{\ell\in L}$ satisfy $\alpha_\ell \neq \alpha_{\ell'}$ for $\ell\neq \ell' \in L$,
then the map $\alpha^*: \mathcal{C}^\infty(X,\mathbb{R}^N)\to \mathcal{C}^\infty(X,\mathbb{R}^n)$ is just a projection onto the
$\alpha_\ell$ coordinates of $\varphi=(\varphi_\alpha)$, hence it is an open map, so the preimage
of a dense set is dense. Thus, for all $n \leq N$, with any choice of data  
$\{ \alpha_\ell \}_{\ell\in L}$ satisfying $\alpha_\ell \neq \alpha_{\ell'}$ for $\ell\neq \ell' \in L$,
we have $(\alpha^*)^{-1}(\mathcal{U}^{\epsilon}_n)$ a dense open set in $\mathcal{C}^\infty(X,\mathbb{R}^N)$,
and the intersection of these (finitely many) dense open sets is still a dense open set
on which all these transversality conditions are simultaneously realized.
\end{proof}

This argument is limited to the case where 
$\alpha_\ell \neq \alpha_{\ell'}$ for $\ell\neq \ell' \in L$, which in particular implies $n\leq N$,
imposing a finite cap on the size of the syntactic objects $T\in \mathfrak{T}_{\mathcal{S}\mathcal{O}_0}$
with $\# L(T)\leq N$. 

If we allow cases where $\alpha_\ell = \alpha_{\ell'}$ for some $\ell\neq \ell' \in L$,
then the map $\alpha^*: \mathcal{C}^\infty(X,\mathbb{R}^N)\to \mathcal{C}^\infty(X,\mathbb{R}^n)$ is a projection
followed by an embedding of the image of this projection as a diagonal in $\mathcal{C}^\infty(X,\mathbb{R}^n)$
and embeddings are open maps only when their image is open (which is not the case here),
so the preimage $(\alpha^*)^{-1}(\mathcal{U}^{\epsilon}_n)$ will no longer be a dense set. In fact,
it is clear that, if some conditions $\alpha_\ell = \alpha_{\ell'}$ hold, there can be some trees
$T \neq T'$ that may no longer be distinguishable. 

So this presents two problems from the point of view of encoding syntactic objects: one is the
cap on the size by $\# L(T)\leq N$ and one is (even for trees that meet this size constraint)
the fact that we cannot accommodate the case of some $\alpha_\ell = \alpha_{\ell'}$. 

These two issues can be analyzed by identifying the kind of syntactic objects where
these problems would arise.

\subsection{Repetitions and copies} \label{RepCopSec} 

The assumption that $\alpha_\ell \neq \alpha_{\ell'}$ for $\ell\neq \ell'$ in $L$,
used in Theorem~\ref{Pembed}, is not realistic from the linguistic viewpoint,
as one certainly has syntactic objects where some of the lexical
items are {\em repetitions}, as in  
``{\em John meets John}'', where the two Johns are understood
to not be the same. However, repetitions are not {\em copies}, 
hence one can distinguish them as in {\em John}$_1$ and {\em John}$_2$ 
and assume that they are stored as different $\varphi(\text{John}_1)\neq
\varphi(\text{John}_2)$. So repetitions can be accommodated within
the setting of Theorem~\ref{Pembed}.

A different issue arises in the case of {\em copies}, namely elements
$\alpha_\ell, \alpha_{\ell'}$ that are actually the same. In the linguistics
literature one expresses this in terms of a FormCopy function that
has the effect we described above, to restrict two different coordinates 
$\alpha_\ell, \alpha_{\ell'}$ with $\ell\neq \ell'$ to lie on the diagonal where
$\alpha_\ell = \alpha_{\ell'}$. This occurs in important situations in 
syntax, such as Obligatory Control, Parasitic Gap, Across-the-Board movement.
With the mathematical model of \cite{MCB}, it does {\em not} occur in 
movement by Internal Merge, because in the coproduct terms 
of the form $T_v \otimes T/T_v$ the extraction of the accessible term $T_v$
is accompanied by the deletion of the deeper copy in $T/T_v$ so no
FormCopy is required. However, it does occur in the other mentioned
cases, for example for Obligatory Control, as discussed in \S 3.8.2 of \cite{MCB}. 

In the mathematical formulation of these cases involving
FormCopy, one performs an identification of the copies,
and the coproduct then acts as the resulting coproduct of
a Hopf algebra of graphs, rather than as the coproduct on
trees by extracting and quotienting the identified terms 
as a single subgraph. In particular, this identification means
that, in such cases, we are embedding graphs obtained
from trees by performing these identifications. 

In the setting considered here, the problematic cases are
functions 
\[
 \varphi : X \to \Delta_{\ell,\ell'} \subset \mathbb{R}^n 
\]
where 
$\Delta_{\ell,\ell'} =\{ (x_1,\cdots, x_n) \in \mathbb{R}^n \,|\, x_\ell=x_{\ell'} \}$
is the diagonal, so that $\varphi_\ell(t)=\varphi_{\ell'}(t)$ for all $t\in X$,
or more generally functions
\[
 \varphi : X \to \Delta_{\ell_1,\cdots, \ell_k} \subset \mathbb{R}^n 
\]
for one of the deeper diagonals 
\[
 \Delta_{\ell_1,\cdots, \ell_k}=\{ (x_1,\cdots, x_n) \in \mathbb{R}^n \,|\, x_{i_1}=\cdots = x_{i_k} \}\, . 
 \]
We can think of $\mathbb{R}^n$ as parameterizing the position of an ordered
set of $n$ points in $\mathbb{R}$, and the diagonals as the loci where a certain number
of these points collide together. 

A way to handle these cases within the formalism considered here,
would be to replace these functions with functions that contain more
information about such colliding points. This can be done, for example
by resolving two coincident points as two colliding points that are
infinitesimally displaced along a certain direction. This extra datum
can be seen as a choice of a way to push the colliding points apart.

This idea can be formalized geometrically. There is a mathematical
construction that is precisely designed for the purpose of resolving
the diagonals in this way in configurations of $n$ points in an
ambient space $Y$. This is known as the Fulton--MacPherson compactification 
of the configuration space and it is constructed algebro-geometrically, \cite{FM}. 
Given a smooth projective algebraic variety $Y$ (in our case here it would be
just the projective line $Y=\mathbb{P}^1$) one considers the complement of the diagonals
\[ 
\Conf_n(Y)= Y^n \smallsetminus \bigcup_{\ell,\ell'} \Delta_{\ell,\ell'} 
\]
and a compactification 
\[
 Y[n] = \overline{\Conf_n(Y)}
\]
of this space, obtained as a sequence of blowups of diagonals in $Y^n$. 
There is a projection map $\pi_n: Y[n]  \twoheadrightarrow Y$. 
Under this projection, the preimage $\pi_n^{-1}\Conf_n(Y)\cong \Conf_n(Y)$ so
we can view the configuration space as sitting inside $Y[n]$ and the complement
$Y[n] \smallsetminus \Conf_n(Y)$ is a normal crossings divisor (a union of
hypersurfaces intersecting transversely). These hypersurfaces and their
intersections are the boundary strata of the compactification $Y[n]$ of $\Conf_n(Y)$.
The details of this construction can be found in \cite{FM} and are not needed here.
A more differential geometric formulation of these compactifications is
described in \cite{BottTaubes}.
In our setting we consider this space for the real projective line $Y=\mathbb{P}^1(\mathbb{R})$.

Intuitively, the effect of this compactification is to replace all the diagonals
where two or more of the coordinates agree, with a projective bundle
where the projective spaces that replace each point of the diagonal 
represent the different ``directions'' with which the colliding
points approach the diagonal. For some considerations about 
working with real rather than complex spaces see for instance
the discussion in \cite{VoronFM}. 
These compactifications also 
have a description in terms of operads, see \cite{Markl}. 

Thus, given maps as above, with
\[
\varphi : X \to \Delta_{\ell,\ell'} \hookrightarrow \mathbb{R}^n \hookrightarrow \mathbb{P}^1(\mathbb{R})^n  
\]
or similarly for more general diagonals, we can choose a local section
$\sigma:  \mathbb{P}^1(\mathbb{R})^n \to \mathbb{P}^1(\mathbb{R})[n]$ with $\pi_n\circ \sigma= \id$, over the domain
$\Dom(\sigma)=\mathbb{R}^n \hookrightarrow \mathbb{P}^1(\mathbb{R})^n$. This choice provides us with
a lift
\begin{equation}\label{sigmalift}
	\xymatrix{ & & \mathbb{P}^1(\mathbb{R})[n] \ar[d]^{\pi_n} \\ X \ar[r]_{\varphi} \ar[urr]^{\sigma\circ\varphi} & \mathbb{R}^n \ar@{^{(}->}[r] &  \mathbb{P}^1(\mathbb{R})^n }
\end{equation} 
where the previously colliding $\varphi_\ell = \varphi_{\ell'}$ now become
separated along the direction specified by the section $\sigma$.

Thus, in this model we can think of any FormCopy operation that is performed on
a syntactic object, which causes some identifications $\alpha_\ell =\alpha_{\ell'}$
as consisting not just of the restriction to diagonals $\Delta_{\ell,\ell'}$
(as formulated in \S 3.8.2 of \cite{MCB})
but as incorporating the additional datum of the section $\sigma$, which
lifts $\Delta_{\ell,\ell'}$ to $\sigma(\Delta_{\ell,\ell'}) \subset \mathbb{P}^1(\mathbb{R})[n]$,
so that it results in a restriction to the boundary divisors of the
Fulton--MacPherson compactification of $\Conf_n(\mathbb{R})$, where points do
not collide but maintain mutual $\epsilon$-displacements (with $\epsilon\to 0$)
in a direction specified by $\sigma$.

\subsection{Size constraints and performance} \label{SizeSec} 

We can then also address the problem of the bounded size
$\# L \leq \# \mathcal{S}\mathcal{O}_0$ assumed in Theorem~\ref{Pembed}.
One of the main properties of the computational structure
of syntax is that it generates a countably infinite set of
possible structures (syntactic objects) out of a finite lexicon.
While there are performance constraints in our brains,
we immediately understand that certain
recursions in syntax exist and would produce syntactically
valid sentences of arbitrarily large size. (For example, in
English one can ``help someone help someone help someone $\cdots$'')
and while we will never in practice use more than a couple of iterations,
we all immediately understand that an arbitrary number of them would
remain both grammatically correct and meaningful.) The difference
between the countably infinite set of viable syntactic objects and the
practical limitations on their use is referred to in the linguistics literature
as the competence-performance distinction, where competence refers to
our internal knowledge of abstract syntax rules and performance takes
into account practical memory limitations and other such non-linguistic factors.
Since we are focusing on the question of realizability of the computational
structure of syntax itself, we need to take into account this distinction
and keep track of whether and to what extent, assumptions that account
for performance constraints would affect a computational model we
want to construct that would capture competence. In an actual
neurocomputational setting, performance constraints would 
necessarily have to be taken into account. In the model we
present here, we can pinpoint where this issue manifests itself, which
is explained in this section as the fact that there is a non-uniform 
upper bound that one loses control of when a size cutoff representing
the performance constraints is lifted.

The size cap $\# L \leq \# \mathcal{S}\mathcal{O}_0$ was imposed by the necessity to use only
assignments of lexical items and features at the leaves with
$\alpha_\ell \neq \alpha_{\ell'}$ for $\ell\neq \ell'$ in $L$.
That problem can be bypassed, as we've been arguing
above by resolving the diagonals to the boundary divisors of
the Fulton--MacPherson compactification. 

However, there is a way in which one does see the performance
problem arise in this representation. It is in the use of uniform
bounds and the high-temperature regime discussed in \S \ref{HighTSec}.

As we have seen in \S \ref{HighTSec}, for $X$ a compact
manifold, one can find a uniform bound for the norm
\[
 \sup_{\alpha\in \mathcal{S}\mathcal{O}_0} \sup_{t\in X} | \varphi(\alpha)(t) |  \leq  M_{X,\mathcal{S}\mathcal{O}_0} < \infty .
\]
The existence of this bound ensures that there is a high temperature range
in which the function $u(t)$ of \eqref{ut} satisfies
$\sup_{t\in X} |u(t)| < 1$, so that the optimization value
$\lambda_{\min}( \varphi(\alpha)(t) , \varphi(\alpha')(t)  )$ used in computing
$\varphi(\alpha)(t) \oplus_{S,\beta} \varphi(\alpha')(t)$ of \eqref{lambdaxyu} is contained in $(0,1)$. 
The fact that the minimum occurs in the interior of the interval guarantees that
the resulting $\varphi(\alpha)(t) \oplus_{S,\beta} \varphi(\alpha')(t)$ still contains
the information of both $\varphi(\alpha)(t)$ and $\varphi(\alpha')(t)$.

When we consider iterated applications of $\oplus_{S,\beta}$ rather than a single one,
in the computation of $\varphi_{S,\beta}(T)$, 
at each step the optimal $\lambda_{\min}(x,y)$ of \eqref{lambdaxyu} is computed for
more general $x=\varphi_{S,\beta}(T_1)$ and $y=\varphi_{S,\beta}(T_1)$ for subtrees
$T_1,T_2$ of $T$. Thus, the condition that
\[
  \lambda_{\min}( \varphi_{S,\beta}(T_1)(t) , \varphi_{S,\beta}(T_1)(t)  )\in (0,1) 
\]
now depends on the condition that
\[
 u(t) = \frac{\beta}{2} ( \varphi_{S,\beta}(T_1)(t) - \varphi_{S,\beta}(T_1)(t) ) 
\]
satisfies $\sup_{t\in X} |u(t)|<1$. This no longer follows directly from the
fact that the functions $\varphi(\alpha)$ satisfy the uniform bound. For each
size $n=\# L$, one does have a uniform bound
\[
 \sup_{t\in X} | \varphi_{S,\beta}(T)(t) | \leq M_{X,\mathcal{S}\mathcal{O},n} < \infty 
\]
for all $T\in \mathfrak{T}_{\mathcal{S}\mathcal{O}_0}$ with $\# L(T)\leq n$.
However, for $n\to \infty$ one cannot guarantee a uniform
bound, and typically one expects that $M_{X,\mathcal{S}\mathcal{O},n}\to \infty$ as $n \to \infty$.
This in turn implies that, as larger and larger sizes of syntactic objects are
considered, the high-temperature regime
\[
 \beta < \beta_{X,\mathcal{S}\mathcal{O},n}\coloneqq M_{X,\mathcal{S}\mathcal{O},n}^{-1} 
\]
will shrink as $\beta_{X,\mathcal{S}\mathcal{O},n} \to 0$ for $n\to \infty$. 
This need to progressively shrink $\beta$ (increasing the temperature
parameter $\beta^{-1}$) signals what can be regarded as accounting,
in this model, for a loss of performance in  handling increasingly large
syntactic objects, in addition to the accumulating computational costs
of performing the iterated operations $\oplus_{S,\beta}$.

\section{The magma operation} \label{magmaSec}

Next we can show that, not only the mapping of syntactic objects to a wavelet encoding in the space
$\mathcal{F}_{S,\beta}$ is faithful (at least using for $S$ the second R\'enyi entropy) but it also is compatible
with the magma operation (for more general choices of $S$). Namely, the magma operation on
syntactic objects maps to the semiring addition in $\mathcal{F}_{S,\beta}$. 

\begin{theorem}\label{magmahom}
	Consider a choice of information function $S$ such that the addition $\oplus_{S,\beta}$  of the 
	thermodynamic semiring $\mathbb{T}_{S,\beta}$ is
	commutative but nonassociative (for example the R\'enyi entropy considered above), hence so is $\mathcal{F}_{S,\beta}$.  
	The map $\varphi_{S,\beta}: \mathfrak{T}_{\mathcal{S}\mathcal{O}_0} \to \mathcal{F}_{S,\beta}$ as in \eqref{phiSbeta}, 
	for $\Lambda$ satisfying the optimal \eqref{argmin}, extended as discussed in \S \ref{RepCopSec},
	is a morphism of commutative nonassociative magmas.
\end{theorem}

\begin{proof} The deformed addition $\oplus_{S,\beta}$ gives $\mathcal{F}_{S,\beta}$ the structure of commutative
nonassociative magma, which is unital with unit $0$. The unit $1$ (the formal empty tree) of the magma $\mathfrak{T}_{\mathcal{S}\mathcal{O}_0}$
maps to $0$ under $\varphi_{S,\beta}$. To see that the map is compatible with the magma operation, namely
\[
 \varphi_{S,\beta}( \mathfrak{M}(T_1,T_2) ) =  \varphi_{S,\beta}(T_1) \oplus_{S,\beta} \varphi_{S,\beta}(T_2)\, ,  
\]
we need to compare
\[
 \min_{\lambda_v, \Lambda_1, \Lambda_2} \{ \sum_{\ell \in L(T_1)} \lambda_v a_\ell(\Lambda_1) + \sum_{\ell' \in L(T_2)}  (1-\lambda_v) a_{\ell'}(\Lambda_2) - \beta^{-1} S_{ \mathfrak{M}(T_1,T_2) }(\lambda_v a_\ell(\Lambda_1),(1-\lambda_v) a_{\ell'}(\Lambda_2)) 
\} 
\]
with
\begin{multline*}
 \min_{\lambda_v}\{  \lambda_v \min_{\Lambda_1} \{ \sum_{\ell \in L(T_1)} a_\ell(\Lambda_1) -\beta^{-1} S_{T_1}(A(\Lambda_1)) \} \\
 + (1-\lambda_v) \min_{\Lambda_2} \{ \sum_{\ell' \in L(T_2)}  a_{\ell'}(\Lambda_2) - \beta^{-1} S_{T_2}(A(\Lambda_2)) \} - \beta^{-1} S(\lambda_v,1-\lambda_v) \} \, . 
\end{multline*}
The recursive structure of the entropy functionals $S_T$ shows that (Lemma~10.3 of \cite{MarThor})
\[
 S_{ \mathfrak{M}(T_1,T_2) }(\lambda_v a_\ell(\Lambda_1),(1-\lambda_v) a_{\ell'}(\Lambda_2)) =
S(\lambda_v,1-\lambda_v) + \lambda_v S_{T_1}(A(\Lambda_1)) + (1-\lambda_v) S_{T_2}(A(\Lambda_2)) \, . 
\]
The identification of the two expressions above then follows, by observing that one can separate the
minimization with respect to $\Lambda_1$, $\Lambda_2$ and $\lambda_v$. 
\end{proof}

\section{Encoding Merge} \label{IMsec}

The magma operation $\mathfrak{M}(T,T')$ on syntactic objects suffices to account for structure
building by External Merge, but movement  also requires an Internal Merge operation,
which can be formulated as in \cite{MCB} in terms of a Hopf algebra of workspaces where
the coproduct extracts accessible terms $T_v$ inside a syntactic object $T$. The 
Internal Merge then forms the new structure $\mathfrak{M}(T_v, T/T_v)$.

By Theorem~\ref{magmahom} we know that 
\[
\varphi_{S,\beta}(\mathfrak{M}(T_v, T/T_v))=\varphi_{S,\beta}(T_v)\oplus_{S,\beta} \varphi_{S,\beta}(T/T_v)\, , 
\]
where the two terms are given by 
\[
 \varphi_{S,\beta}(T_v)=\mathcal{B}_{T_v, S,\beta}((x_\ell)_{\ell\in L(T_v)}) \ \ \ \text{ and } \ \ \ 
\varphi_{S,\beta}(T/T_v)=\mathcal{B}_{T/T_v, S,\beta}((x_\ell)_{\ell\in L(T/T_v)}). 
\]

Note, however, that this does not provide us with a procedure for computing 
$\varphi_{S,\beta}(T_v)$ from $\varphi_{S,\beta}(T)$ since the addition $\oplus_{S,\beta}$
not being invertible in the semiring, one cannot directly extract  
$\varphi_{S,\beta}(T_v)$ from $\varphi_{S,\beta}(T)$ simply via the semiring operations.
The extraction of accessible terms and their representations
has to come from additional algebraic structure (as we already know, from the
use of Hopf algebras to model this extraction procedure at the level of the
syntactic objects and workspaces in \cite{MCB}).

\subsection{Algebra over an operad}\label{CircuitsSec}

Thus, we first describe what additional algebraic structures we have available on
$\mathcal{F}_{S,\beta}$ and $\varphi_{S,\beta}(\mathfrak{T}_{\mathcal{S}\mathcal{O}_0}) \subset \mathcal{F}_{S,\beta}$. 
A first thing we can observe is that, just like the set of syntactic objects $\mathfrak{T}_{\mathcal{S}\mathcal{O}_0}$
is an algebra over the Merge operad (see \cite{MCB}), its image $\varphi_{S,\beta}(\mathfrak{T}_{\mathcal{S}\mathcal{O}_0})$
also has the same structure. These algebra over an operad structure of syntactic objects
is useful for important purposes, including modeling the assignment of theta roles for
semantics, \cite{MarLar}, and the structure of head, complement, specifiers, and the resulting
nested structure of phases. So it is important that
the same structure is maintained in the representation of syntactic objects in $\mathcal{F}_{S,\beta}$.

\begin{lemma}\label{OperadAlg}
	Let $\mathcal{M}$ denote the Merge operad of nonplanar binary rooted trees (with unlabeled leaves), 
	with the operad compositions $\gamma(T; (T_\ell)_{\ell\in L(T)})$ that
	graft the root vertex of each $T_\ell$ into the leaf $\ell \in L(T)$ of $T$. 
	The image $\varphi_{S,\beta}(\mathfrak{T}_{\mathcal{S}\mathcal{O}_0}) \subset \mathcal{F}_{S,\beta}$
	is an algebra over the operad $\mathcal{M}$.
\end{lemma}

\begin{proof} We consider $T\in \mathcal{M}(n)$ as an operation with $n=\#L(T)$ inputs and one output. 
We write this operation in the form $\mathcal{B}_{T,S,\beta}(x_1,\cdots,x_n)$, where we consider
$x_1, \cdots, x_n$ as input variables in $\mathcal{F}_{S,\beta}$. For example, $x\oplus_{S,\beta} y$ 
is the operation (cherry tree) in $\mathcal{M}(2)$ that takes inputs $x,y \in \mathcal{F}_{S,\beta}$ and provides
an output that is also in $\mathcal{F}_{S,\beta}$. This gives $\mathcal{F}_{S,\beta}$ the structure of an
algebra over the Merge operad $\mathcal{M}$, with 
$\varphi_{S,\beta}(\mathfrak{T}_{\mathcal{S}\mathcal{O}_0}) \subset \mathcal{F}_{S,\beta}$ a subalgebra over the same
operad, since for $x_i=\varphi_{S,\beta}(T_i)$ the result 
$\gamma(T; (\varphi_{S,\beta}(T_\ell))_{\ell\in L(T)})$ is also in $\varphi_{S,\beta}(\mathfrak{T}_{\mathcal{S}\mathcal{O}_0})$.
\end{proof}

Note that, here, we consider $\mathcal{B}_{T,S,\beta}(x_1,\cdots,x_n)$ as an
operation with input variables $x_i$. It is only when these variables
are specialized to chosen inputs $x_i =\varphi_{S,\beta}(T_i)$ with $T_i\in \mathfrak{T}_{\mathcal{S}\mathcal{O}_0}$
(including the case of $x_i =\varphi_{S,\beta}(\alpha_i)$ with $\alpha_i\in \mathcal{S}\mathcal{O}_0$),
that $\mathcal{B}_{T,S,\beta}(\varphi_{S,\beta}(T_1),\cdots, \varphi_{S,\beta}(T_n))$ is actually
evaluated by computing the optimizing $A(\Lambda)$ as in \eqref{argmin}.

Here one should interpret the operations $\mathcal{B}_{T,S,\beta}(x_1,\cdots,x_n)$ as {\em circuits}
that input certain waves $x_i=\varphi_{S,\beta}(T_i)$ (functions in $\mathcal{C}(X,\mathbb{R})$) and output 
a now ``combined'' wave form. We know, as discussed above, that these circuits are all
obtained by compositions (according to the structure of the tree $T$) of one simple
{\em gate} that maps a pair of inputs $x,y$ to the output $x\oplus_{S,\beta} y$. As we discuss
in \S \ref{WorkHopfMergeSec} below, we can then think of the Merge action as an action
that {\em transforms these circuits} into new circuits.

Note that the binary gates $(x,y) \mapsto x\oplus_{S,\beta} y$ mentioned here are
to be understood algorithmically as formal primitives and not implementationally
as literal neural components.

\subsection{Workspaces, coproduct, and Merge} \label{WorkHopfMergeSec}

We then discuss how to implement the Hopf algebra structure that gives the extraction of
accessible terms, and the Merge action as a Hopf algebra Markov chain.

\begin{definition}\label{MergeCirc}
	Consider first the set $\mathfrak{B}_{S,\beta}$ of all the operations 
	$\mathcal{B}_{T,S,\beta}: \mathcal{F}_{S,\beta}^{\# L(T)} \to \mathcal{F}_{S,\beta}$.
	We form the vector space $\mathcal{V}_{S,\beta}=\Span_\mathbb{R} (\mathfrak{B}_{S,\beta})$
	spanned by this set. We then form the symmetric algebra over this
	vector space, which we can identify with the polynomial algebra
	in the $\mathfrak{B}_{S,\beta}$,
	\[
	 \mathcal{W}_{S,\beta}\coloneqq\Sym(\mathcal{V}_{S,\beta}) = \mathbb{R} [ \mathfrak{B}_{S,\beta} ] \, . 
	 \]
	We refer to the set $\mathfrak{B}_{S,\beta}$ as the {\em set of thermodynamic syntax circuits}
	with thermodynamic datum $(S,\beta)$, 
	and to $\mathcal{W}_{S,\beta}$ as the {\em space of thermodynamic syntax circuits}.
\end{definition}

\begin{theorem}\label{HopfCirc}
	The space $\mathcal{W}_{S,\beta}$ of thermodynamic syntax circuits has a bialgebra (Hopf algebra) 
	structure that is isomorphic to the one defined on workspaces. The Merge action on workspaces
	is faithfully realized by a Hopf algebra Markov chain in $\mathcal{W}_{S,\beta}$.
\end{theorem}

\begin{proof} 
For a forest $F=T_1 \sqcup \cdots \sqcup T_N$, we write
\[
 \mathcal{B}_{F,S,\beta}\coloneqq \mathcal{B}_{T_1 ,S,\beta}\cdots \mathcal{B}_{T_N,S,\beta} 
 \]
by interpreting the monomial as a basis element in $\mathcal{W}_{S\,\beta}$. 
The elements $\mathcal{B}_{F,S,\beta}$ are the representations of workspaces $F$.

It is then possible to endow $\mathcal{W}_{S,\beta}$ with a coproduct
and a bialgebra structure (or Hopf algebra, depending on slightly 
different forms of the coproduct, as discussed in \S 1.s of \cite{MCB}) 
using the same coproduct defined on workspaces in \cite{MCB}, by setting
\begin{equation}\label{coprod}
	\Delta(\mathcal{B}_{F,S,\beta})=\mathcal{B}_{F,S,\beta}\otimes 1 + 1 \otimes \mathcal{B}_{F,S,\beta} +\sum_{F_{\underline{v}}}
	\mathcal{B}_{F_{\underline{v}},S,\beta}\otimes \mathcal{B}_{F/F_{\underline{v}},S,\beta} \, . 
\end{equation}
We then represent the Merge action in the form
\begin{equation}\label{MSbeta}
	\mathfrak{M}^{(S,\beta)}_{T,T'} = \mu \circ ( (- \oplus_{S,\beta} - )\otimes \id ) \circ \delta_{T,T'} \circ \Delta \, ,
\end{equation}
where $\mu$ is the multiplication in $\mathbb{R} [ \mathfrak{B}_{S,\beta} ]$ and $\delta_{T,T'}$ is the identity on
terms of the coproduct of the form $\mathcal{B}_{T,S,\beta}\mathcal{B}_{T',S,\beta} \otimes \mathcal{B}_{F',S,\beta}$
and zero on all other terms. The notation
$(- \oplus_{S,\beta} - )$ stands for the binary operation $\oplus_{S,\beta}$ that takes input
$\mathcal{B}_{T_1,S,\beta}\mathcal{B}_{T_2,S,\beta}$ and computes
$\mathcal{B}_{T_1,S,\beta}(x_1,\cdots, x_n)\oplus_{S,\beta}\mathcal{B}_{T_2,S,\beta}(y_1,\cdots, y_m)$. 
These Merge operations can be assembled into a Hopf algebra Markov chain (as discussed
in \S 1.9 of \cite{MCB}) of the form 
\begin{equation}\label{MergeSbeta}
	\mathcal{K} =  \mu \circ ( (- \oplus_{S,\beta} - )\otimes \id ) \circ \Pi_{(2)} \circ \Delta \, ,
\end{equation}
where $\Pi_{(2)}$ is the
projection of $\mathcal{W}_{S,\beta}\otimes \mathcal{W}_{S,\beta}$ onto the subspace spanned
by elements of the form $\mathcal{B}_{T_1,S,\beta}\mathcal{B}_{T_2,S,\beta} \otimes \mathcal{B}_{F',S,\beta}$ 
for some arbitrary forest $F'$ and arbitrary trees $T_1,T_2$. 
\end{proof}

\section{Realization as phase synchronizations} \label{ROSEsec}

Our goal in this section is to make the previous discussion slightly more
concrete, by showing that a form of the embodiment of Merge on a
nonassociative commutative thermodynamic semiring addition can
be implemented on an especially simple and explicit class of wavelets,
in the form of phase synchronizations on superpositions of 
sinusoidal waves. 

This very simple realization is again in many ways an abstract toy model,
but it has the advantage of dealing with objects (sinusoidal waves) and
operations (phase locking, cross-frequency synchronization, phase-amplitude coupling)
that are widely used in the neurocomputational context. 
Oscillations are a way of describing neural activities, and modulations of oscillatory activity 
can be measured for perceptual and cognitive functions, linguistic and otherwise. 

A large amount of work in language neuroscience has pointed to phase synchronization, 
phase coupling, phase-amplitude coupling (PAC) being modulated in speech 
and language processing, 
\cite{Baasti1,Baasti2,BreMartin,CoopMartin,CoopMartin2,Ding,Kau,Martin1,Meyer1,Meyer2,Molinaro,Oever,Poe,Prysta,Rimme,Sega,Slaats,Zhao1,Zhao2}.
From the theoretical linguistics perspective, it is natural to expect to see a signature of the
hierarchical structure formation of syntax reflected in such operations on wavelets, if these
are to be involved in sentence formation and parsing activities. However, 
theoretical models of language have not predicted any specific mechanisms
through which such a signature could be realized. 
Moreover, a lot of the neuroscience literature that deals with PAC and language focuses on PAC effects for speech listening.
Empirical results linking PAC and phase synchronization to syntax include
\cite{BaiMartin,BreMartin,WeissMartin}, while empirical results on the relation 
between synchronization and syntax are discussed in \cite{MeyerGum}.
It is in general a very difficult problem to directly relate observable wavelet data 
from oscillations to an underlying algorithmic computational model embodied and
implemented in neural population activity. In this sense the realization we have
described of Merge as a semiring addition on a space of wavelets is   
limited, as it does not address how these operations would be encoded in receptive 
fields and population activity. 

Our mathematical model here is not yet designed, at this stage, to fit any particular theoretical
model of language neuroscience, or proposal for the interpretation of the relation between
syntax and brain waves, with synchronization and modulation operations. 

We will, however, recall briefly some more specific examples of recent work that shares 
some conceptual aspects with the model we will describe in this section, based on 
an operation of synchronization.

\subsection{Martin's compositional neural architecture proposal}\label{MartinSec}

In \cite{MartinDou1,MartinDou2,MartinDou3}, a computational model
of low frequency brain oscillations generating linguistic structure was developed, and
a related theoretical framework was developed in \cite{Martin}. The model is centered
on time-based binding operations, which means that in a neural ensemble the relative
time of firing can encode the relation between different patterns in the system, with
temporal proximity and phase synchronization reflecting relations of word level inputs into
higher hierarchical structures. This process of structure building or parsing can be
modeled by a network where nodes at a higher level show more phase synchronization
encoding structural relations in phrases and sentences. This ``rhythmic computation''
model uses time and synchronization/desynchronization to carry information about
hierarchical structures in language. In this model a phrase or sentence can be formed,
starting from a vector representation of the input words via a conjoining operation
resulting in a conjunctive code on a higher level of the network. In particular, it is 
shown in \cite{BaiMartin} that the level of synchronization and power connectivity
is higher for sentences than for phrases, for comparable semantic and statistical
properties. The precise nature
of these synchronization operations and whether they satisfy the same
algebraic properties (commutativity and nonassociativity)  
of the syntactic Merge are not explicitly discussed in \cite{Martin}, but the kind of
circuits that we discussed in \S \ref{CircuitsSec} above are in principle compatible
with this kind of model. Other aspects of this model include the interplay between 
slower rhythm oscillations and higher frequencies providing a modulation as part
of the generation of higher-level linguistic structures. 

An interesting observation 
made in \cite{Martin} on the nature of the structure building operations is that
words and phrases compose additively and not through a multiplicative
operation (like a tensor product). It is posed as a question there whether
vector addition (linear or convex combinations) would be sufficient. What we
have shown in this paper confirms that the Merge operation on vector representations
(in a vector space of functions) of lexical items is indeed not a tensor product
type operation but an addition-type operation. 
However, we have also shown here that simple vector addition, as in convex combinations,
is not sufficient, as we discussed in \S \ref{CombOp}. On the other hand,  
a modification of a convex combination of vectors, performed via an optimization
of such combinations involving an entropy functional, provides an
additive operation (the addition in a thermodynamic semiring) that has all the
right algebraic properties required for compositionality in natural language.
In this section we will also show that such an operation is implementable
in the form of a synchronization.

\subsection{Murphy's ROSE proposal}\label{caseROSEsec}

In  \cite{Murphy2,Murphy4} Murphy proposes a possible theoretical model, called ROSE, for 
neurocomputational realizations of Merge, based on neurosurgery experiments measuring 
phase shifts towards increased phase synchronization and modulation effects of brain waves.
In essence, Murphy's ROSE model (Representation, Operation, Structure, Encoding)
proposes a multi-step process, with a first process of spike-phase coupling that
bundles atomic features together (R and O part), followed by phase-coupling
of a slower theta rhythm with this morphological spike-phase complex (implementing
lexical access and memory) where spikes (encoding features) couple with gamma phases 
(encoding morphemes) and the amplitude of this gamma wave couples to a theta phase 
(encoding lexical items). In this part of the process (S and E part) lexical items are encoded
as theta-gamma-spike synchronized complexes, spatially separated as different cellular 
clusters that oscillate in the theta range.
Structure formation in syntax is then modeled as 
coupling to delta waves, which involves a delta phase-phase locking for merging, while 
delta phase-locking with the amplitude of both  these theta rhythms implements
the presence of a head function for syntactic objects.

For the purposes of the present paper, we do not analyze these models in any detail,
as we are not trying to formalize or validate any specific existing model. The main reason
why we mention them here is because of one particular aspect, that is directly relevant to
the discussion of the previous sections, namely the fact that both of these approaches develop 
a modeling of hierarchical structure building that involves phase synchronization and 
cross-frequency coupling. This kind of approach and results as one sees in these types
of models is what motivates and inspires the construction that we describe in this section.

Indeed, our goal here is to show that, in our mathematical model, if lexical items are encoded 
through simple sinusoidal waves, then a model of Merge as discussed in the previous 
sections can be implemented by synchronizing phases via cross-frequency synchronization, where the phases are
locked to a common weighted average expressed as a sum $\oplus_{\Ry_2,\beta}$.  
The operation we describe is different from  what is proposed in \cite{Murphy2,Murphy4}, 
where it is suggested that a mechanism for bracketing would rely on time-ordering, 
but a direct algebraic verification of nonassociativity is not directly available.  In the construction we
discuss here nonassociativity is realized implementing a phase synchronization via semiring addition.

\subsection{The successor function in thermodynamic semirings} \label{succSec}

As shown in \S 9 of \cite{MarThor}, the algebraic properties of the thermodynamic semiring
$\mathbb{T}_{S,\beta}$ can be encoded in its {\em successor function} (the Legendre transform of
the entropy functional $\beta^{-1} S$). It is defined as
\begin{equation}\label{succS}
	\Upsilon(x,\beta)= x \oplus_{S,\beta} 0 = \min_{\lambda\in [0,1]} \{ \lambda x - \beta^{-1} S(\lambda) \} \, .
\end{equation}
It is called ``successor function'' since it is the operation of adding (with the semiring addition $\oplus_{S,\beta}$)
a copy of the multiplicative unit (which is $0$ in the semiring multiplication $\odot$), just as the usual successor
function of arithmetic $x\mapsto x+1$ is the operation of adding the multiplicative unit.

\begin{figure}[h]
	\begin{center}
		\includegraphics[scale=0.35]{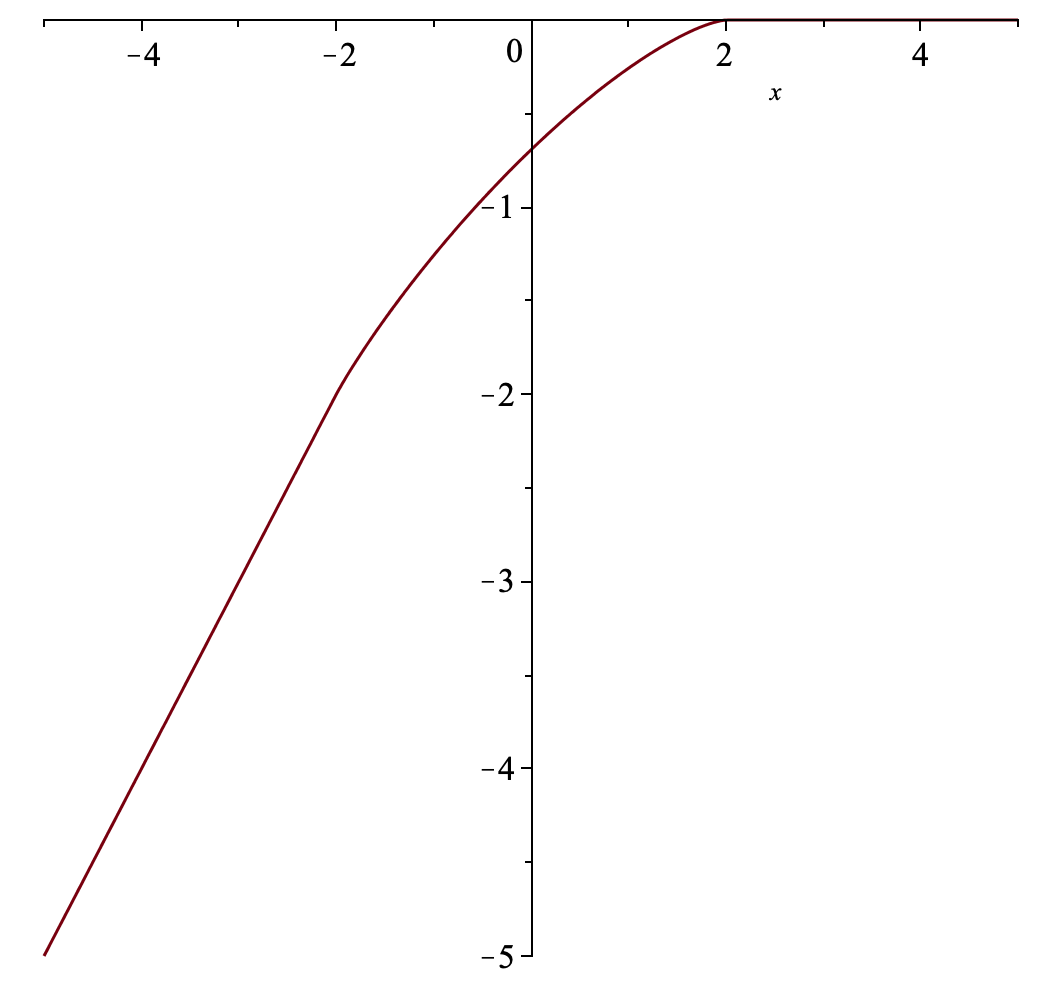} 
		\caption{The behavior of the successor function $\Upsilon(x,\beta)$ for $S=\Ry_2$, shown in the case of $\beta=1$.
			\label{FigSucc}}
	\end{center}
\end{figure}

\begin{figure}[h]
	\begin{center}
		\includegraphics[scale=0.35]{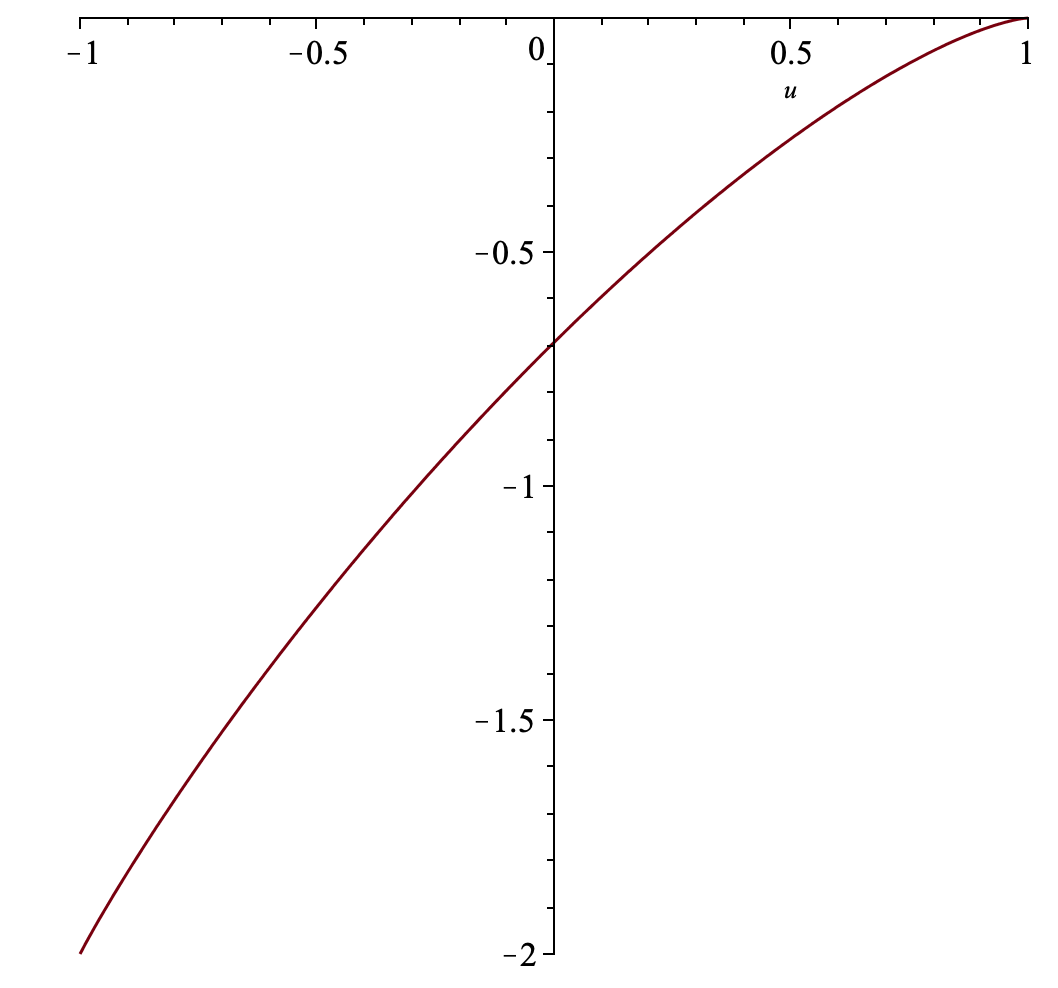} 
		\caption{The successor function $\Upsilon(x,\beta)$ for $S=\Ry_2$ and $\beta=1$, in the
			interval $|u|<1$ for $u=\beta x/2$.
			\label{FigSuccu}}
	\end{center}
\end{figure}

The behavior in the case of
$S=\Ry_2$ is shown in Figure~\ref{FigSucc}. The property that $\Upsilon(x,\beta)\sim x$ for $x\to -\infty$
and $\Upsilon(x,\beta)\sim 0$ for $x\to +\infty$ is common to all thermodynamic semirings: these properties
are in fact related to the existence of an additive unit in the semiring (Proposition~9.1 of \cite{MarThor}).
In the specific case of $S=\Ry_2$, we obtain from the minimum $\lambda_{\min}(u)$ in the form \eqref{lambdaxyu2}
that in fact $\Upsilon(x,\beta)= x$ for
all $x< -2\beta^{-1}$ and $\Upsilon(x,\beta)=0$ for all $x> 2\beta^{-1}$ and that in the interval 
$|x|< 2\beta^{-1}$ it is given by the function
\begin{equation}\label{eqSuccu}
	\Upsilon(x,\beta)=\beta^{-1}  \left(  -1 +u + \sqrt{1-u^2} - \log\left(1+ \sqrt{1-u^2}  \right) \right) 
\end{equation} 
for $u=\beta x/2$, with $| u  |\leq 1$, and $0$ for $u>1$ and $x$ for $u<-1$ which is shown in Figure~\ref{FigSuccu}. (While this is less evident
in this plot than in the one of Figure~\ref{FigLambdaMin}, the function \eqref{eqSuccu} is only
piecewise smooth, with singularities at $u=\pm 1$.)

A direct computation from \eqref{eqSuccu} with $u=\beta x/2$ gives 
the following property of the successor function, which will be useful 
in implementing Minimal Search, as we discuss in \S \ref{AccTermsSec}.

\hspace*{12pt}
\begin{lemma}\label{lemYbeta}
	The successor function $\Upsilon(x,\beta)$  with $S=\Ry_2$ has the following expansion in $\beta$
	\begin{equation}\label{expandYbetaall}
		\Upsilon(x,\beta) \sim -\log(2) \beta^{-1} + \frac{x}{2} - \sum_{n\geq 1} \frac{C_{n-1}\, x^{2n}}{n\, 2^{4n}} \, \beta^{2n-1} \, ,
	\end{equation}	
	where $C_k$ are the Catalan numbers
	\[
	 C_k = \frac{1}{k+1} \binom{2k}{k}  \, . 
	 \]
\end{lemma}

\begin{proof} 
Consider the function
\begin{equation}\label{hu}
 h(u)=\sqrt{1-u^2}-1
 -\log\left(\frac{1+\sqrt{1-u^2}}2\right).
\end{equation}
It satisfies $h(0)=0$ and
\[
 h'(u)=-\frac{u}{1+\sqrt{1-u^2}}
 =-\frac{1-\sqrt{1-u^2}}u.
\]
The generating series of the Catalan numbers is given by
\[
 1-\sqrt{1-u^2} =\sum_{n\geq1}\frac{C_{n-1}}{2^{2n-1}}u^{2n}.
\]
This gives the series expansion of $h'(u)$, and integrating it 
term by term, for $|u|<1$, we obtain 
\begin{equation}\label{huser}
 h(u)=-\sum_{n\geq1}\frac{C_{n-1}}{n\,4^n}u^{2n} \, .
\end{equation}
Then using
\[
 \Upsilon(x,\beta) = \beta^{-1} ( -\log(2) + u + h(u)) \ \ \ \text{ with } u=\beta x/2 \, , 
 \]
gives the expansion \eqref{expandYbetaall}.
\end{proof}

In particular, this gives
	\begin{equation}\label{expandYbeta}
		\Upsilon(x,\beta) \sim -\log(2) \beta^{-1} + \frac{x}{2} -\frac{x^2}{16}  \beta  
		- \frac{x^4}{512}  \beta^3  - \frac{x^6}{6144} \beta^5   + O(\beta^6)\, .
	\end{equation}

\begin{remark}\label{distrY}
		The distributivity of the multiplication $\odot$ over the sum
		$\oplus_{S,\beta}$, in a  thermodynamic semiring $\mathbb{T}_{S,\beta}$, namely the property
		\begin{align*}
		 z \odot (x\oplus_{S,\beta} y) & = z+ \min_\lambda \{ \lambda x+ (1-\lambda) y - \beta^{-1} S(\lambda) \}  \\
		  & = \min_\lambda \{ z+ \lambda x+ (1-\lambda) y - \beta^{-1} S(\lambda) \} \\
		& = \min_\lambda \{  \lambda (x+z)+ (1-\lambda) (y+z) - \beta^{-1} S(\lambda) \} \\
		  &= (x+z) \oplus_{S,\beta} (y+z) = (x\odot z) \oplus_{S,\beta} (y\odot z) \, ,  
		\end{align*}  
		implies that the addition $x\oplus_{S,\beta} y$ is completely determined by the successor function via
		\[
		 x\oplus_{S,\beta} y =\Upsilon(x-y,\beta) + y \, . 
		 \]
		%(We changed slightly the notation here with respect to \cite{MarThor}, where the successor is called
		%$\lambda$ instead of $\Upsilon$ to avoid clashes of notation with our use of $\lambda$ in this paper
		%as the probabilities at the nodes.)
\end{remark}

We can then provide a model of Merge as phase-synchronization of waves based on the
construction of the previous sections and the properties of the successor function.

\subsection{Merge represented as phase synchronization} 

Here we assume, as in the rest of this paper, that there is a representation
\[
 \varphi: \mathcal{S}\mathcal{O}_0 \to \mathcal{F}=\mathcal{C}^\infty (X,\mathbb{R}) 
 \]
of lexical items and syntactic features. 

These functions $\varphi(\alpha)=\varphi_\alpha(t,x,y,z)$ can be thought of, in first
approximation, as sinusoidal waves $\varphi_\alpha(t,x,y,z)= A_\alpha\sin(\nu_\alpha t +\omega_\alpha)$
where $A_\alpha=A_\alpha(x,y,z)$ is the amplitude, $\nu_\alpha=\nu_\alpha(x,y,z)$ is the frequency, and
$\omega_\alpha=\omega_\alpha(x,y,z)$ is the phase.  This assumption is not too unrealistic, as superpositions of a   	
few pure sinusoidal waves may in fact be adequate for word representation, \cite{Suppes}.

We also do not discuss here the dependence
of amplitude, frequency and phase on the spatial location: 
the variability may be encoded as a statistical model. This is
adequate since the transversality arguments we use rely on
a generic case (dense open set).
We do assume, however, that the variation is sufficiently
regular (at least continuous and piecewise smooth for the resulting representations of
syntactic objects, and smooth for the waves $\varphi_\alpha$ representing lexical items
and features). 

\subsubsection{Phase synchronization}

Consider first the case of two waves with the same frequency $\nu_{\alpha_1}=\nu_{\alpha_2}$.
In this case, phase-synchronization simply means that the phases are shifted so that the
phase difference 
\[
 \Delta\omega_{\alpha_1\alpha_2}=\omega_{\alpha_1}-\omega_{\alpha_2} 
 \]
becomes zero. Such a phase shift could be achieved by an effect that moves both phases towards
an average of the two. In view of the discussion on the algebraic properties of such averaging
operations, one can instead consider an averaging that is entropy-optimized, as discussed above,
so that $\omega_{\alpha_1}$ and $\omega_{\alpha_2}$ are replaced by a common synchronized phase
\begin{equation}\label{synchphases}
	\omega_{\alpha_1} \oplus_{\Ry_2, \beta} \omega_{\alpha_2} \, . 
\end{equation} 
Since in general one can evaluate {\em phase differences} rather than phases, one can
equivalently write the phase synchronization \eqref{synchphases} in terms of the 
successor function. 
%We can assume that, say, $\omega_{\alpha_2} > \omega_{\alpha_1}$ 
%(the opposite case is analogous by commutativity of the $\oplus_{\Ry_2, \beta}$ operation). 
We can write \eqref{synchphases} as 
\begin{equation}\label{synchphasesY}
	\omega_{\alpha_1} \oplus_{\Ry_2, \beta} \omega_{\alpha_2} = \omega_{\alpha_2} + \Upsilon( \omega_{\alpha_1}-\omega_{\alpha_2} , \beta) = \omega_{\alpha_2} \odot \Upsilon(\Delta\omega_{\alpha_1\alpha_2},\beta) \, ,
\end{equation}
which expresses the phase synchronization as applying the successor function of the thermodynamic semiring 
to the phase difference. 

\subsubsection{Cross-frequency phase synchronization} 

More generally, for different frequencies $\nu_{\alpha_1}\neq \nu_{\alpha_2}$, one can still
achieve phase synchronization via the mechanism of {\em cross-frequency phase synchronization}.
For this, we assume that the two frequencies are in a rational relation, namely there are relatively 
prime integers $n_{(\alpha_1,\alpha_2)}, n_{(\alpha_2,\alpha_1)}\in \mathbb{N}$ 
with $\gcd (n_{(\alpha_1,\alpha_2)}, n_{(\alpha_2,\alpha_1)})=1$, such that
\begin{equation}\label{ratrelnu}
	n_{(\alpha_1,\alpha_2)} \nu_{\alpha_1} = n_{(\alpha_2,\alpha_1)} \nu_{\alpha_2} \, . 
\end{equation}
In this case the phase difference is measured as
\begin{equation}\label{phasediff}
	\Delta\omega_{\alpha_1\alpha_2} \coloneqq n_{(\alpha_1,\alpha_2)} \omega_{\alpha_1}- n_{(\alpha_2,\alpha_1)} \omega_{\alpha_2} \, .
\end{equation}
We can then again synchronize the phases as above, with the synchronized phase
\begin{align}\nonumber
\omega_{\alpha_1}^{\odot n_{(\alpha_1,\alpha_2)}} \oplus_{\Ry_2, \beta} 
\omega_{\alpha_2}^{\odot n_{(\alpha_2,\alpha_1)}}  
= n_{(\alpha_1,\alpha_2)} \omega_{\alpha_1} \oplus_{\Ry_2, \beta} n_{(\alpha_2,\alpha_1)} \omega_{\alpha_2} \\
\label{synchphasesY2}
	= \omega_{\alpha_2}^{\odot n_{(\alpha_2,\alpha_1)}} \odot \Upsilon(\Delta\omega_{(\alpha_1,\alpha_2)},\beta) \, ,
\end{align}
with the phase difference $\Delta\omega_{\alpha_1\alpha_2}$ defined as in \eqref{phasediff}. 
In the case where the phases (and frequencies and amplitudes) of the sinusoidal waves have
a dependence on the spatial coordinates (with locally constant $n_{\alpha_i}$), the
semiring operations are applied pointwise, as discussed in the previous sections.

To see the action of Merge as phase synchronization, assume that we
start with a packet of wave functions
\[
 \Phi(t) = \sum_\alpha c_\alpha \, \varphi_\alpha(t) 
\]
with coefficients $c_\alpha\in \mathbb{R}_+$ (these can be absorbed into the amplitudes $A_\alpha$), 
and with wave functions $\varphi_\alpha(t)=A_\alpha\sin(\nu_\alpha t +\omega_\alpha)$,
where we leave implicit any possible dependence of $A,\nu,\omega$ on the spatial coordinates. 

The merging $\mathfrak{M}(\alpha_1,\alpha_2)$ then replaces a packet
$\sum_\alpha \varphi_\alpha$ with a new packet in which the contribution 
$\varphi_{\alpha_1}+\varphi_{\alpha_2}$ of the two waves 
\[
 \varphi_{\alpha_i}=A_{\alpha_i}\sin(\nu_{\alpha_i} t +\omega_{\alpha_i}) 
 \]
for $i=1,2$, is replaced, via cross-frequency phase synchronization, by the contribution of the phase shifted waves
\begin{equation}\label{mergewaves}
	\mathfrak{M}(\varphi_{\alpha_1},\varphi_{\alpha_2})=A_{\alpha_1}\sin(\nu_{\alpha_1} t +\omega_{\{\alpha_1,\alpha_2\}})+
	A_{\alpha_2}\sin(\nu_{\alpha_2i} t +\omega_{\{\alpha_1,\alpha_2\}}) 
\end{equation}
where
\[ 
\omega_{\{\alpha_1,\alpha_2\}} \coloneqq \omega_{\alpha_1}^{\odot n_{(\alpha_1,\alpha_2)}} \oplus_{\Ry_2, \beta} \omega_{\alpha_2}^{\odot n_{(\alpha_2,\alpha_1)}} 
\]
is obtained as above from the successor function of the thermodynamic semiring applied to the phase shift of \eqref{phasediff}. 

More generally, one can assume that the $\varphi_\alpha$ are superpositions of a few sinusoidal waves
$\varphi_\alpha=\sum_i A_{\alpha,i} \sin(\nu_{\alpha,i}t + \omega_{\alpha,i})$ and that the Merge
operation gives a superposition of waves of the form
$\varphi_{\mathfrak{M}(\alpha_1,\alpha_2)}=\sum_{ij} ( A_{\alpha_1,i} \sin(\nu_{\alpha_1,i}t + \omega_{\{ (\alpha_1,i), (\alpha_2,j)\}})+
A_{\alpha_1,j} \sin(\nu_{\alpha_1,j}t + \omega_{\{ (\alpha_1,i), (\alpha_2,j)\}}) )$ where the frequencies $\nu_{\alpha_1,i}$
and $\nu_{\alpha_1,j}$ have rational ratios. 
For simplicity of notation we just write the case where the $\varphi_\alpha$ are single sinusoidal waves, as this 
superposition case is analogous.

\begin{proposition}\label{algMphases}
	The operation \eqref{mergewaves} is commutative. It is also nonassociative, for a dense open set 
	of functions $\omega\in \mathcal{C}^\infty(\Omega,\mathbb{R}^N)$,
	with $\Omega\subset \mathbb{R}^3$ the spatial region where the waves are located, and $N=\#\mathcal{S}\mathcal{O}_0$,
	such that $\varphi_\alpha=A_\alpha \sin(\nu_\alpha t +\omega_\alpha)$.
\end{proposition}

\begin{proof}
The commutativity of the operation \eqref{mergewaves} 
follows from the fact that, while $n_{(\alpha_1,\alpha_2)}\neq n_{(\alpha_2,\alpha_1)}$,
exchanging $\varphi_{\alpha_1}$ and $\varphi_{\alpha_2}$ exchanges both $\omega_{\alpha_1}$ and $\omega_{\alpha_2}$
as well as $\nu_{\alpha_1}$ and $\nu_{\alpha_2}$ hence it exchanges also 
$n_{(\alpha_1,\alpha_2)}$ and $n_{(\alpha_2,\alpha_1)}$, so that the result is the same, 
$\omega_{\alpha_1\alpha_2} =\omega_{\alpha_2\alpha_1}$, 
so we obtain as needed that 
\[
 \mathfrak{M}(\varphi_{\alpha_1},\varphi_{\alpha_2}) = \mathfrak{M}(\varphi_{\alpha_2},\varphi_{\alpha_1}) \, . 
\]

The nonassociativity of the operation  \eqref{mergewaves}  lies in the nonassociativity of the iterated phase shifts
\begin{multline*}
 \omega_{\alpha_3}^{\odot n_{(\alpha_3,(\alpha_1,\alpha_2))}} \oplus_{\Ry_2, \beta} ( \omega_{\alpha_1}^{\odot n_{(\alpha_1,\alpha_2)}} \oplus_{\Ry_2, \beta} \omega_{\alpha_2}^{\odot n_{(\alpha_2,\alpha_1)}} )^{\odot n_{((\alpha_1,\alpha_2),\alpha_3)}} \neq  \\
(\omega_{\alpha_3}^{\odot n_{(\alpha_3,\alpha_1)}} \oplus_{\Ry_2, \beta}  \omega_{\alpha_1}^{\odot n_{(\alpha_1,\alpha_3)}})^{\odot n_{((\alpha_3,\alpha_1),\alpha_2)}}  \oplus_{\Ry_2, \beta} \omega_{\alpha_2}^{\odot n_{(\alpha_2,(\alpha_3,\alpha_1))}}  \, , 
\end{multline*}
where the multiplicities are determined by comparing the frequencies
\[
 n_{((\alpha_1,\alpha_2),\alpha_3)}  n_{(\alpha_1,\alpha_2)} \nu_{\alpha_1} =  n_{((\alpha_1,\alpha_2),\alpha_3)}   n_{(\alpha_2,\alpha_1)} \nu_{\alpha_2} = n_{(\alpha_3,(\alpha_1,\alpha_2))} \nu_{\alpha_3} 
\]
and  
\[ n_{((\alpha_3,\alpha_1),\alpha_2)} n_{(\alpha_3,\alpha_1)} \nu_3 = n_{((\alpha_3,\alpha_1),\alpha_2)} n_{(\alpha_1,\alpha_3)} \nu_1 = n_{(\alpha_2, (\alpha_3,\alpha_1))} \nu_2\, . 
\]
This is again obtainable from a transversality argument as in Theorem~\ref{3Tcase}, extended as
in Theorem~\ref{Pembed} to the case of iterations. In this case, the hypersurfaces where transversality
needs to occur are of the form (for the case of Theorem~\ref{3Tcase}) 
\begin{equation}\label{Hnu}
	H_{\underline{n}}=\{ (x_1,x_2,x_3)\in \mathbb{R}^3 \,|\, 
	x_3^{\odot u} \oplus_{\Ry_2, \beta} ( x_1^{\odot a} \oplus_{\Ry_2, \beta} x_2^{\odot b} )^{\odot v} = 
	(x_3^{\odot c} \oplus_{\Ry_2, \beta}  x_1^{\odot d})^{\odot p}  \oplus_{\Ry_2, \beta} x_2^{\odot q} \}
\end{equation}
with $\underline{n}=(n_{ij})_{i\neq j\in \{ 1,2,3 \}}$ satisfying
\begin{equation}\label{nijrels}
	\frac{n_{ij}}{n_{ji}} \frac{n_{jk}}{n_{kj}} = \frac{n_{ik}}{n_{ki}} 
\end{equation} 
for $i\neq j \neq k$, 
and with  $a=n_{12}$, $b=n_{21}$, $c=n_{31}$, $d=n_{13}$, and $u,v \in \mathbb{N}$ and $p,q \in \mathbb{N}$ are, respectively, pairs of relatively 
prime integers determined by the identities
\[
 \frac{v}{u}= \frac{n_{13}}{n_{31} n_{12}} =\frac{n_{23}}{n_{32} n_{21}}      \ \ \ \ \text{ and } \ \ \ \   \frac{p}{q} = \frac{n_{32}}{n_{23} n_{31}}= \frac{n_{12}}{n_{21} n_{13}} \, .  
 \]
While there is a countable set of such $H_{\underline{n}}$, for varying choices of the integers in $\underline{n}$,
in fact only a finite set of $H_{\underline{n}}$ is relevant and is determined by the set of
frequencies $\{ \nu_\alpha \}_{\alpha\in \mathcal{S}\mathcal{O}_0}$ of the functions $\varphi_\alpha$, through
their ratios $\nu_j/\nu_i = n_{ij}/n_{ji}$.  (In the case
where we allow frequencies to vary with the spatial location in the compact region $\Omega$, we assume their ratios are
locally constant, hence in this case too only finitely many $\underline{n}$ are involved.)
In the case of trees $T,T'$ with $T\neq T'$ and with $L=L(T)=L(T')$ with the same assignment $\{ \alpha_\ell \}_{\ell\in L}$
of $\alpha_\ell \in \mathcal{S}\mathcal{O}_0$ at the leaves, we similarly have hypersurfaces
$H_{T,T',\underline{n}}$ for $\underline{n}=( n_{ij} )_{i\neq j\in \{ 1, \cdots, n \}}$ for $n=\# L$, with the relations \eqref{nijrels}.
These hypersurfaces  extend the case of \eqref{Hnu}, by the appropriate modification of \eqref{HTT},
using a version of $\mathcal{B}_{T,S,\beta, \underline{n}}(x_1,\cdots, x_n)$ that incorporates the multiplicities $n_{ij}$.
We do not write these explicitly, but the construction is clear by iterating the expression in \eqref{Hnu}. 
Again, the number of hypersurfaces $H_{T,T',\underline{n}}$ that need to be taken into account for the
transversality argument is finite, since the ratios $n_{ij}/n_{ji}$ correspond to the ratios of frequencies $\nu_j/\nu_i$
in the finite set $\{ \nu_\alpha \}_{\alpha\in \mathcal{S}\mathcal{O}_0}$. 
\end{proof}

In this model, the hierarchical construction of syntactic objects then proceeds in the following way.
Given a finite set $\{ \alpha_\ell \}_{\ell \in L}$
(assuming distinct $\alpha_\ell \neq \alpha_{\ell'}$ for $\ell\neq \ell'$), a syntactic object $T$ with bracket notation 
$\mathcal{B}_T(\alpha_1,\cdots,\alpha_n)$ is mapped to the function 
\begin{equation}\label{phiTomega}
	\varphi_T\coloneqq\sum_{\ell\in L}  \omega_T(\varphi_{\alpha_\ell}) \, , 
\end{equation} 
where $\omega_T$ here denotes the phase shift
\[
 \omega_T(\varphi_{\alpha_\ell})(t)= A_{\alpha_\ell} \sin(\nu_{\alpha_\ell} t + \omega_T) 
\]
where the synchronized phase $\omega_T$ takes the form
\[
 \omega_T= \mathcal{B}_{T,\Ry_2,\beta, \underline{n}}(\omega_{\alpha_1}, \cdots, \omega_{\alpha_n})\, , 
 \]
for $\underline{n}=(n_{\ell \ell'})$ with $n_{\ell \ell'}/n_{\ell' \ell}=\nu_{\alpha_{\ell'}}/\nu_{\alpha_\ell}$.

\subsection{Accessible terms}\label{AccTermsSec}

In addition to realizing the magma of syntactic objects, one needs to be able to identify
accessible terms inside an already formed structure. In other words, one needs a Minimal
Search algorithm for reconstructing the terms $\varphi_{T_v}$ for $T_v \subset T$ an
accessible term, given the wave function $\varphi_T$.

Since $\varphi_T$ is obtained by repeated application of the semiring addition $\oplus_{S,\beta}$
to the phases, to implement cross-frequency phase synchronization, it is necessary to ``undo''
this operation to recover the original wave functions before synchronization. The semiring
addition is not invertible, which seems to be a main obstacle to this operation. However, one can
use some of the properties of the successor function analyzed above to extract the accessible terms. 

We can view this as the question of whether the general result of Theorem~\ref{Pembed} could
be replaced, in the case of the construction of $\varphi_T$ via repeated $\oplus_{S,\beta}$-combination
of phases, by a stronger result, providing an explicit algorithmic way of reconstructing the input
phases $\omega_{\alpha_i}$ and their combinations $\omega_{T_v}$ over accessible terms
$T_v \subset T$ from $\varphi_T$ as a function of $\beta$ (in particular through the
coefficients of the $\beta$-expansion). We show that while this is the case in simple
examples, this kind of reconstruction fails for larger trees. This fact suggests that this simple
model with cross-frequency phase synchronization of sinusoidal waves does not suffice to
capture the full properties of the general model described in the previous sections. On the
other hand, the fact that the first example that fails reconstructibility on a dense open set of 
inputs occurs only at $9$ leaves indicates that this simple sinusoidal wave model remains
valid for trees up to $8$ leaves, a size bound that is sufficient for capturing a wide range 
of syntactic phenomena. 

We first consider the case of a syntactic object obtained from elements of $\mathcal{S}\mathcal{O}_0$ via a single Merge operation, 
namely of the form $T=\mathfrak{M}(\alpha_1,\alpha_2)$. This syntactic
object has only two accessible terms, $\alpha_1$ and $\alpha_2$. We need to show how to extract
them from the wave function $\varphi_T$. 

\begin{lemma}\label{simpleMacc}
	For a syntactic object $T\in \mathcal{S}\mathcal{O}$ of the form $T=\mathfrak{M}(\alpha_1,\alpha_2)$ with $\alpha_1,\alpha_2\in \mathcal{S}\mathcal{O}_0$,
	it is possible to obtain the wave functions $\varphi_{\alpha_1}$ and $\varphi_{\alpha_2}$ from only observing
	the result of Merge $\varphi_T$, provided $\varphi_T$ can be observed at variable values of the parameter $\beta$.
\end{lemma}

\begin{proof}
We know by \eqref{phiTomega} that, in order to be able to access $\varphi_{\alpha_1}$ and $\varphi_{\alpha_2}$
we need to obtain the phases $\omega_{\alpha_1}$ and $\omega_{\alpha_2}$ from the synchronized phase
$\omega_T = \omega_{\alpha_1} \oplus_{\Ry_2,\beta} \omega_{\alpha_2}$. We can assume without
loss of generality that $\omega_{\alpha_2} > \omega_{\alpha_1}$, and we write $\omega_T= \omega_{\alpha_2} + \Upsilon(\omega_{\alpha_1}-\omega_{\alpha_2},\beta)$, with $\omega_{\alpha_1}-\omega_{\alpha_2} <0$. 
Let
\[
 A_{2n} \coloneqq \frac{(\omega_{\alpha_1}-\omega_{\alpha_2})^{2n}}{2^{2n-1}} \, . 
 \]
The $\beta$-expansion of \eqref{expandYbetaall} gives
\[
\omega_T = - \frac{\log(2)}{\beta} + \frac{\omega_{\alpha_1}+\omega_{\alpha_2}}{2} - \sum_{n\geq 1}
\frac{C_{n-1}}{n 2^{2n+1}} \, A_{2n} \, \beta^{2n-1} \, . 
\]
We write the first terms of this expansion as
\[
  \omega_T = - \frac{\log(2)}{\beta} + a_0 + a_1 \beta + O(\beta^3) \, ,
\]
where 
\[
 a_0=\frac{\omega_{\alpha_1}+\omega_{\alpha_2}}{2} \ \ \  \text{ and } \ \ \  a_1=-\frac{(\omega_{\alpha_1}-\omega_{\alpha_2})^2}{16} \, . 
\]
The $\omega_{\alpha_1}$ and $\omega_{\alpha_2}$ are then the two roots of the polynomial equation
\[
(x-a_0)^2 + 4 a_1 =0 \, , 
\]
hence they are reconstructed from $\omega_T$ through the first terms of the $\beta$-expansion.
This shows that, for a single Merge operation, 
$T=\mathfrak{M}(\alpha_1,\alpha_2)$ the wave functions $\varphi_{\alpha_1}$ and $\varphi_{\alpha_2}$ representing
the accessible terms are indeed accessible given the wave function $\varphi_T$.
\end{proof}

The argument of Lemma~\ref{simpleMacc} relies on the properties of the $\beta$-expansion of the
successor function. It does not directly apply to hierarchical structures $T=\mathfrak{M}(T_1,T_2)$
with $\omega_T = \omega_{T_1}\oplus_{\Ry_2,\beta} \omega_{T_2}$, where the
$\omega_{T_1}, \omega_{T_2}$ themselves are iterations of previous $\oplus_{\Ry_2,\beta}$
operations, hence they incorporate a $\beta$-dependence.
Indeed, it is difficult to iterate Lemma~\ref{simpleMacc} to syntactic objects that involve multiple repeated applications
of the magma operation $\mathfrak{M}$, because successive applications of $\mathfrak{M}$ compose the
$\beta$-expansions. One can see this more explicitly by considering the example of a syntactic
object of the form 
\begin{equation}\label{4tree}
	T = \Tree[ [ $\alpha_1$ $\alpha_2$ ] [ $\alpha_3$ $\alpha_4$ ] ] = \Tree[ $T_1$ $T_2$ ] = \mathfrak{M}(T_1,T_2)
\end{equation}
where the corresponding compositions that implement the phase synchronization would be of
the form
\[
 \omega_T = \omega_{T_1} \oplus_{\Ry_2, \beta} \omega_{T_2}\, 
\]
with
\[
 \omega_{T_1}= \omega_{\alpha_1} \oplus_{\Ry_2, \beta} \omega_{\alpha_2} \ \ \ \text{ and } \ \ \
\omega_{T_2}= \omega_{\alpha_3} \oplus_{\Ry_2, \beta} \omega_{\alpha_4} \, . \]
In terms of the successor function, assuming $\omega_{T_2}>\omega_{T_1}$ and $\omega_{\alpha_2}> \omega_{\alpha_1}$
and $\omega_{\alpha_4}> \omega_{\alpha_3}$ (the other cases are analogous using commutativity 
of $ \oplus_{\Ry_2, \beta}$), we have
\begin{align*}
 \omega_T & =  \omega_{T_2} + \Upsilon( \omega_{T_1} - \omega_{T_2}, \beta)  \\
& = \omega_{\alpha_4} + \Upsilon( \omega_{\alpha_3} - \omega_{\alpha_4}, \beta) + 
\Upsilon\big( \omega_{\alpha_2} + \Upsilon( \omega_{\alpha_1} - \omega_{\alpha_2}, \beta)
- \omega_{\alpha_4} - \Upsilon( \omega_{\alpha_3} - \omega_{\alpha_4}, \beta) , \beta\big)\, . 
\end{align*}

\begin{lemma}\label{expand4}
Consider the expressions
\begin{equation}\label{muomegai}
\mu \coloneqq \frac{1}{4} \sum_{i=1}^4 \omega_{\alpha_i}, \ \ \ S_k =\sum_{i=1}^4 (\omega_{\alpha_i} -\mu)^k, \ \ \  Q=( (\omega_{\alpha_1}-\omega_{\alpha_2})^2 - (\omega_{\alpha_3}-\omega_{\alpha_4})^2 )^2 \, .
\end{equation}
The $\beta$-expansion of $\omega_T$ satisfies
\begin{align*}
\omega_T & \sim -\frac{2\log(2)}{\beta} + \mu -\frac{S_2}{16}  \beta +\frac{S_3}{96}  \beta^2   +     \frac{4 S_4 - 3 S_2^2 - Q}{1024}      \beta^3    \\  & 
+     \frac{S_5}{1280}       \beta^4  +     \frac{ 96 S_6 - 30 S^3_2 - 4 S^2_3 - 27 S_2 Q  }{147456}        \beta^5 +
O(\beta^6) \, .
\end{align*}
The individual $ \omega_{\alpha_i}$, for $i=1,\ldots,4$ can be recovered from the $\beta$-expansion of $\omega_T$ up to order $\beta^5$.
The coefficients of the $\beta$-expansion of $\omega_T$ up to fifth order determine the
$\omega_{\alpha_i}$, $i=1,2,3,4$. 
\end{lemma}

\begin{proof}
Let
\begin{equation}\label{rhobetax}
 \rho_\beta(x)=\Upsilon(x,\beta)+\frac{\log2}{\beta}-\frac{x}{2} \, ,
\end{equation}
so that $x\oplus_{\beta, {\Ry}_2} y=(x+y)/2 - \log(2)/\beta +\rho_\beta(x-y)$. 
We can then write
\begin{multline*}
 \omega_T = \mu - \frac{2\log(2)}{\beta} + \frac{ \rho_\beta(\omega_{\alpha_1}-\omega_{\alpha_2}) + \rho_\beta (\omega_{\alpha_3}-\omega_{\alpha_4}) }{2} \\
 + \rho_\beta\left( \frac{ \omega_{\alpha_1}+\omega_{\alpha_2}-\omega_{\alpha_3}-\omega_{\alpha_4}}{2} 
+ \rho_\beta(\omega_{\alpha_1}-\omega_{\alpha_2}) - \rho_\beta (\omega_{\alpha_3}-\omega_{\alpha_4}) \right) \, . 
\end{multline*}
The $\beta$-expansion of \eqref{expandYbetaall} gives 
\[
 \rho_\beta(x) = -\frac{x^2}{16}\beta - \frac{x^4}{512} \beta^3 - \frac{x^6}{6144} \beta^5 + O(\beta^7) \, . 
 \]
Thus we have an expansion 
\[
 \omega_T \sim -\frac{2\log(2)}{\beta} + c_0 + c_1 \beta + c_2 \beta^2 + c_3 \beta^3 + c_4 \beta^4 + c_5 \beta^5 + O(\beta^6), 
\]
where the coefficients $c_k$ are given by
 \begin{align*}
c_0  & = \mu \\
c_1 & = -\frac{S_2}{16} \\
c_2 & = \frac{S_3}{96} \\
c_3 & = \frac{4S_4-3S_2^2-Q}{1024} \\
c_4 & = \frac{S_5}{1280} \\
c_5 & =\frac{96S_6-30S_2^3-4S_3^2-27S_2Q}{147456}\, . 
\end{align*}
To see that the coefficients $c_i$, $i=0,\ldots, 5$ determine the $\omega_{\alpha_i}$, $i=1,2,3,4$, the argument
is similar to the case of Lemma~\ref{simpleMacc}. One shows that the  $\omega_{\alpha_i}-\mu$ are the roots of
the polynomial equation
\begin{equation}\label{poly4}
x^4 + 8 c_1 x^2 - 32 c_2 x - 192 c_3 - \frac{112}{3} c_1^2 + \frac{112 c_2^2 - 64 c_5}{c_1} =0 \, ,
\end{equation}
through the identity
\[
 \prod_{i=1}^4 (x - (\omega_{\alpha_i}-\mu)) = x^4 - \frac{S_2}{2} x^2 
- \frac{S_3}{3} x + \frac{S^2_2}{8} - \frac{S_4}{4} \, . 
\]
\end{proof}

A similar reconstruction argument works in the case of three leaves, though we do not
write it out here explicitly.

In general, the terms of the $\beta$-expansion provide polynomial expressions in the $\omega_{\alpha_i}$. Thus,
a possible way to determine the $\omega_{\alpha_i}$ would then be to solve these
polynomial constraints. However, when the number of inputs $\omega_{\alpha_i}$ grows, it is
not always possible to extend this reconstruction argument.

\begin{proposition}\label{reconprop}
Let $T$ be a nonplanar full binary rooted tree with leaves labelled by input phases
$\omega_{\alpha_\ell}$ and internal nodes $v$ labelled by $\omega_{T_v}=\omega_{T_{v_1}}\oplus_{S,\beta} \omega_{T_{v_2}}$, with $v_1,v_2$ the two children vertices below $v$ and $S$ the second R\'enyi
entropy, and with $\omega_T$ labeling the root of $T$. 
For number of leaves $n=\# L(T)\leq 8$, there is a dense open set of inputs $\omega_{\alpha_\ell}$,
$\ell\in L(T)$, such that the set $\{ \omega_{\alpha_\ell} \}_{\ell\in L(T)}$ can be reconstructed from the
first $n$ coefficients of the $\beta$-expansion
\begin{equation}\label{omegaTexpandn}
\omega_T = \frac{\gamma_T}{\beta} + c_{0,T} + c_{1,T} \beta + \cdots + c_{n,T} \beta ^n + O(\beta^{n+1}) \, .
\end{equation}
For $n=9$ leaves, consider the tree $T=\mathfrak{M}(T',\alpha)$ where $T'$ is the balanced tree on $8$ leaves
\[
 T' = \Tree[ [ [ $\alpha_1$  $\alpha_2$ ]  [ $\alpha_3$  $\alpha_4$   ]  ] [ [  $\alpha_5$  $\alpha_6$  ]   [  $\alpha_7$  $\alpha_8$   ] ] ] 
\]
and $\alpha$ is a single leaf. Then for all choices of $\omega_{\alpha_i}$ and $\omega_\alpha$ with 
\[
 \omega_\alpha \geq \frac{1}{8} \sum_{i=1}^8 \omega_{\alpha_i} 
\]
the input $\omega_\alpha$ is not reconstructible from $\omega_T$. 
\end{proposition}

A proof of Proposition~\ref{reconprop} is given in Appendix~\ref{App}. 

\smallskip

More generally, one can pose the following question.

\begin{question}\label{MSalg} 
		Under what conditions is there a Minimal Search algorithm that extracts a given $\omega_{T_v}$, for any chosen 
		accessible term $T_v\subset T$, from the synchronized phase $\omega_T$, observed as a function of the variable
		parameter $\beta$?
\end{question}

Note that this problem looks like an inversion of power series problem, which can 
be seen as essentially a Hopf algebraic renormalization problem, see \cite{Frab} and the
discussion of the Lagrange inversion formula in \cite{Gessel}  and its interpretation as antipode in
the Fa\`a di Bruno Hopf algebra, \cite{JoniRota}, and the relation between the
Fa\`a di Bruno Hopf algebra and the Connes--Kreimer Hopf algebra of rooted trees, in
the context of Dyson--Schwinger equations, \cite{Foissy}.
We will not discuss this problem further in this paper.

\begin{remark}\label{multivalbeta}
		If we allow a multivariable version, where each application of $\oplus_{S,\beta}$ corresponding to
		each non-leaf node $v$ of the tree $T$ can carry an independent parameter $\beta_v$, then the
		argument of Lemma~\ref{simpleMacc} would extend directly to the extraction of  
		$\omega_{T_v}$ from $\omega_T$, for any accessible term $T_v$. The more difficult
		Question~\ref{MSalg} pertains to the case where a single $\beta$-parameter is used.
\end{remark}

Whenever an algorithm as discussed above exists, that can extract $\omega_{T_v}$ from $\omega_T$
for an accessible term $T_v \subset T$, we then have the Internal Merge operation 
that maps 
\[
 \omega_T \mapsto \omega_{T_v}\oplus_{\Ry_2,\beta} \omega_{T/T_v}. 
\]

\subsection{Remark on Merge and the successor function} \label{succSec2}

In this framework, each application of Merge is realized as an application
of the successor function of the thermodynamic semiring, applied to the
phases, to achieve synchronization. As we have discussed, the
synchronization of phases is achieved by an averaging optimized by a
R\'enyi information functional $S=\Ry_2$, in the form of the
thermodynamic semiring addition $\oplus_{S,\beta}$. Equivalently,
both the External Merge operation $\omega_T=\omega_{T_1}\oplus_{\Ry_2,\beta} \omega_{T_2}$
for $T=\mathfrak{M}(T_1,T_2)$ and the Internal Merge operation 
$\omega_T \mapsto \omega_{T_v}\oplus_{\Ry_2,\beta} \omega_{T/T_v}$ are
expressible in terms of the {\em successor function of the semiring}.

Formal similarities between Internal Merge and the {\em successor function
	of arithmetic} have been observed (for instance in \cite{Chomsky2020}). Indeed
the von Neumann set theoretic construction of the integers and the
successor function of arithmetic, where one defines $0=\emptyset =\{ \ \ \}$,
$1=\{ \emptyset \}$, $2=\{ \emptyset , \{ \emptyset \} \}$, $3 =\{ \emptyset , \{ \emptyset \}, \{ \emptyset , \{ \emptyset \} \} \}$
and so on with the successor function $n \mapsto n+1$ defined in this set-theoretic form as
$\Upsilon: A_n \mapsto A_{n+1}=A_n \cup \{ A_n \}$.  Since the only information that finite sets carry
is their cardinality, the von Neumann construction indeed identifies a set-theoretic operation that is fully equivalent to the 
step $n \mapsto n+1$ leading from cardinality $n$
to cardinality $n+1$, namely the successor function of arithmetic. A formal similarity with Internal Merge
can be seen using the bracket notation for full binary rooted trees and writing the simplest Internal Merge in the form
$\{ \alpha_1, \alpha_2 \} \mapsto \{ \alpha_1, \{ \text{\sout{ $\alpha_1$ }}, \alpha_2 \}\}$, with \sout{ $\alpha_1$ }
denoting the trace left by the cancellation of the deeper copy. 
While this is an interesting and suggestive analogy, these two operations have different 
algebraic properties, hence they cannot be directly compared.

On the other hand, the model discussed shows that there is a precise way in which {\em both}
Internal Merge {\em and} External Merge behave like a successor function, not the one of arithmetic
but an appropriate one through which the algebraic properties of the Merge operation can be
realized. This will not provide a pathway to the successor function of arithmetic via the syntactic
Merge, as these remain different kinds of operation, but it does show that there is a sense in
which the successor function analogy is rigorous, if applied to the appropriate algebraic structure.

\subsection{Workspaces and Merge action as synchronization}

Consider the set of workspaces $\mathfrak{F}_{\mathcal{S}\mathcal{O}_0}$, namely binary rooted forests $F=\sqcup_a T_a$ whose
components are syntactic objects in $\mathcal{S}\mathcal{O}=\mathfrak{T}_{\mathcal{S}\mathcal{O}_0}$, and the linear span
$\mathcal{V}(\mathfrak{F}_{\mathcal{S}\mathcal{O}_0})$ (with real coefficients). We know from \cite{MCB} that Merge (in all its forms:
the optimal External and Internal Merge and the optimality violating Sideward Merge) can be realized
as a Hopf algebra Markov chain
\begin{equation}\label{KMergeWork}
	\mathcal{K} = \sqcup \circ (\mathcal{B} \otimes \id) \circ \Pi_{(2)} \circ \Delta 
\end{equation}
on $\mathcal{V}(\mathfrak{F}_{\mathcal{S}\mathcal{O}_0})$ endowed with the disjoint union product $\sqcup$
and the coproduct $\Delta$ that extracts forests of disjoint accessible terms via admissible cuts on
workspaces.

In the model discussed in this section, where we realize the magma operation $\mathfrak{M}$
as a phase synchronization on sinusoidal waves, we can also realize the Merge operations
in the same Hopf algebraic terms discussed in \cite{MCB}.

\begin{proposition}\label{HopfMergephi}
	The span $\mathcal{W}$ of finite products of wave functions $\varphi_T$ is endowed with a
	bialgebra structure, as the polynomial algebra in the $\varphi_T$ with coproduct
	\begin{equation}\label{coprodphiF}
		\Delta \varphi_F = \sum_{F_{\underline{v}}} \varphi_{T_{v_1}}\cdots  \varphi_{T_{v_n}} \otimes \varphi_{F/F_{\underline{v}}} \, , 
	\end{equation} 
	for $\underline{v}=(v_1,\cdots, v_n)$ with $F_{\underline{v}}=T_{v_1}\sqcup \cdots \sqcup T_{v_n}$ a forest of disjoint
	accessible terms. The map $\varphi: T \mapsto \varphi_T$ gives a bialgebra homomorphism
	$\varphi: \mathcal{V}(\mathfrak{F}_{\mathcal{S}\mathcal{O}_0}) \to \mathcal{W}$, from workspaces to their representation as waves. 
	The Merge action on workspaces is then implemented as the unique linear map that makes the diagram commute:
	\begin{equation}\label{diagrMergephi}
		\xymatrix{ \mathcal{V}(\mathfrak{F}_{\mathcal{S}\mathcal{O}_0}) \ar[d]_{\mathcal{K}} \ar[r]^{\varphi} & \mathcal{W} \ar[d]^{\mathcal{K}_\varphi} \\
			\mathcal{V}(\mathfrak{F}_{\mathcal{S}\mathcal{O}_0}) \ar[r]^{\varphi} & \mathcal{W} } 
	\end{equation}
	where $\mathcal{K}$ is the Hopf algebra Markov chain \eqref{KMergeWork},
	with $\Pi_{(2)}$ the projection on the terms of the coproduct with a forest with two
	components in the left-channel. 
	The map $\mathcal{K}_\varphi$ is also of the same
	Markov chain form. 
\end{proposition}

\begin{proof}
The first step is identifying what should play the role of workspaces $F\in \mathfrak{F}_{\mathcal{S}\mathcal{O}_0}$.
We need to associate to $F=\sqcup_a T_a$ a function $\varphi_F$,
built out of the sine waves $\varphi_{T_a}$. It is natural to expect that this mapping
$F \mapsto \varphi_F$ should give a morphism of commutative algebras. This means
that the target space should be a polynomial algebra in the functions $\varphi_T$ for
$T\in \mathcal{S}\mathcal{O}$ and that $\varphi_F =\prod_a \varphi_{T_a}$. Thus, this compatibility
with the commutative algebra structure of the space of workspaces $\mathcal{V}(\mathfrak{F}_{\mathcal{S}\mathcal{O}_0})$
implies that we encode workspaces $F$ as {\em products} of sinusoidal waves. 
Thus, we consider as the target space $\mathcal{W}$ where workspaces are represented
the linear span of products of sinusoidal waves $\varphi_T$ for $T\in \mathcal{S}\mathcal{O}$. 
On this space $\mathcal{W}$, we consider a coproduct of the form \eqref{coprodphiF}.

The extraction of accessible terms from a product $\varphi_F =\prod_a \varphi_{T_a}$ is possible,
given an algorithm for the extraction of the phases $\omega_{T_v}$ of the accessible terms $T_v$ of the
individual components $T_a$ from the phase $\omega_{T_a}$ of the wave function $\varphi_{T_a}$,
as discussed in \S \ref{AccTermsSec}. Indeed, it suffices to show that we can recover the phases
$\omega_{T_a}$ from the product $\varphi_F$, since then all the phases $\omega_{T_v}$ would also
be recoverable, while the frequencies and amplitudes are determined by those of the $\varphi_\alpha$
for the terms $\alpha$ at the leaves of $F$. All the $\varphi_{T_v}$ are then recoverable,
so that the extraction of accessible terms in \eqref{coprodphiF} is fully computable. 
For sine waves $\varphi_{T_a}=\sum_i A_{T_{a},i}\sin(\nu_{a,i} t +\omega_{T_a})$, 
the phases $\omega_{T_a}$ are recoverable from the finite product $\varphi_F =\prod_a \varphi_{T_a}$
via Fourier transform, which for a product of sine waves is given by a combination of delta functions
from whose weights the phases $\omega_{T_a}$ can be directly obtained (we can assume that
the $\varphi_{T_a}$ all have distinct frequencies $\nu_a$). 
Thus, the map $\varphi: \mathcal{V}(\mathfrak{F}_{\mathcal{S}\mathcal{O}_0}) \to \mathcal{W}$ then gives a bialgebra homomorphism.

Given the Hopf algebra Markov chain $\mathcal{K} : \mathcal{V}(\mathfrak{F}_{\mathcal{S}\mathcal{O}_0}) \to \mathcal{V}(\mathfrak{F}_{\mathcal{S}\mathcal{O}_0})$ of the Merge
action on workspaces \eqref{KMergeWork}, there is a corresponding $\mathcal{K}_\varphi: \mathcal{W} \to \mathcal{W}$
that makes the diagram \eqref{diagrMergephi} commute.
The map $\mathcal{K}_\varphi: \mathcal{W} \to \mathcal{W}$ takes the form
\[
 \mathcal{K}_\varphi = \mu \circ (\mathbb{B} \otimes \id) \circ \Pi_{(2)} \circ \Delta\, , 
\]
for $\Delta$ the coproduct in \eqref{coprodphiF} and $\mu$ the multiplication in
the algebra $\mathcal{W}$, and with
\[
 \mathbb{B} (\varphi_{T_1} \cdot \varphi_{T_2}) = \varphi_{\mathfrak{M}(T_1,T_2)} \, ,
\]
where $\varphi_{\mathfrak{M}(T_1,T_2)}$ is computed from the product $\varphi_{T_1} \cdot \varphi_{T_2}$
by first extracting the phases $\omega_{T_1}$ and $\omega_{T_2}$, then computing from
them $\omega_{\mathfrak{M}(T_1,T_2)}=\omega_{T_1} \oplus_{\Ry_2, \beta} \omega_{T_2}$ and
from this the resulting $\varphi_{\mathfrak{M}(T_1,T_2)}$.
\end{proof}

\subsection{Further question: introducing head functions} \label{HeadSec} 

We have mentioned head functions on syntactic objects in \S \ref{PolyTree}, where
a head function provided us with a choice at each internal node $v$ of either a factor
$\lambda_v$ or $1-\lambda_v$, so that the resulting probability distribution $A=(a_\ell)_{\ell\in L(T)}$
on the leaves of the tree depends not just on $T$, but on the pair $(T, h_T)$. This dependence,
however, is rather mild, and it eventually disappears entirely when the optimization 
\[
 \min_{\lambda_v} \{\lambda_v x_v + (1-\lambda_v) y_v - \beta^{-1} S(\lambda_v) \} 
\]
is computed, because of the symmetry $S(\lambda_v) =S(1-\lambda_v)$.

On the other hand, the presence of a head function is important, both for what is called the
``phase structure'' in syntax and the labeling algorithm, and for parsing in semantics. 
We know (\S 1.13 of \cite{MCB} and \cite{MHL}) that 
$\Dom(h)\subset \mathcal{S}\mathcal{O}$ is not a submagma. For $T,T'\in \Dom(h)$, 
the failure of $\mathfrak{M}(T,T')$ being in $\Dom(h)$ is coming from exocentric
constructions $\{ \mathrm{XP}, \mathrm{YP} \}$ where there is no a priori indication of which 
should be the head. The hypermagma structure discussed in \cite{MHL}
accounts for this issue.

The operation $\oplus_{\Ry_2,\beta}$ wipes out any explicit 
information on the head function encoded in the probability distribution $A=(a_\ell)_{\ell \in L(T)}$
(hence making it possible to encode the magma structure), so the presence of
the head has to be encoded here in a different way that is not directly involved in the
phase-to-phase synchronization where $\oplus_{\Ry_2,\beta}$ is acting.

\begin{figure}[h]
	\begin{center}
		\includegraphics[scale=0.68]{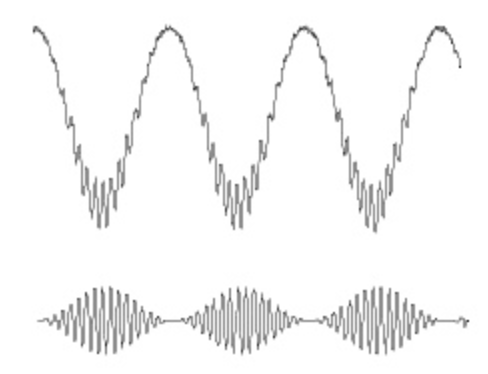} 
		\caption{Phase-amplitude coupling in sinusoidal waves. 
			\label{FigPAC}}
	\end{center}
\end{figure}

For example, the model of \cite{Murphy2,Murphy4}  proposes phase-amplitude coupling (PAC) as a possible
realization of headedness, where the phase of a lower frequency wave modulates the amplitude of 
a higher frequency wave, see Figure~\ref{FigPAC}. 

In the setting we are describing here, this would mean that when two
waves $\varphi_{T_1}$ and $\varphi_{T_2}$ are merged into the resulting
wave $\varphi_{\mathfrak{M}(T_1,T_2)}$ according to the phase synchronization 
$\omega_{\mathfrak{M}(T_1,T_2)}=\omega_{T_1}\oplus_{\Ry_2, \beta} \omega_{T_2}$,
as discussed above, if the head function $h_{\mathfrak{M}(T_1,T_2)}$ assigns to the
root vertex $v$ of $\mathfrak{M}(T_1,T_2)$ the same head $h_{\mathfrak{M}(T_1,T_2)}(v) = h_{T_1} (v_1)$,
where $v_1$ is the root vertex of $T_1$, then a lower frequency wave
modulates the amplitude of the wave $\omega_T(\varphi_{T_1})$ in the resulting 
superposition wave 
$\varphi_{\mathfrak{M}(T_1,T_2)}=\omega_T(\varphi_{T_1}) + \omega_T(\varphi_{T_2})$. 

Implementing this idea, however, has the problem that it leaves open the question 
of how the slower wave modulating the
faster wave $\omega_T(\varphi_{T_1})$ that carries the head
is to be determined: a mechanism for selecting such a modulating wave  
is not part of the type of model we have been discussing in this paper.

An alternative approach to incorporating head functions may be based on the
results of \cite{MarLar} and \cite{MHL}, where the filtering of syntactic objects by theta roles
assignments and by head, complement, specifier structure and well formed
phases can be formulated in terms of a colored operad and its generating
system. This suggests that in a representation of syntactic objects in terms
of superposition of phase-synchronized waves, the different local coloring rules
that mark head, complement, specifier, and modifier positions in the phase
structure, and theta role assignments, may possibly correspond to different 
PAC modulations. 

At present, the proposed theoretical models of Merge implementation available in the 
literature on neuroscience of language are not yet in a form that can be directly
translated into a full mathematical formalization. Thus, for the moment, we have just
provided some general possible models of mathematical operations that can be 
realized in a space of wavelets and that we can show faithfully capture the same 
algebraic properties of Merge. 
In particular, the cross-frequency phase synchronization that we described in this
section is not yet directly applicable for empirical comparison with data currently
available in the neuroscience literature, but it does provide a theoretical proof-of-concept
that such operations of phase synchronization can be abstractly designed to encode
the free symmetric Merge action.

\section{Conclusions}\label{ConcludeSec}

The main results we presented in this paper abstractly construct embeddings of
the nonassociative commutative magma of syntactic objects in function spaces of
wavelet-like functions, under the assumption that the set of lexical items can be
encoded in such a space. We focus on two main aspects of this construction:
\begin{enumerate}
	\item realizing the correct algebraic properties, namely nonassociativity, which is needed to
	have hierarchical structures in syntax, and commutativity, which characterizes the
	specific model of free symmetric Merge we are considering, in contrast to older 
	formulations of Minimalism;
	\item providing a good encoding of syntactic objects, in the sense that it should be
	possible from the wavelet-type function we associate to a given syntactic object to
	correctly reconstruct the syntactic object as a combinatorial structure, to ensure
	that the encoding we construct really preserves the information we wish to encode.
\end{enumerate}
We achieve these two goals in a chain of intermediate steps, that we summarize here.
\begin{itemize}
	\item It is first shown, in \S \ref{PolyTree},
	that syntactic objects can be faithfully encoded as convex combinations of the 
	functions associated to the lexical items at the leaves, according to a probability 
	distribution on the leaves determined by the tree structure of the syntactic object, 
	and depending polynomially on auxiliary parameters associated to the non-leaf 
	vertices of the tree. This encoding is not yet what we want, both because it does
	not realize the correct algebraic properties and because it requires the introduction
	of large numbers of additional arbitrary variables.
	\item We then show in \S \ref{RenyiSec} 
	that these additional variables can be optimized over (hence obtaining an actual
	object in the same function space where the lexical items are represented),
	through an optimization involving an entropy functional, the second R\'enyi
	entropy. This operation of optimized combination is itself 
	a nonassociative commutative magma operation and the resulting
	representation of syntactic objects becomes a magma homomorphism, so
	the algebraic properties are preserved. This nonassociative
	magma structure is compatible with the underlying linear structure of
	the target space of functions, in the sense that the vector space
	addition and the magma operation together form a (nonassociative)
	semiring. 
	\item While the above takes care of the correct algebraic properties,
	the question of being able to fully reconstruct syntactic objects from
	their image in this representation is more subtle. We first show in \S \ref{TransvSec}
	that this faithfulness property holds over a certain range of syntactic
	objects, namely those with a fixed (possibly very large but finite) 
	bound on the size (in number of leaves) and with all distinct lexical
	items at the leaves (no repetitions). The faithfulness, under these
	assumptions, is proved using topological transversality arguments,
	and holds in a generic sense (for a dense open set of choices
	of the representation of lexical items).
	\item The ``no repetitions'' assumption is not realistic linguistically,
	as one certainly does want to be able to represent syntactic
	objects containing repetitions of lexical items at the leaves. We show
	in \S \ref{RepCopSec} 
	that this condition can be circumvented by introducing more
	geometry, a configuration space of points (a product minus the
	diagonals, describing assignments with no repetitions) with 
	its compactification, in which identical occurrences are resolved
	by additional distinguishing data. This mirrors to some extent 
	the conceptual distinction in linguistics between copies and repetitions. 
	The size bound of \S \ref{HighTSec}, on the other hand, is more directly related to the
	distinction made in linguistics between competence and performance:
	we identify in \S \ref{SizeSec} an intrinsic source, in our model, of a progressive loss
	of performance as the size of the syntactic objects grows, in the
	form of estimates that typically are not uniform in the thermodynamic
	parameter of the construction, which has the effect of
	progressively shrinking the range in which faithfulness is realizable.
	\item We then consider other aspects of the action of Merge,
	besides the magma of syntactic objects, in particular workspaces
	and the action of Merge on workspaces that also realizes 
	movement via Internal Merge. In \S \ref{IMsec}, we represent workspaces, in 
	our model, as circuits built out of a basic binary gate that
	performs the entropy-optimized combination we described
	in \S\ref{RenyiSec} and the action of Merge as operations
	on such circuits. 
	\item In \S \ref{ROSEsec},  
	inspired by two current proposed neurolinguistics models, 
	in which hierarchical structures are related to operations
	of cross-frequency phase synchronizations, we describe
	more explicitly a special case of the general construction
	presented in this paper, where the basic functions representing
	lexical items are assumed to be simple sinusoidal waves
	and the Merge operation is implemented as a 
	cross-frequency phase synchronization involving
	the same kind of entropy-optimized combination 
	applied to phases. 
	\item In the process of developing this particular construction
	with cross-frequency phase synchronization, we also show in \S \ref{succSec} and \S \ref{succSec2}
	that the basic magma operation can be described in terms of the
	``successor function'' of the nonassociative semiring (which
	directly generalizes the usual successor function of
	arithmetic, which corresponds to the associative and
	commutative semiring on natural numbers). 
\end{itemize}

\appendix

\section{Reconstructing accessible terms in phase-synchronization}\label{App}

The purpose of this Appendix is to present the proof of Proposition~\ref{reconprop}.
Since the proof involves a case by case verification, performing a computation over 
each of the $47$ different tree topologies with up to $8$ leaves, we have automated this 
task by running the argument on GPT-6 Astra. This presentation is the
authors' writing based on that result.

We break Proposition~\ref{reconprop} into the two separate statements for
the failure of reconstructibility at $9$ leaves and for 
the reconstructibility for a dense open set of inputs, for up to $8$ leaves. 

\begin{lemma}\label{lem9leaves}
Consider the balanced tree on $8$ leaves 
\[
 T' = \Tree[ [ [ $\alpha_1$  $\alpha_2$ ]  [ $\alpha_3$  $\alpha_4$   ]  ] [ [  $\alpha_5$  $\alpha_6$  ]   [  $\alpha_7$  $\alpha_8$   ] ] ] 
\]
with input phases $\omega_{\alpha_i}$, $i=1,\ldots, 8$, and the tree
\[
 T= \Tree[ $T'$ $\alpha$ ] 
\]
with the additional input frequency $\omega_\alpha$. 
The input $\omega_\alpha$ is not reconstructibe from the resulting $\omega_T$, whenever
\begin{equation}\label{omegaaverage}
 \omega_\alpha > \frac{1}{8} \sum_{i=1}^8 \omega_{\alpha_i} \, .
\end{equation} 
\end{lemma}

\begin{proof}
Since $x\oplus_{S,\beta} y =\min_{\lambda\in [0,1]} \{ \lambda x + (1-\lambda)y + \beta^{-1} \log (\lambda^2 + (1-\lambda)^2) \}$ for $S={\Ry}_2$, we in particular have, by taking $\lambda=1/2$,
\begin{equation}\label{averbound}
x\oplus_{S,\beta} y \leq \frac{x+y}{2} - \frac{\log(2)}{\beta} \, . 
\end{equation}
Repeatedly applying \eqref{averbound} to each $\oplus_{S,\beta}$ operation at each
internal node of the balanced tree $T'$ with $\omega_{\alpha_i}$ inputs at the leaves, we
obtain
\begin{equation}\label{Tprimebound}
\omega_{T'} \leq \frac{1}{8} \sum_{i=1}^8 \omega_{\alpha_i} - \frac{ 3 \log(2)}{\beta} \, .
\end{equation}
Writing $x\oplus_{S,\beta} y =x + \Upsilon(y-x,\beta)$, and using 
the form \eqref{eqSuccu} of the successor function, with $u=\beta (x-y)/2$,
we see that  $x\oplus_{S,\beta} y 	= \min \{ x, y \}$ if  $ |x-y| \geq 2/\beta$.  
In particular
\begin{equation}\label{cutoffbound2}
  y-x \geq \frac{2}{\beta} \Longrightarrow x\oplus_{S,\beta} y 	= x \, .
\end{equation} 
Suppose  that $\omega_{\alpha_i}$ and $\omega_\alpha$ satisfy \eqref{omegaaverage}.
Then from \eqref{Tprimebound} we have
\[
 \omega_\alpha-\omega_{T'} \geq \omega_\alpha - \frac{1}{8} \sum_{i=1}^8 \omega_{\alpha_i} 
+ \frac{ 3 \log(2)}{\beta} \geq \frac{ 3 \log(2)}{\beta}  > \frac{2}{\beta} \, .  \]
Thus, for 
\[ \omega_T = \omega_{T'} \oplus_{S,\beta}  \omega_\alpha = \omega_{T'}
+ \Upsilon(\omega_\alpha-\omega_{T'},\beta)  \, , \]
we obtain from \eqref{cutoffbound2} that
\[
  \omega_T = \omega_{T'} 
  \]
so that the input $\omega_\alpha$ is not seen at all by $\omega_T$ and reconstruction fails. 
\end{proof}

The argument in this example can be adapted to obtain examples of failures of reconstructibility 
of  phases $\omega_{T_v}$ for accessible terms $T_v\subseteq T$ of some sufficiently large 
trees. We then consider the cases with up to $8$ leaves.

We define
\begin{equation}\label{varphix}
\varphi(x) \coloneqq \Upsilon(x, \beta)|_{\beta=1}  \ \ \  \text{ and } \ \ \   H(x,y)\coloneqq y + \varphi(x-y) \, ,
\end{equation}
so that we have
\begin{equation}\label{varphix2}
\Upsilon(x, \beta) = \beta^{-1}\, \varphi(\beta\, x)  \ \ \  \text{ and } \ \ \  x \oplus_{S,\beta} y = \beta^{-1}\, H(\beta\, x, \beta\, y) \, . 
\end{equation}
For a nonplanar full binary rooted tree $T$ with $n$ leaves with input 
$\underline{\omega}=(\omega_{\alpha_i})_{i=1}^n$, we also write 
\begin{equation}\label{PhiTv}
\Phi_{T_v}(\underline{\omega},\beta)\coloneqq \beta\, \omega_{T_v}(\underline{\omega},\beta) \, ,
\end{equation}
which satisfy $\Phi_\ell(\underline{\omega},\beta)=\beta \omega_{\alpha_\ell}$ at each
leaf $\ell \in L(T)$, and recursively
\begin{equation}\label{PhiTv12}
\Phi_{T_v} = H(\Phi_{T_{v_1}}, \Phi_{T_{v_2}} ) \, ,
\end{equation}
for $v_1, v_2$ the two children vertices below the vertex $v$. 
We also define $\gamma_v$ for all vertices of $T$ recursively by taking $\gamma_\ell =0$ at
the leaves and 
\begin{equation}\label{gammavH}
\gamma_v = H(\gamma_{v_1}, \gamma_{v_2}) 
\end{equation}
for the two children vertices $v_1, v_2$. We write $\gamma_T$ for the value at the root vertex of $T$,
and we write
\begin{equation}\label{deltav}
\delta_v \coloneqq\gamma_{v_1}-\gamma_{v_2}\, .
\end{equation}
The $\gamma_v$
of \eqref{gammavH} are the values $\Phi_{T_v}$ computed with all inputs $\omega_{\alpha_i}=0$.
For the balanced tree  on $n=2^k$ leaves $\gamma_T=-k \log(2)=-\log n$. 
The $\beta$-expansion of $\omega_{T_v}$ has leading term $\gamma_v/\beta$.

The gaps $\delta_v$ have the property that, if $|\delta_v|<2$, then $H$is real analytic
near the point $(\gamma_{v_1},\gamma_{v_2})$ and the composition of Taylor series
applies to compute $\Phi_{T_v}=H(\Phi_{T_{v_1}},\Phi_{T_{v_2}})$. If $\delta_v>2$ then
$\Phi_{T_v}=\Phi_{T_{v_2}}$ and if $\delta_v< -2$ then $\Phi_{T_v}=\Phi_{T_{v_1}}$. These
are the cases like what we have seen in Lemma~\ref{lem9leaves}, where the contribution $\Phi_{T_{v_i}}$ 
of one of the two branches of the tree $T_v=\mathfrak{M}(T_{v_1},T_{v_2})$ is not seen at all
by $\Phi_{T_v}$. The limiting cases where $\delta_v=\pm 2$ can have a more
complex behavior and we will not discuss it explicitly here. 

One can first observe that the same phenomenon that occurs the case of $9$ leaves in 
Lemma~\ref{lem9leaves} cannot happen for trees with at most $8$ leaves by
checking that, in all these cases, the gaps have $|\delta_v|<2$. 

\begin{lemma}\label{nooutgap8}
For any nonplanar full binary rooted tree $T$ with $n\leq 8$ leaves, the value $\gamma_T$ satisfies
\begin{equation}\label{gammaTbound8}
-\log(n) \leq \gamma_T \leq 0 \, .
\end{equation}
Thus, all gaps $\delta_v$ satisfy $|\delta_v|<2$. 
\end{lemma}

\begin{proof} 
Writing $C_T\coloneqq-\gamma_T$, the recursion \eqref{gammavH} is
\[ C_{T_v}=\max_{\lambda_v \in [0,1]} \{ \lambda_v C_{T_{v_1}} + (1-\lambda_v)  C_{T_{v_2}} + \Ry_2(\lambda_v) \} \, , \]
with $C_\ell=0$ at the leaves, so $C_T\geq 0$ inductively. The second R\'enyi entropy is bounded by
the Shannon entropy, $\Ry_2(\lambda)\leq \Sh(\lambda)$ by concavity of the logarithm,
\[ \log\bigl(\lambda^2+(1-\lambda)^2\bigr)
 \geq\lambda\log\lambda+(1-\lambda)\log(1-\lambda) \, . \]
Thus we have, for $n_i=\# L(T_{v_i})$, 
\[
 C_T \leq\max_{\lambda \in [0,1]}
 \{\lambda\log(n_1)+(1-\lambda)\log(n_2)
 +\Sh(\lambda)\} =\log(n_1+n_2)=\log(n). \]
Thus, all the gaps satisfy $|\delta_v|=| C_{T_{v_1}}- C_{T_{v_2}} |\leq \log(7) < 2$, unlike the case of
$9$ leaves where one of the gaps can reach value $\log(8)>2$. 
\end{proof}

We can then show reconstruction of the inputs over a dense open set for all trees with
at most $8$ leaves, from the $\beta$-expansion of $\omega_T$.

In terms of the $\beta$-normalized $\Phi_{T_v}$ we write the $\beta$-expansion as
\begin{equation}\label{PhiTvbetaexp}
\Phi_{T_v} (\underline{\omega},\beta) = \gamma_v + \sum_{k\geq 1} A_{T_v,k}(\underline{\omega}) \, \beta^k \, , 
\end{equation}
where the $A_{T_v,k}$ are polynomials in the $\omega_{\alpha_i}$.

\begin{lemma}\label{upto8}
Let $T$ be a nonplanar full binary rooted tree with $n\leq 8$ leaves. 
There is a dense open set $U_T$ of input phases
$\underline\omega=(\omega_{\alpha_i})_{i=1}^n$ that are reconstructible from the first
$n$ coefficients of the $\beta$-expansion \eqref{PhiTvbetaexp}. 
\end{lemma}

\begin{proof} The case $n=1$ is immediate since the first term is the input $\omega$ itself. 
For $2\leq n \leq 8$, consider the function 
\[ F_T(\underline\omega) =(A_{T,1}(\underline\omega), \ldots, A_{T,n}(\underline\omega))\, , \]
where $A_{T,k}(\underline\omega)$ are the coefficients of the $\beta$-expansion \eqref{PhiTvbetaexp}
of $\Phi_T$, and let $\det JF_T$ be the Jacobian determinant of $F_T$. Then one has local
invertibility on the open set
\[ U_T =\{ \underline\omega \, : \, \det JF_T(\underline\omega)) \neq 0 \} \, . \]
If $U_T\neq \emptyset$ it is dense in the Euclidean topology, 
since the $A_{T,k}$ are polynomials, so $U_T$ is Zariski open.

Thus, one just needs to verify that for every tree $T$ with up to $8$ leaves, $U_T\neq \emptyset$.
To this purpose, it suffices to choose a probe-input $\underline\omega^{(0)}=(\omega^{(0)}_i)$
on which to verify that 
\begin{equation}\label{probeJF}
\det JF_T(\underline\omega^{(0)}) \neq 0\, . 
\end{equation}
We choose probe-points 
$\omega^{(0)}_i\in [0,1]$ for $i=1,\ldots,n$, 
\begin{equation}\label{probes}
 \omega^{(0)}_i =  \frac{2^{i-1}-1}{2^{n-1}-1}  \,  .
\end{equation} 
For each $2 \leq n\leq 8$ the number of different tree topologies is $1,1,2,3,6,11,23$, so
there are in total $47$ cases for which to check that \eqref{probeJF} holds
at the points \eqref{probes}.

For each of these cases the polynomials $A_{T_v,k}$ can be computed through a recursion. 
For $A_{T_v,k}$ as in \eqref{PhiTvbetaexp}, let
\begin{equation}\label{Dvk}
D_{v,k}=A_{T_{v_1},k} - A_{T_{v_2},k} \, ,
\end{equation}
with $v_1, v_2$ the two children vertices of $v$. The $A_{T_v,k}(\underline{\omega})$ are homogeneous
polynomials of degree $k$, with $A_{\ell,1}=\omega_{\alpha_\ell}$ and $A_{\ell,k}=0$ for $k>1$ and $\ell\in L(T)$
a leaf, while at a non-leaf vertex 
\[ A_{T_v,1}=\lambda_v A_{T_{v_1},1}+ (1-\lambda_v) A_{T_{v_2},1} , \ \ \ \text{ with } \ \
\lambda_v= \varphi^\prime(\delta_v), \]
for $\delta_v =\gamma_{v_1}-\gamma_{v_2}$. The composition of Taylor series then gives the recursion
formula for the polynomials $A_{T_v,k}$ of the form
\begin{equation}\label{recurAtvk}
A_{T_v,k} = A_{T_{v_2},k} + \sum_{r=1}^k \frac{\varphi^{(r)}(\delta_v)}{r!} \sum_{k_1+\cdots+k_r=k} 
D_{v,k_1} \cdots D_{v,k_r} \, . 
\end{equation}
A case by case computation run on GPT-6 Astra verified \eqref{probeJF} on all $47$ cases. So 
reconstruction of the input phases holds on the dense open set $U_T$. 
\end{proof}

%%%%%%%%%%%%%%%%%%%%%%%%%%%%%%%%%%%%%%%%%%%%%%%%%%%%%%%%%%%%%%%%%%%
%%                                                               %%
%% Use the two commands below for producing your bibliography    %%
%% with bibtex, then comment again the commands and include the  %%
%% content of the .bbl file in this file below the commands.     %%
%%                                                               %%
%%%%%%%%%%%%%%%%%%%%%%%%%%%%%%%%%%%%%%%%%%%%%%%%%%%%%%%%%%%%%%%%%%%

%\bibliographystyle{amsplainhyper}  
%\bibliography{bibliomna}

% add below the content of your .bbl file produced by bibtex.
%

\providecommand{\bysame}{\leavevmode\hbox to3em{\hrulefill}\thinspace}
\providecommand{\MR}{\relax\ifhmode\unskip\space\fi MR }
\providecommand{\MRhref}[2]{%
	\href{http://www.ams.org/mathscinet-getitem?mr=#1}{#2}
}
\providecommand{\bysame}{\leavevmode\hbox to3em{\hrulefill}\thinspace}

%%%%%%%%%%%%%%%%%%%%%%%%%%%%%%%%%%%%%%%%%%%%%%%%%%%%%%%%%%%%%%%%%%%
%%                                                               %%
%% You may add acknowledgments (optional).                       %%
%%                                                               %%
%%%%%%%%%%%%%%%%%%%%%%%%%%%%%%%%%%%%%%%%%%%%%%%%%%%%%%%%%%%%%%%%%%%

\ACKNO{The first author is supported by the National Science Foundation, 
	grants DMS-2104330 and DMS-2506176, by Caltech's Center of Evolutionary Science,
	and by Caltech's T\&C Chen Center for Systems Neuroscience, and by an 
	OpenAI Academic Researchers grant [GPT-6 Astra used in Appendix~\ref{App}]. We thank Andrea Martin and 
	Elliot Murphy for very useful discussions and for providing extensive comments on a draft of this paper. 
	We acknowledge helpful comments from Juan Pablo Vigneaux and several other participants 
	in the workshop on {\em Mathematics of Generative Linguistics} at Caltech's Merkin Center, 
	May 31--June 1, 2025. We thank Etienne Tanr\'e,  as editor at Mathematical Neuroscience 
	and Applications, for another careful reading and helpful comments that led to further
	improvements to the manuscript. }

%%%%%%%%%%%%%%%%%%%%%%%%%%%%%%%%%%%%%%%%%%%%%%%%%%%%%%%%%%%%%%%%%%%
%%                                                               %%
%% You have reached the end of your document.                    %%
%%                                                               %%
%%%%%%%%%%%%%%%%%%%%%%%%%%%%%%%%%%%%%%%%%%%%%%%%%%%%%%%%%%%%%%%%%%%
%\nocite{*}
\end{document}